\documentclass[11pt]{article}

\usepackage[utf8]{inputenc} 
\usepackage[T1]{fontenc}    
\usepackage{hyperref}       
\usepackage{url,authblk}            
\usepackage{booktabs}       
\usepackage{amsfonts}       
\usepackage{nicefrac}       
\usepackage{microtype}      
\usepackage{xcolor}         

\usepackage[round]{natbib}
\usepackage{graphicx}
\usepackage{amsmath,amssymb,mathtools,amsthm}
\usepackage{mathrsfs}
\usepackage{dirtytalk}
\usepackage{tikz}
\usepackage{placeins}
\usepackage{csvsimple}
\usepackage{siunitx}
\usepackage{caption}
\usepackage{subcaption}
\newcommand{\R}{\mathbb{R}}
\newcommand{\E}{\mathbb{E}}
\newcommand{\N}{\mathbb{N}}
\newcommand{\I}{\mathfrak{I}}
\newcommand{\HS}{\mathcal{L}^{2}}
\newcommand{\Op}{\mathcal{L}^{\infty}}
\newcommand{\Hil}{\mathcal{H}}

\newcommand{\repone}{\phi}
\newcommand{\reptwo}{\psi}
\newcommand{\repthree}{\varphi}
\newcommand{\Hone}{\mathcal{H}_{\phi}}
\newcommand{\Htwo}{\mathcal{H}_{\psi}}
\newcommand{\Hf}{\mathcal{H}_{f}}
\newcommand{\Hg}{\mathcal{H}_{g}}
\newcommand{\Hthree}{\mathcal{H}_{\varphi}}
\newcommand{\norm}[1]{\left\|#1\right\|}
\newcommand{\inprod}[1]{\left \langle #1 \right\rangle}

\DeclareMathOperator*{\argmin}{arg\,min}
\DeclareMathOperator*{\Tr}{\text{Tr}}
\newcommand{\LPtwo}{L^{2}(P_{X})}
\newcommand{\metricstname}{UKP }
\newcommand{\metricfullname}{Uniform Kernel Prober}

\theoremstyle{plain}

\newcounter{lemmano}
\newcounter{theoremno}
\newcounter{propositionno}

\newcounter{corollarynno}
\newcounter{definitionno}

\newtheorem{theorem}[theoremno]{Theorem}
\newtheorem{lemma}[lemmano]{Lemma}
\newtheorem{proposition}[propositionno]{Proposition}

\newtheorem{corollary}[corollarynno]{Corollary}
\newtheorem{definition}[definitionno]{Definition}

\newenvironment{theorem*}{{\bf Lemma:}}

\title{Uniform Kernel Prober}

\author{Soumya Mukherjee}
\author{Bharath K.~Sriperumbudur}
\affil{Department of Statistics\\  
Pennsylvania State University, 
University Park, PA 16802, USA.\\
\texttt{\{szm6510,bks18\}@psu.edu}}

\begin{document}

\maketitle

\begin{abstract}
 The ability to identify useful features or representations of the input data based on training data that achieves low prediction error on test data across multiple prediction tasks is considered the key to multitask learning success. In practice, however, one faces the issue of the choice of prediction tasks and the availability of test data from the chosen tasks while comparing the relative performance of different features. In this work, we develop a class of pseudometrics called Uniform Kernel Prober (UKP) for comparing features or representations learned by different statistical models such as neural networks when the downstream prediction tasks involve kernel ridge regression. The proposed pseudometric, UKP, between any two representations, provides a uniform measure of prediction error on test data corresponding to a general class of kernel ridge regression tasks for a given choice of a kernel without access to test data. Additionally, desired invariances in representations can be successfully captured by UKP only through the choice of the kernel function and the pseudometric can be efficiently estimated from $n$ input data samples with $O(\frac{1}{\sqrt{n}})$ estimation error. We also experimentally demonstrate the ability of UKP to discriminate between different types of features or representations based on their generalization performance on downstream kernel ridge regression tasks.
\end{abstract}

\section{Introduction}

Model comparison is a classical problem in Statistics and Machine Learning \cite{burnham1998practical,pfahringer2000meta,spiegelhalter2002bayesian,caruana2006empirical,fernandez2014we}. This question has received tremendous attention from the scientific community, especially after the widespread adoption and implementation of modern general-purpose large-scale models such as deep neural networks (DNNs). Faced with the vast complexity and variation in mathematical representations (functional forms), sizes (no. of trainable parameters), and levels of model transparency (open to public vs. black-box/query access), it is an ongoing challenge to develop criteria for model comparison that is general and widely applicable to a large class of models as well as choice of learning tasks. 

In the supervised learning setting, where the goal is to predict the correct outputs given some inputs, it is natural to compare models based on relative differences in predictive performance, as this aligns directly with the objective of maximizing model accuracy on the supervised learning task. It is now well understood that the key to success for training models with good generalization ability over multiple tasks (i.e. achieves low prediction error on test data across multiple prediction tasks) is directly correlated to the ability of models to identify useful features or representations of the input data based on training data \cite{bengio2013representation,lecun2015deep,maurer2016benefit}. Therefore, one can attempt to resolve the question of model comparison by considering metrics (more precisely, pseudometrics) on the space of features or representations, and there is extensive literature in this area \cite{laakso2000content,li2015convergent,morcos2018insights,wang2018towards,kornblith2019similarity,GULP}.

  An ideal pseudometric must be interpretable and efficiently computable based on a reasonably small amount of data samples. It must also be sensitive only to differences in features that will lead to differences in predictive performance, but be fairly insensitive to any other differences in features that do not affect predictive performance. Finally, it must be flexible enough to accommodate available prior knowledge about the class of prediction tasks that is of interest to the model users. However, most pseudometrics fall short of fulfilling this extensive set of desiderata. In this work, we develop a class of pseudometrics on the space of representations called Uniform Kernel Prober (UKP) that can be used to compare features or representations learned by any class of statistical models. 

 The proposed pseudometric is motivated by the need for a distance measure over representations of differing 
 dimensionalities that captures the ability of a model to generalize over a general and flexible class of prediction tasks, specifically, the class of kernel ridge regression-based tasks. Depending on the choice of the kernel, one can probe which models share ``similar" features, with similarity being understood in the following sense: If the features or representations for a pair of models are similar, then, if they are both trained to perform kernel ridge regression tasks, their predictive performances will be close to each other.
 
 The proposed \metricstname pseudometric is a unique distance measure over features or representations and is a useful contribution to the existing literature since it has the following desirable characteristics:
\begin{enumerate}
    
     \item The proposed pseudometric offers a uniform guarantee of performance similarity for a wide range of regression functions, irrespective of whether the tasks are kernel ridge-regression or not. This is particularly beneficial when the prediction tasks align with models whose representations share similar characteristics with the kernel used to compute the \metricstname distance.

     \item The pseudometric is adaptable to incorporate inductive biases that help identify models suited for specific tasks. A simple choice of the kernel parameter of the \metricstname distance can help us encode these inductive biases. For example, suppose we are interested in image classification tasks where the rotation of the images should not affect the model prediction. In that case, we can encode this inductive bias into the pseudometric by choosing a rotationally invariant kernel, such as a Gaussian RBF kernel, as the kernel parameter for UKP. This results in the creation of two clusters: one for models with rotationally invariant features and another for models without such features.
    
    To the best of our knowledge, ours is the first pseudometric on the space of representations in the ML literature that can flexibly encode a wide range of inductive biases and treat them within a single framework.

    \item \metricstname distance has a practical prediction-based interpretation in addition to usual mathematical interpretations of similarity or dissimilarity in terms of inner product or pseudometric.
    
    \item Computation of the estimate of \metricstname distance only requires unlabelled data, i.e., data samples from the input domain, and therefore preserves labeled data for model training/fitting. Moreover, the computation of the estimate of \metricstname distance only requires black-box access to model representations, i.e., pairs of inputs and outputs to the model. 
    \item It is possible to design a statistically efficient estimator for the \metricstname distance based on a finite number ($n$) of samples from the input domain, that enjoys an estimation error rate of $n^{-1/2}$.
    
    \item The \metricstname distance enables us to even compare representations that differ in their dimensionalities. 
    
\end{enumerate}

The paper is organized as follows. In Section \ref{Problem Setup}, we formally define the \metricstname distance. In Section \ref{Properties}, we provide different characterizations of the \metricstname distance and prove that it satisfies all criteria of being a pseudometric. Then using Lemma \ref{Invariance of regularized kernel inner product}, we also find the type of transformations under which the \metricstname distance remains invariant. We propose a statistical estimator of the \metricstname distance in Section \ref{statistical estimation of UKP}. In Sections \ref{Relation to other comparison measures} and \ref{Finite sample convergence rate}, we mathematically demonstrate its relationship to other pseudometrics used for model comparison and show that our proposed estimator converges to the true \metricstname distance as the sample size goes to infinity. Finally, in Section \ref{experiments}, we provide numerical experiments that validate our theory. Proofs of all lemmas, propositions and theorems are provided in Section \ref{Proofs} of the Appendix. 

\section{Problem setup}\label{Problem Setup}

Let the input/predictor of the model be $X \in \R^d$ and $P_{X}$ be the distribution of the input. Let $\repone: \R^d \to \R^k$ and $\reptwo: \R^d \to \R^l$ be two instances of a representation map that transforms an input to a feature representation used in a trained/fitted model. Let $Y$ be the random real-valued response corresponding to the input $X$ generated from the nonparametric regression model $Y=\eta(X) + \epsilon$ where $\epsilon$ is mean-zero noise, where $\eta(x) = \E(Y \mid X = x)$ is the population regression function of $Y$ on $X$.

Let $K(\cdot,\cdot)$ be a positive definite, symmetric, bounded, and continuous kernel function, mapping pairs of vectors in Euclidean spaces of different dimensions to real numbers. Examples of radial kernels include the Gaussian RBF kernel $K_{RBF,h}(x,y) = \exp(-\frac{1}{2h}\norm{x-y}_{2}^{2})$ and the Laplace kernel $K_{Lap,h}(x,y) = \exp(-\frac{1}{2h}\norm{x-y}_{1})$, where $x,y\in \R^{d}$ for any $d\in \N$. By the Moore-Aronszajn Theorem \citep{aronszajn1950theory} and Lemma 4.33 of \citet{steinwart2008support}, there exists a unique separable Reproducing Kernel Hilbert Space (RKHS) $\Hil$ of functions such that $K(\cdot,\cdot)$ is its unique reproducing kernel. Theorem 5.7 of \citet{paulsen2016introduction}
ensures that $K_{\repone}(\cdot,\cdot) \coloneq K(\repone(\cdot),\repone(\cdot))$ and $K_{\reptwo}(\cdot,\cdot) \coloneq K(\reptwo(\cdot),\reptwo(\cdot))$ are the unique reproducing kernels corresponding to the ``pullback'' RKHS's $\Hone \coloneq \Hil\left(K \circ \left(\repone \times \repone\right)\right)$ and $\Htwo \coloneq \Hil\left(K \circ \left(\reptwo \times \reptwo\right)\right)$. Further, let $\Hil^{k}$ and $\Hil^{l}$ be the RKHS's associated with the kernel $K$ when the domain is restricted to $\R^{k} \times \R^{k}$ and $\R^{l} \times \R^{l}$, respectively. Then, for any $f_{\repone} \in \Hone$, we have $\norm{f_{\repone}}_{\Hone}=\underset{f \in \Hil^{k}: f \circ \repone = f_{\repone}}{\min}\norm{f}_{\Hil^{k}}$ and for any $f_{\reptwo} \in \Htwo$, we have $\norm{f_{\reptwo}}_{\Htwo}=\underset{f \in \Hil^{l}: f \circ \reptwo = f_{\reptwo}}{\min}\norm{f}_{\Hil^{l}}$. 

For any $\lambda>0$, let $\alpha_{\lambda}$ and $\beta_{\lambda}$ be the population kernel ridge regression estimators of the regression function $\eta$, given by
\begin{equation}\label{Defintion of alpha}
    \alpha_{\lambda} = \underset{\alpha \in \Hone}{\argmin} \hspace{2pt} \E\left[Y-\alpha(X)\right]^{2} + \lambda \norm{\alpha}_{\Hone}^{2}
\end{equation}
and
\begin{equation}\label{Defintion of beta}
\beta_{\lambda} = \underset{\beta \in \Htwo}{\argmin} \hspace{2pt} \E\left[Y-\beta(X)\right]^{2} + \lambda \norm{\beta}_{\Htwo}^{2},
\end{equation}
respectively. The prediction loss being the squared error loss, $\alpha_{\lambda}$ and $\beta_{\lambda}$ depend on the distribution of $Y$ only through the population regression function $\eta$.

We now define the kernel ridge regression-based pseudometric between the two representations of the input $\repone$ and $\reptwo$, based on the difference between predictions for $Y$ uniformly over all regression functions $\eta \in \LPtwo$ such that its $\LPtwo$ norm is bounded above by 1.
\begin{definition}\label{Definition of pseudometric}
    For any $\lambda>0$ and choice of kernel $K(\cdot,\cdot)$, the \metricstname (\metricfullname) distance between representations $\repone(X)$ and $\reptwo(X)$ is defined as, 
    \[
    d_{\lambda,K}^{\metricstname}(\repone,\reptwo) \coloneq \underset{\norm{\eta}_{\LPtwo} \leq 1}{\sup} \left(\E\left[\alpha_{\lambda}(X)-\beta_{\lambda}(X)\right]^{2}\right)^{\frac{1}{2}},
    \]
    where $\alpha_{\lambda}$ and $\beta_{\lambda}$ are defined in Equations \eqref{Defintion of alpha} and \eqref{Defintion of beta}, respectively.
\end{definition}

\section{Properties of $d_{\lambda,K}^{\text{\metricstname}}$} \label{Properties}

Let $\I_{\repone}: \Hone \to \LPtwo, f \to f$ be the inclusion operator, which maps any $f \in \Hone$ to its representation $f \in \LPtwo$. Then the adjoint of the inclusion operator is given by $\I_{\repone}^{*}:\LPtwo \to \Hone, f \to \int K_{\repone}(\cdot,x)f(x)dP_{X}(x)$. The inclusion operator $\I_{\reptwo}$ and the corresponding adjoint operator $\I_{\reptwo}^{*}$ can be analogously defined. 

Let us define the covariance operators corresponding to the RKHS's $\Hone$ and $\Htwo$ as
\[
\begin{aligned}
    \Sigma_{\repone} \coloneq \int K_{\repone}(\cdot,x) \otimes_{\Hone}  K_{\repone}(\cdot,x) dP_{X}(x)= \int K(\repone(\cdot),\repone(x)) \otimes_{\Hone} K(\repone(\cdot),\repone(x)) dP_{X}(x)
\end{aligned}
\]
and 
\[
\begin{aligned}
    \Sigma_{\reptwo} \coloneq \int K_{\reptwo}(\cdot,x) \otimes_{\Htwo}  K_{\reptwo}(\cdot,x) dP_{X}(x)= \int K(\reptwo(\cdot),\reptwo(x)) \otimes_{\Htwo} K(\reptwo(\cdot),\reptwo(x)) dP_{X}(x).
\end{aligned}
\]

$\Sigma_{\repone}: \Hil_{\repone} \to \Hil_{\repone}$ and $\Sigma_{\reptwo}: \Hil_{\reptwo} \to \Hil_{\reptwo}$ are the unique operators that satisfy 
\[
    \inprod{\Sigma_{\repone}f_{1},g_{1}}_{\Hone} = \E\left[f_{1}(X)g_{1}(X)\right]
\]
and
\[
    \inprod{\Sigma_{\reptwo}f_{2},g_{2}}_{\Htwo} = \E\left[f_{2}(X)g_{2}(X)\right],
\]
where $f_{1},g_{1} \in \Hone$ and $f_{2},g_{2} \in \Htwo$, respectively. In terms of inclusion operators, it can be easily shown that $\Sigma_{\repone} = \I_{\repone}^{*}\I_{\repone}$ and $\Sigma_{\reptwo} = \I_{\reptwo}^{*}\I_{\reptwo}$.

Let us define the integral operators corresponding to the RKHS's $\Hone$ and $\Htwo$ as follows:

\[
\begin{aligned}
    \mathcal{T}_{\repone} f &\coloneq \int K_{\repone}(\cdot,x) f(x) dP_{X}(x)
\end{aligned}
\]
and 
\[
\begin{aligned}
    \mathcal{T}_{\reptwo} f &\coloneq \int K_{\reptwo}(\cdot,x) f(x) dP_{X}(x),
\end{aligned}
\]

for any $f \in \LPtwo$.
It is also easy to show that $\mathcal{T}_{\repone} = \mathfrak{I}_{\repone}\mathfrak{I}_{\repone}^{*}$ and $\mathcal{T}_{\reptwo} = \mathfrak{I}_{\reptwo}\mathfrak{I}_{\reptwo} ^{*}$. The boundedness and continuity of the kernel $K$ ensures that $\Sigma_{\repone}$, $\Sigma_{\reptwo}$, $\mathcal{T}_{\repone}$ and $\mathcal{T}_{\reptwo}$ are all compact trace-class operators, which consequently ensures that they are also Hilbert-Schmidt operators. Further, each of $\Sigma_{\repone}$, $\Sigma_{\reptwo}$, $\mathcal{T}_{\repone}$ and $\mathcal{T}_{\reptwo}$ are self-adjoint positive operators and therefore have a spectral representation \citep[Theorems~VI.16,VI.17]{reed1980methods}.

For any $\lambda>0$, the regularized inverse covariance operators are defined as $\Sigma_{\repone}^{-\lambda} \coloneq \left(\Sigma_{\repone} + \lambda I\right)^{-1}$ and $\Sigma_{\reptwo}^{-\lambda} \coloneq \left(\Sigma_{\reptwo} + \lambda I\right)^{-1}$ , while the corresponding square roots are defined as  $\Sigma_{\repone}^{-\frac{\lambda}{2}} \coloneq \left(\Sigma_{\repone} + \lambda I\right)^{-\frac{1}{2}}$ and $\Sigma_{\reptwo}^{-\frac{\lambda}{2}} \coloneq \left(\Sigma_{\reptwo} + \lambda I\right)^{-\frac{1}{2}}$. Further, let us  define $\widetilde{K}_{\repone}(x,y) \coloneq \Sigma_{\repone}^{-\frac{\lambda}{2}}K_{\repone}(x,y)$ and $\widetilde{K}_{\reptwo}(x,y)\coloneq \Sigma_{\reptwo}^{-\frac{\lambda}{2}}K_{\reptwo}(x,y)$.

The \metricstname distance has the following characterization:
\begin{lemma}\label{Characterization of pseudometric in terms of expected squared distance}
For any $\lambda>0$, the squared \metricstname distance $d_{\lambda,K}^{\emph{\metricstname}}(\repone,\reptwo)$ between representations $\repone(X)$ and $\reptwo(X)$ can be expressed as
\[
\begin{aligned}
     \left[d_{\lambda,K}^{\emph{\metricstname}}(\repone,\reptwo)\right]^{2}
    &= \E  \left[\inprod{\Sigma_{\repone}^{-\frac{\lambda}{2}}K_{\repone}(\cdot,X),\Sigma_{\repone}^{-\frac{\lambda}{2}}K_{\repone}(\cdot,X^{\prime})}_{\Hone} -\inprod{\Sigma_{\reptwo}^{-\frac{\lambda}{2}}K_{\reptwo}(\cdot,X),\Sigma_{\reptwo}^{-\frac{\lambda}{2}}K_{\reptwo}(\cdot,X^{\prime})}_{\Htwo}\right]^{2}\\
    & = \E  \left[\inprod{K_{\repone}(\cdot,X),\Sigma_{\repone}^{-\lambda}K_{\repone}(\cdot,X^{\prime})}_{\Hone} -\inprod{K_{\reptwo}(\cdot,X),\Sigma_{\reptwo}^{-\lambda}K_{\reptwo}(\cdot,X^{\prime})}_{\Htwo}\right]^{2},
\end{aligned}
\]
where $X$ and $X^{\prime}$ are i.i.d observations drawn from $P_{X}$.
\end{lemma}

The proof is provided in Section \ref{Proof of Lemma 1} of the Appendix. The above characterization shows that the \metricstname induces an isometric embedding $\repone \mapsto \inprod{\Sigma_{\repone}^{-\frac{\lambda}{2}}K_{\repone}(\cdot,X),\Sigma_{\repone}^{-\frac{\lambda}{2}}K_{\repone}(\cdot,X^{\prime})}_{\Hone}$ of $\repone$ into $L^{2}(P_{X}^{\otimes 2})$. This characterization allows us to prove Proposition \ref{Proposition: Squared pseudometric using trace operators}, which will be useful throughout the rest of the paper.

Next, we show that the \metricstname distance can be expressed in terms of the trace operator, which will be essential for developing a statistical estimator of the pseudometric based on random samples from the input distribution $P_{X}$. 

To do so, we define the cross-covariance operators $\Sigma_{\repone\reptwo}: \Htwo \to \Hone$ and $\Sigma_{\reptwo\repone}: \Hone \to \Htwo$ as follows:
\[
\begin{aligned}
    &\Sigma_{\repone\reptwo} \coloneq\int K_{\repone}(\cdot,x) \otimes_{\HS(\Htwo,\Hone)}  K_{\reptwo}(\cdot,x) dP_{X}(x)\\ &= \int K(\repone(\cdot),\repone(x)) \otimes_{\HS(\Htwo,\Hone)}  K(\reptwo(\cdot),\reptwo(x)) dP_{X}(x)
\end{aligned}
\]and
\[
\begin{aligned}
    &\Sigma_{\reptwo\repone} \coloneq \int K_{\reptwo}(\cdot,x) \otimes_{\HS(\Hone,\Htwo)} K_{\repone}(\cdot,x) dP_{X}(x)\\
    &= \int  K(\reptwo(\cdot),\reptwo(x))\otimes_{\HS(\Hone,\Htwo)}  K(\repone(\cdot),\repone(x))dP_{X}(x)
    = \Sigma_{\repone\reptwo}^{*}.
\end{aligned}
\]

\begin{proposition}\label{Proposition: Squared pseudometric using trace operators}
    For any $\lambda>0$, the squared \emph{\metricstname} distance $d_{\lambda,K}^{\emph{\metricstname}}(\repone,\reptwo)$ between representations $\repone(X)$ and $\reptwo(X)$ can be expressed as
    \[
\begin{aligned}
        \left[d_{\lambda,K}^{\emph{\metricstname}}(\repone,\reptwo)\right]^{2}
        = \emph{Tr}\left(\Sigma_{\repone}^{-\lambda}\Sigma_{\repone}\Sigma_{\repone}^{-\lambda}\Sigma_{\repone}\right) + \emph{Tr}\left(\Sigma_{\reptwo}^{-\lambda}\Sigma_{\reptwo}\Sigma_{\reptwo}^{-\lambda}\Sigma_{\reptwo}\right)-2\emph{Tr}\left(\Sigma_{\repone}^{-\lambda}\Sigma_{\repone\reptwo}\Sigma_{\reptwo}^{-\lambda}\Sigma_{\reptwo\repone}\right).
    \end{aligned}
    \]
\end{proposition}

The proof is provided in Section \ref{Proof of Proposition 1} of the Appendix. The following theorem serves to show that the \metricstname distance does satisfy the axioms of a pseudometric.

\begin{theorem}\label{Theorem: Pseudometric}
    For any $\lambda>0$, the $d_{\lambda,K}^{\emph{\metricstname}}$ distance satisfies the following properties:
    \begin{enumerate}
        \item For any function $\repone: \R^d \to \R^k$ for some $k \in \N$, $d_{\lambda,K}^{\emph{\metricstname}}(\repone,\repone) = 0$,
        \item (Non-negativity) For any two functions $\repone: \R^d \to \R^k$ and $\reptwo: \R^d \to \R^l$ for some $k,l \in \N$, $d_{\lambda,K}^{\emph{\metricstname}}(\repone,\reptwo) \geq 0$,
        \item (Symmetric) For any two functions $\repone: \R^d \to \R^k$ and $\reptwo: \R^d \to \R^l$ for some $k,l \in \N$, $d_{\lambda,K}^{\emph{\metricstname}}(\repone,\reptwo) = d_{\lambda,K}^{\metricstname}(\reptwo,\repone)$,
        \item (Triangle inequality) For any three functions $\repone: \R^d \to \R^k$, $\reptwo: \R^d \to \R^l$ and $\repthree: \R^d \to \R^m$ for some $k,l,m \in \N$, $d_{\lambda,K}^{\emph{\metricstname}}(\repone,\reptwo) \leq d_{\lambda,K}^{\emph{\metricstname}}(\repone,\repthree) + d_{\lambda,K}^{\emph{\metricstname}}(\repthree,\reptwo)$.
    \end{enumerate}
    Hence, $d_{\lambda,K}^{\emph{\metricstname}}$ is a pseudometric over the space of all functions that maps $\R^d$ to some Euclidean space $\R^t$ for any $t \in \N$.
\end{theorem}

The proof is provided in Section \ref{Proof of Theorem 1} of the Appendix. We now analyze the invariance properties of the pseudometric $d_{\lambda,K}^{\text{\metricstname}}$ and identify the transformations of the representations $\repone$ and $\reptwo$ that leave its value unchanged. To this end, the following lemma will be useful, whose proof is provided in Section \ref{Proof Lemma 2} of the Appendix.

\begin{lemma}\label{Invariance of regularized kernel inner product}

Let $f: \R^d \to \R^k$ and $g: \R^d \to \R^l$ be any two functions. Consider a positive definite, symmetric, bounded and continuous kernel function $K(\cdot,\cdot)$ defined on the domain $\cup_{d} \left\{\mathcal{X}_{d} \times \mathcal{X}_{d}\right\}$, where $\mathcal{X}_{d} \subset \mathbb{R}^{d}$ is a separable space for $d \in \mathbb{N}$. Let $K_{f}(\cdot,\cdot) \coloneq K(f\cdot),f(\cdot))$ and $K_{g}(\cdot,\cdot) \coloneq K(g\cdot),g(\cdot))$ be the unique reproducing kernels corresponding to the ``pullback" RKHS's $\Hf \coloneq \Hil\left(K \circ \left(f \times f\right)\right)$ and $\Hg \coloneq \Hil\left(K \circ \left(g \times g\right)\right)$. For any $\lambda>0$, let $\Sigma_{f}^{-\frac{\lambda}{2}}$ and $\Sigma_{g}^{-\frac{\lambda}{2}}$ denote the square roots of the $\lambda$-regularized covariance operators corresponding to the kernels $K_{f}$ and $K_{g}$, respectively. For any $x,x^{\prime} \in \mathbb{R}^{d}$ and $\lambda>0$, define the operator $\mathcal{I}$ as follows: 
\[
\begin{aligned}
\mathcal{I}(f)(x,x^{\prime}) &= \inprod{\Sigma_{f}^{-\frac{\lambda}{2}}K_{f}(\cdot,x),\Sigma_{f}^{-\frac{\lambda}{2}}K_{f}(\cdot,x^{\prime})}_{\Hf} \\
& = \inprod{K_{f}(\cdot,x),\Sigma_{f}^{-\lambda}K_{f}(\cdot,x^{\prime})}_{\Hf}. 
\end{aligned}
\]
Then, a necessary and sufficient condition for $f$ and $g$ to satisfy $ \mathcal{I}(f) = \mathcal{I}(g)$ is that $ K_{f}(\cdot, \cdot) = K_{g}(\cdot, \cdot)$. 
\end{lemma}

As an easy corollary of Lemma \ref{Invariance of regularized kernel inner product}, we can identify representations that \metricstname treats as equivalent in terms of prediction-based performance for a general collection of kernel ridge regression tasks corresponding to a particular kernel $K$.

\begin{corollary}
\label{corollary 1}
Let $\mathcal{H}$ be the class of transformations under which the kernel $K$ is invariant, i.e., $\mathcal{H} = \left\{h : K(\cdot,\cdot) = K(h(\cdot),h(\cdot))\textrm{ a.e. } P_{X}\right\}$. Then, the \emph{\metricstname} distance $d_{\lambda,K}^{\emph{\metricstname}}(\repone,\reptwo)$ between representations $\repone(X)$ and $\reptwo(X)$ is invariant under the same class of transformations that the kernel $K$ is invariant for, i.e., for any $h_{1},h_{2} \in \mathcal{H}$, 
\[
d_{\lambda,K}^{\emph{\metricstname}}(h_{1} \circ \repone,h_{2} \circ \reptwo) = d_{\lambda,K}^{\emph{\metricstname}}(\repone,\reptwo)
\]
and if either $h_{1}$ or $h_{2}$ does not belong to $\mathcal{H}$, \[
d_{\lambda,K}^{\emph{\metricstname}}(h_{1} \circ \repone,h_{2} \circ \reptwo) \neq d_{\lambda,K}^{\emph{\metricstname}}(\repone,\reptwo).
\]
\end{corollary}

The proof of Corollary \ref{corollary 1} is provided in Section  \ref{Proof of Corollary 1} of the Appendix. Based on these results, the following corollary of Lemma \ref{Invariance of regularized kernel inner product} then provides an exact characterization of the representations that lead to $d_{\lambda,K}^{\text{\metricstname}}=0$.

\begin{corollary}\label{corollary 2}
    A necessary and sufficient condition for the \emph{\metricstname} distance $d_{\lambda,K}^{\emph{\metricstname}}(\repone,\reptwo)$ between representations $\repone(X)$ and $\reptwo(X)$ to be zero is that $K_{\repone}(\cdot,\cdot) = K_{\reptwo}(\cdot,\cdot)$ a.e. $P_{X}$.
\end{corollary}

The proof is straightforward, similar to that of Corollary \ref{corollary 1}, and is therefore omitted.

\section{Statistical estimation of $d_{\lambda,K}^{\text{\metricstname}}$} \label{statistical estimation of UKP}

In practice, when comparing the prediction-based utility of different representations, we consider the realistic scenario where one only has access to a random sample $X_{1},\dots, X_{n} \overset{i.i.d}{\sim}P_{X}$ and a statistical estimator of the proposed distance measure is required. In supervised learning settings, the goal is to allocate most of the data for training and model fitting while minimizing the amount of data used for diagnostics and exploratory analysis.

Using the empirical covariance and cross-covariance operators $\hat{\Sigma}_{\repone}$, $\hat{\Sigma}_{\reptwo}$, $\hat{\Sigma}_{\repone\reptwo}$ and $\hat{\Sigma}_{\reptwo\repone} = \hat{\Sigma}_{\repone\reptwo}^{*}$ as plug-in estimators of $\Sigma_{\repone}$, $\Sigma_{\reptwo}$, $\Sigma_{\repone\reptwo}$ and $\Sigma_{\reptwo\repone}$ in the trace operator based expression of $d_{\lambda,K}^{\text{\metricstname}}(\repone,\reptwo)$ as derived in Proposition \ref{Proposition: Squared pseudometric using trace operators}, we arrive at the following V-statistic type estimator of $d_{\lambda,K}^{\text{\metricstname}}(\repone,\reptwo)$:
\begin{equation}\label{Pseudometric V-statistic estimator}
    \begin{aligned}
        \hat{d}_{\lambda,K}^{\text{\metricstname}}(\repone,\reptwo)
        = \left[\Tr\left(\hat{\Sigma}_{\repone}^{-\lambda}\hat{\Sigma}_{\repone}\hat{\Sigma}_{\repone}^{-\lambda}\hat{\Sigma}_{\repone}\right) + \Tr\left(\hat{\Sigma}_{\reptwo}^{-\lambda}\hat{\Sigma}_{\reptwo}\hat{\Sigma}_{\reptwo}^{-\lambda}\hat{\Sigma}_{\reptwo}\right)-2\Tr\left(\hat{\Sigma}_{\repone}^{-\lambda}\hat{\Sigma}_{\repone\reptwo}\hat{\Sigma}_{\reptwo}^{-\lambda}\hat{\Sigma}_{\reptwo\repone}\right)\right]^{\frac{1}{2}},
    \end{aligned}
\end{equation}
where
\[
\begin{aligned}
    \hat{\Sigma}_{\repone}=\frac{1}{n}\sum_{i=1}^{n}K_{\repone}(\cdot,X_{i}) \otimes_{\Hone}  K_{\repone}(\cdot,X_{i})
    =\frac{1}{n}\sum_{i=1}^{n}K(\repone(\cdot),\repone(X_{i})) \otimes_{\Hone} K(\repone(\cdot),\repone(X_{i})),
\end{aligned}
\]
\[
\begin{aligned}
    \hat{\Sigma}_{\reptwo}=\frac{1}{n}\sum_{i=1}^{n}K_{\reptwo}(\cdot,X_{i}) \otimes_{\Htwo}  K_{\reptwo}(\cdot,X_{i})
    =\frac{1}{n}\sum_{i=1}^{n}K(\reptwo(\cdot),\reptwo(X_{i})) \otimes_{\Htwo} K(\reptwo(\cdot),\reptwo(X_{i})),
\end{aligned}\]
\[
\begin{aligned}
    \hat{\Sigma}_{\repone\reptwo}=\frac{1}{n}\sum_{i=1}^{n}K_{\repone}(\cdot,X_{i}) \otimes_{\HS(\Htwo,\Hone)} K_{\reptwo}(\cdot,X_{i})
    =\frac{1}{n}\sum_{i=1}^{n}K(\repone(\cdot),\repone(X_{i})) \otimes_{\HS(\Htwo,\Hone)} K(\reptwo(\cdot),\reptwo(X_{i})),
\end{aligned}\] and 
\[\begin{aligned}
    \hat{\Sigma}_{\reptwo\repone} &=\frac{1}{n}\sum_{i=1}^{n} K_{\reptwo}(\cdot,X_{i}) \otimes_{\HS(\Hone,\Htwo)}  K_{\repone}(\cdot,X_{i})
    =\frac{1}{n}\sum_{i=1}^{n}K(\reptwo(\cdot),\reptwo(X_{i}))\otimes_{\HS(\Hone,\Htwo)}  K(\repone(\cdot),\repone(X_{i}))\\
    &= \hat{\Sigma}_{\repone\reptwo}^{*}.
\end{aligned}\]

It is an easy exercise to show that the V-statistic type estimator $\hat{d}_{\lambda,K}^{\text{\metricstname}}(\repone,\reptwo)$ can be expressed in terms of the number of input data points $n$, the chosen regularization parameter $\lambda$ and the empirical Gram matrices $K_{n,\repone}$ and $K_{n,\reptwo}$ whose $(i,j)$-th elements are the kernel evaluations for the $(i,j)$-th input data pair $(X_{i},X_{j})$, i.e., $\left(K_{n,\repone}\right)_{ij} = K(\repone(X_{i}),\repone(X_{j}))$ and $\left(K_{n,\reptwo}\right)_{ij} = K(\reptwo(X_{i}),\reptwo(X_{j}))$. If $\lambda=0$, one is required to ensure the invertibility of $K_{n,\repone}$ and $K_{n,\reptwo}$.

\begin{proposition}\label{Proposition: Estimator in terms of Gram matrices}
    For any $\lambda>0$, the V-statistic type estimator $\hat{d}_{\lambda,K}^{\emph{\metricstname}}(\repone,\reptwo)$ of  $d_{\lambda,K}^{\emph{\metricstname}}(\repone,\reptwo)$ between representations $\repone(X)$ and $\reptwo(X)$ can be expressed as
    \[
    \begin{aligned}
        &\hat{d}_{\lambda,K}^{\emph{\metricstname}}(\repone,\reptwo)\\
        =& \left[\emph{Tr}\left(K_{n,\repone}(K_{n,\repone}+n\lambda I)^{-1}K_{n,\repone}(K_{n,\repone}+n\lambda I)^{-1}\right) + \emph{Tr}\left(K_{n,\reptwo}(K_{n,\reptwo}+n\lambda I)^{-1}K_{n,\reptwo}(K_{n,\reptwo}+n\lambda I)^{-1}\right)\right.\\
        &-2\left.\emph{Tr}\left(K_{n,\repone}(K_{n,\repone}+n\lambda I)^{-1}K_{n,\reptwo}(K_{n,\reptwo}+n\lambda I)^{-1}\right)\right]^{\frac{1}{2}}.
    \end{aligned}
    \]
\end{proposition}
 
\subsection{Relation to other comparison measures} 
\label{Relation to other comparison measures}

In this subsection, we discuss the relationship between the \metricstname distance and some popular distances between representations that are popularly used in Machine Learning.

The \metricstname distance is a generalization of the GULP distance, as proposed in \citet{GULP}, in the sense that, if we choose the kernel for the \metricstname to be the linear kernel $K_{lin}(x,y) = x^{T}y$, we exactly recover the GULP distance. Our proposed pseudometric $\hat{d}_{\lambda,K}^{\text{\metricstname}}$ provides the additional flexibility of choosing other kernel functions, such as the Gaussian RBF kernel $K_{RBF,h}$ and the Laplace $K_{Lap,h}$, for understanding the relative difference between the generalization performance on different classes of kernel ridge regression-based prediction tasks. 

Let $K_{n,\repone} = U_{\repone} \Lambda_{n,\repone}U_{\repone}^{T}$ and $K_{n,\reptwo} = U_{\reptwo} \Lambda_{n,\reptwo}U_{\reptwo}^{T}$ be the eigenvalue decompositions of $K_{n,\repone}$ and 
$K_{n,\reptwo}$, respectively. Here $\Lambda_{n,\repone} = \operatorname{diag}\left\{\mu_{\repone}^{(1)},\dots,\mu_{\repone}^{(n)}\right\}$ and $\Lambda_{n,\reptwo} = \operatorname{diag}\left\{\mu_{\reptwo}^{(1)},\dots,\mu_{\reptwo}^{(n)}\right\}$. Define $c_{\repone,\reptwo}^{(i),(j)} = \left(u_{\repone}^{(i)}\right)^{T}u_{\reptwo}^{(j)}$, as the inner product between the $i$-th eigenvector $u_{\repone}^{(i)}$ corresponding to the $i$-th eigenvalue $\mu_{\repone}^{(i)}$ of $K_{n,\repone}$ and $j$-th eigenvector $u_{\reptwo}^{(j)}$ corresponding to the $j$-th eigenvalue $\mu_{\reptwo}^{(i)}$  of $K_{n,\reptwo}$. In the following proposition, we express the V-statistic type estimator $\hat{d}_{\lambda}^{\text{\metricstname}}(\repone,\reptwo)$ exclusively in terms of the inner products $c_{\repone,\reptwo}^{(i),(j)}$'s, the regularization parameter $\lambda$ and the eigenvalues $\mu_{\repone}^{(i)}$'s and $\mu_{\reptwo}^{(j)}$'s, which is useful for understanding the effect of changing the regularization parameter $\lambda$ on the estimate and its relation to other popular pseudometrics on the space of representations.

\begin{proposition}\label{Proposition: Estimator in terms of eigenvalues and eigenvectors of Gram matrices}
    For any $\lambda>0$, the V-statistic type estimator $\hat{d}_{\lambda,K}^{\emph{\metricstname}}(\repone,\reptwo)$ of $d_{\lambda,K}^{\emph{\metricstname}}(\repone,\reptwo)$ between representations $\repone(X)$ and $\reptwo(X)$ can be expressed as
    \[
    \begin{aligned}
        &\hat{d}_{\lambda,K}^{\emph{\metricstname}}(\repone,\reptwo)\\
        =& \left[\sum_{i=1}^{n} \left(\frac{\mu_{\repone}^{(i)}}{\mu_{\repone}^{(i)}+ n\lambda}\right)^{2} + \sum_{j=1}^{n} \left(\frac{\mu_{\reptwo}^{(j)}}{\mu_{\reptwo}^{(i)}+ n\lambda}\right)^{2}\right.-2\left.\sum_{i=1}^{n}\sum_{j=1}^{n}\frac{\mu_{\repone}^{(i)}\mu_{\reptwo}^{(j)}}{\left(\mu_{\repone}^{(i)} + n\lambda\right)\left(\mu_{\reptwo}^{(j)} + n\lambda\right)}\left(c_{\repone,\reptwo}^{(i),(j)} \right)^{2}\right]^{\frac{1}{2}}.
    \end{aligned}
    \]
\end{proposition}

The proof is straightforward, relying on the spectral decomposition of $K_{n,\repone}$ and $K_{n,\reptwo}$ and the properties of the trace operator, and is thus omitted.

The general kernelized version of the Ridge-CCA (Canonical Correlation Analysis) distance, introduced by \citet{vinod1976canonical} and later discussed in \citet{kuss2003geometry}, is defined as
\[
\begin{aligned}
 \hat{d}^{\text{RCCA}}_{\lambda,K}(\repone,\reptwo) &= \Tr\left(\hat{\Sigma}_{\repone}^{-\lambda}\hat{\Sigma}_{\repone\reptwo}\hat{\Sigma}_{\reptwo}^{-\lambda}\hat{\Sigma}_{\reptwo\repone}\right) \\
    &= \sum_{i=1}^{n}\sum_{j=1}^{n}\frac{\mu_{\repone}^{(i)}\mu_{\reptwo}^{(j)}}{\left(\mu_{\repone}^{(i)} + n\lambda\right)\left(\mu_{\reptwo}^{(j)} + n\lambda\right)}\left(c_{\repone,\reptwo}^{(i),(j)} \right)^{2}.
\end{aligned}
\]
However, the machine learning literature has largely focused on the original Ridge-CCA formulation with a linear kernel, as discussed in \citet{kornblith2019similarity}. The classical CCA distance $\hat{d}^{\text{CCA}}$ can be derived from the kernelized Ridge-CCA distance $\hat{d}^{\text{RCCA}}_{\lambda,K}$ by selecting a linear kernel and setting $\lambda=0$. From these definitions, it is clear that \metricstname is a distance measure on the Hilbert space of representations, while the kernelized Ridge-CCA serves as the corresponding inner product on the Hilbert space when the kernel and regularization parameter $\lambda$ are the same for both.

Another related notion of distance, as proposed in \citet{cristianini2001kernel} and popularized  by \citet{kornblith2019similarity}, is known as CKA (Centered Kernel Alignment) and is defined as 
\[
\begin{aligned}
    \hat{d}_{K}^{\text{CKA}}(\repone,\reptwo) = \frac{\Tr\left(K_{n,\repone}H_{n}K_{n,\reptwo}H_{n}\right)}{\sqrt{\Tr\left(K_{n,\repone}H_{n}K_{n,\repone}H_{n}\right) \Tr\left(K_{n,\reptwo}H_{n}K_{n,\reptwo}H_{n}\right)}}
\end{aligned}
\]
where $H_{n} = I_{n} - \frac{1}{n}1_{n}1_{n}^{T}$. We can equivalently express $\hat{d}_{K}^{\text{CKA}}(\repone,\reptwo) $ as 
\[
\begin{aligned}
    \hat{d}^{\text{CKA}}(\repone,\reptwo) = \frac{\sum_{i=1}^{n}\sum_{j=1}^{n}\mu_{\repone}^{(i)}\mu_{\reptwo}^{(j)}\left(c_{\repone,\reptwo}^{(i),(j)} \right)^{2}}{\sqrt{\sum_{i=1}^{n} \left(\mu_{\repone}^{(i)}\right)^{2}}\sqrt{\sum_{j=1}^{n} \left(\mu_{\reptwo}^{(j)}\right)^{2}}}.
\end{aligned}
\]

If the kernelized Ridge-CCA distance is normalized by dividing it by the product of the norms of the pair of representations, taking the regularization parameter $\lambda$ to $+\infty$ recovers the CKA measure $\hat{d}_{K}^{\text{CKA}}(\repone,\reptwo)$ in the limit. This can be shown by expressing $\hat{d}_{\lambda,K}^{\text{\metricstname}}(\repone,\reptwo)$ and $\hat{d}_{K}^{\text{CKA}}(\repone,\reptwo)$ in terms of the eigenvalues and eigenvectors of the empirical Gram matrices $K_{n,\repone}$ and $K_{n,\reptwo}$ and then taking the limit as $\lambda \to +\infty$. The kernelized Ridge-CCA distance thus serves as a bridge between the CKA measure, interpreted as a normalized inner product, and the \metricstname distance, understood as an unnormalized pseudometric in the space of representations. This connection implies a linear correlation between the two measures for sufficiently high value of the regularization parameter. While the CKA and kernelized Ridge-CCA measures naturally reflect similarity between representations via inner products, the \metricstname distance offers a broader perspective. Beyond functioning as a distance on the space of representations, it provides a relative measure of generalization performance uniformly across a wide range of prediction tasks involving kernel ridge regression—something other comparison measures fail to deliver.

It is desirable for discrepancy measures to satisfy pseudometric properties, particularly when comparing representations or features learned by DNN models. The UKP metric enables the assessment of similarity in generalization performance between two representations, even if they were not directly compared during experiments. This is especially useful when a sequence of proposed models is compared to a baseline but not to each other. For instance, suppose $\phi_{1}$ represents a baseline model's representation. If one experimenter uses the UKP metric to compare $\phi_{1}$ with a second representation $\phi_{2}$, while another experimenter compares 
$\phi_{1}$ with a third representation $\phi_{3}$, the triangle inequality provides an upper bound for the UKP distance between $\phi_{2}$ and $\phi_{3}$, even without directly comparing them. This eliminates the need for additional experiments, a valuable feature in the context of deep learning and large-scale data. In contrast, CKA cannot reuse such pairwise comparisons to approximate the similarity between $\phi_{2}$ and $\phi_{3}$. 

Most importantly, the \metricstname distance can differentiate between the generalization ability of models based on their associated representations/features without requiring any ``training'' on particular prediction-based tasks, which makes it efficient in terms of data and computational requirements.

\subsection{Finite sample convergence rate of $\hat{d}_{\lambda,K}^{\text{\metricstname}}$} \label{Finite sample convergence rate}

From a statistical estimation viewpoint, it is possible that the estimator $\hat{d}_{\lambda,K}^{\text{\metricstname}}$ converges to $d_{\lambda,K}^{\text{\metricstname}}$ as the number of data samples $X_{1},\dots,X_{n}$ from the input domain grows to infinity. In addition, we also provide a rate of convergence of the order of $O(\frac{1}{\sqrt{n}})$, which is a parametric rate of convergence. The following theorem, proved in Section \ref{Proof of Theorem 2} of the Appendix, combines these two results and consequently illustrates the finite sample concentration of the  estimator proposed in Equation \eqref{Pseudometric V-statistic estimator} around the population $d_{\lambda,K}^{\text{\metricstname}}$.

\begin{theorem} \label{Theorem: Finite sample convergence}
    Let $\kappa$ be an upper bound on the kernel function $K(\cdot,\cdot)$. Then, for any $\lambda>0$ and $\delta>0$, with probability atleast $1-\delta$, the V-statistic estimator $\hat{d}_{\lambda}^{\emph{\metricstname}}(\repone,\reptwo)$ satisfies
    \[
    \begin{aligned}
    \left|\left(d_{\lambda,K}^{\emph{\metricstname}}(\repone,\reptwo)\right)^{2}-\left(\hat{d}_{\lambda}^{\emph{\metricstname}}(\repone,\reptwo)\right)^{2}\right|\leq \frac{8\kappa^{3}}{\lambda^{3}}\left[\frac{2\log(\frac{6}{\delta})}{n} + \sqrt{\frac{2\log(\frac{6}{\delta})}{n}}\right]+ \frac{4\kappa^{2}}{\lambda^{2}}\left[\frac{2}{n} + \sqrt{\frac{2\log(\frac{6}{\delta})}{n}}\right] .
    \end{aligned}
    \]
\end{theorem}

\subsection{Computational complexity of $\hat{d}_{\lambda,K}^{\text{\metricstname}}$}

From the expression of the estimator $\hat{d}_{\lambda,K}^{\text{\metricstname}}$ in Proposition \ref{Proposition: Estimator in terms of Gram matrices}, it can be shown that its computational complexity is $O(n^3)$, where $n$ is the sample size. Notably, the GULP distance proposed in \citet{GULP} shares the same complexity. The primary computational cost arises from inverting the Gram matrix, which can be reduced using kernel approximation techniques like Random Fourier Features (RFF) or Nystr\"{o}m approximation. For example, by using $D$ RFF samples from the spectral distribution of the kernel $K$ or $D$ subsamples from the $n$ data samples in the Nystr\"{o}m method, the complexity of the UKP distance estimator $\hat{d}_{\lambda,K}^{\text{\metricstname}}(\phi,\psi)$ can be reduced from $O(n^3)$ to $O(nD^{2} + D^{3})$, which is significantly lower than $O(n^3)$ when $D \ll n$. Exploring the tradeoff between the statistical accuracy of \metricstname distance estimation and the computational efficiency of kernel approximation methods is a promising direction for future research.

\section{Experiments} \label{experiments}

 In this section, we present experimental results that showcase the efficacy of the \metricstname distance in identifying similarities and differences between representations relevant to generalization performance on prediction tasks. Additional experiments, including model architecture details and training, are provided in the Appendix. All computations were performed on a single A100 GPU using Google Colab.

\subsection{Ability of \metricstname to predict generalization performance by kernel ridge regression-based predictors} \label{MNIST experiments}
The \metricstname pseudometric gives a uniform bound on the difference in predictions generated by a pair of models, based on kernel ridge regression-based estimators that utilize the respective representations of the two models. It is a natural question to ask if this uniform or worst-case guarantee on the difference in prediction performance between representations is useful on a per-instance basis, i.e., given a specific kernel ridge regression task, whether the \metricstname distance is positively correlated with the generalization performance of different models.

We consider 50 fully-connected neural networks with ReLU activation, each having uniform widths of 200, 400, 700, 800, or 900 and depths ranging from 1 to 10. These networks are trained on 60,000 $28 \times 28$-pixel training images from the MNIST handwritten digits dataset \cite{deng2012mnist} for 50 epochs. Representations are then extracted from the penultimate (final hidden) layer of each network, and the CCA, linear CKA (CKA with a linear kernel), GULP, and UKP distances are estimated for each pair of representations using 5,000 test images from the same dataset.

\begin{figure}[h]
\begin{center}
\includegraphics[width=10cm]{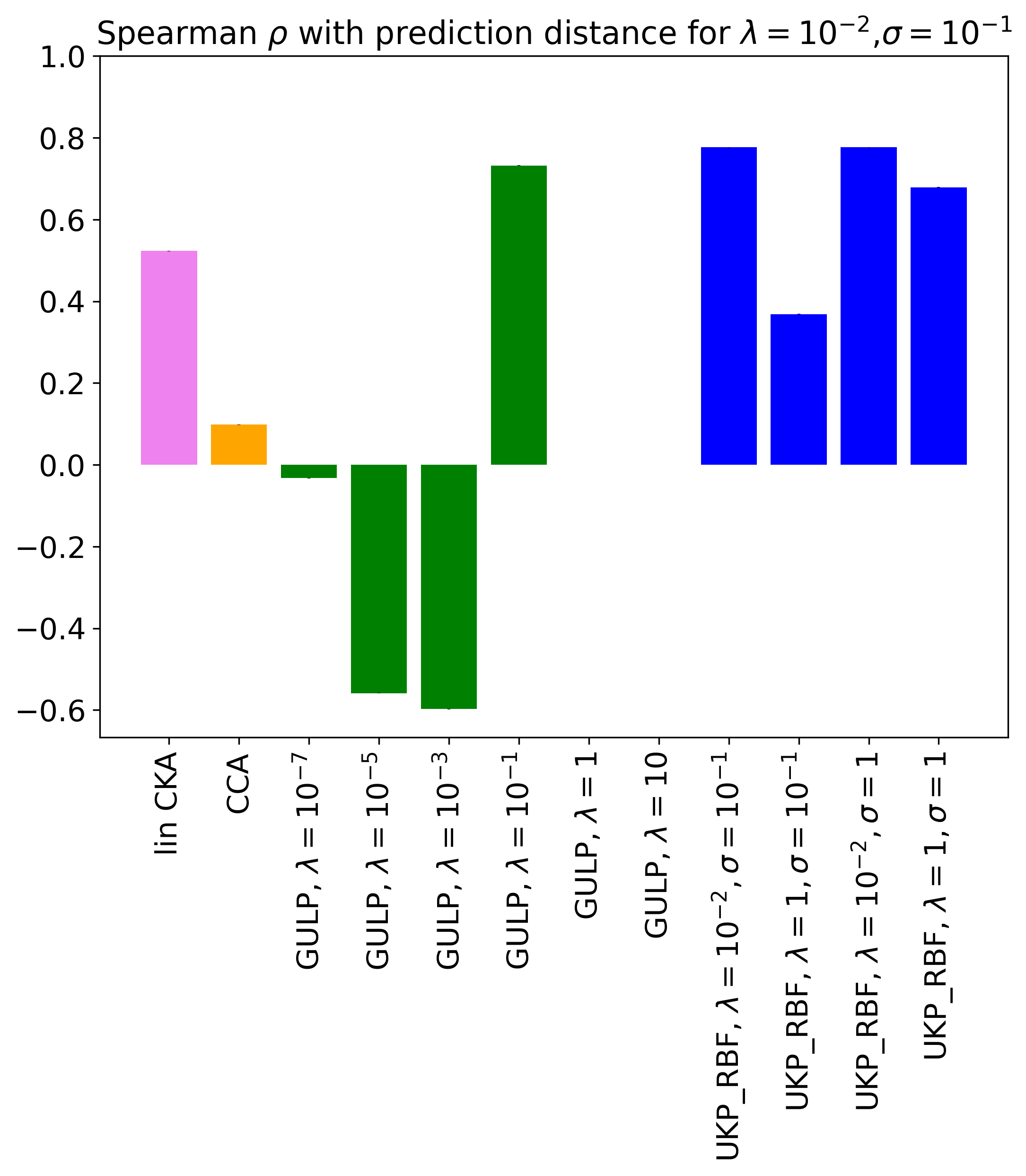}
\caption{Generalization of kernel ridge regression-based predictors is strongly positively correlated with \metricstname distance values. We report the average correlation across 10 random synthetic kernel ridge regression tasks. Error bars are negligibly small and hence not visible.}
\label{Generalization plot}
\end{center}
\vspace{-4mm}
\end{figure}

We create synthetic kernel ridge regression tasks where we randomly sample 5000 images and randomly assign a standard Gaussian label to each image to create the synthetic label/target vector. We obtain the kernel ridge regression estimator for each representation with ridge penalty $\lambda \in \{10^{-2},1\}$ and Gaussian RBF kernel with bandwidth $\sigma \in \{10^{-1}, 1\}$. The empirical mean of the squared difference between predictions based on a pair of representations (say $\repone$ and $\reptwo$) is then computed using 5000 test images to estimate $err_{\repone,\reptwo} = \E_{X \sim P_{X}}\left[\alpha_{\lambda}^{\repone}(X)-\alpha_{\lambda}^{\reptwo}(X)\right]^{2}$, where $\alpha_{\lambda}^{\repone}$ and $\alpha_{\lambda}^{\reptwo}$ are the kernel ridge regression based predictors. 

In Fig. \ref{Generalization plot}, we plot the Spearman's $\rho$ rank correlation coefficient between the $err_{\repone,\reptwo}$'s and the pairwise distances between the representations using CCA, linear CKA, GULP and UKP distances. For this particular regression task, we chose the synthetic ridge penalty to be $\lambda=10^{-2}$ and used a Gaussian RBF kernel with $\sigma=10^{-1}$. For the \metricstname distance, we use the Gaussian RBF kernel as the choice of kernel.

We observe that the pairwise \metricstname distance is highly positively correlated with the collection of $err_{\repone,\reptwo}$'s, as evident from the large positive values of the blue bars, with the largest correlation being observed when the ridge penalty used in the \metricstname distance matches with the synthetic ridge penalty we chose, i.e., $\lambda=10^{-2}$. In contrast, GULP distances exhibit inconsistent behavior across varying levels of regularization, while CCA and linear CKA distances show a significantly weaker positive correlation with generalization performance. As expected, due to the relationship between CKA and \metricstname discussed in Section \ref{Relation to other comparison measures}, the CKA distance with a Gaussian RBF kernel performs comparably to UKP. 
Experiments with the remaining combinations of tuning parameters $\lambda$ and $\sigma$ are presented in Fig. \ref{MNIST generalization plots} in Section \ref{MNIST Experiments additional} of the Appendix, yielding qualitatively similar conclusions.

\begin{figure}[t]
\begin{center}
\includegraphics[width=10cm]{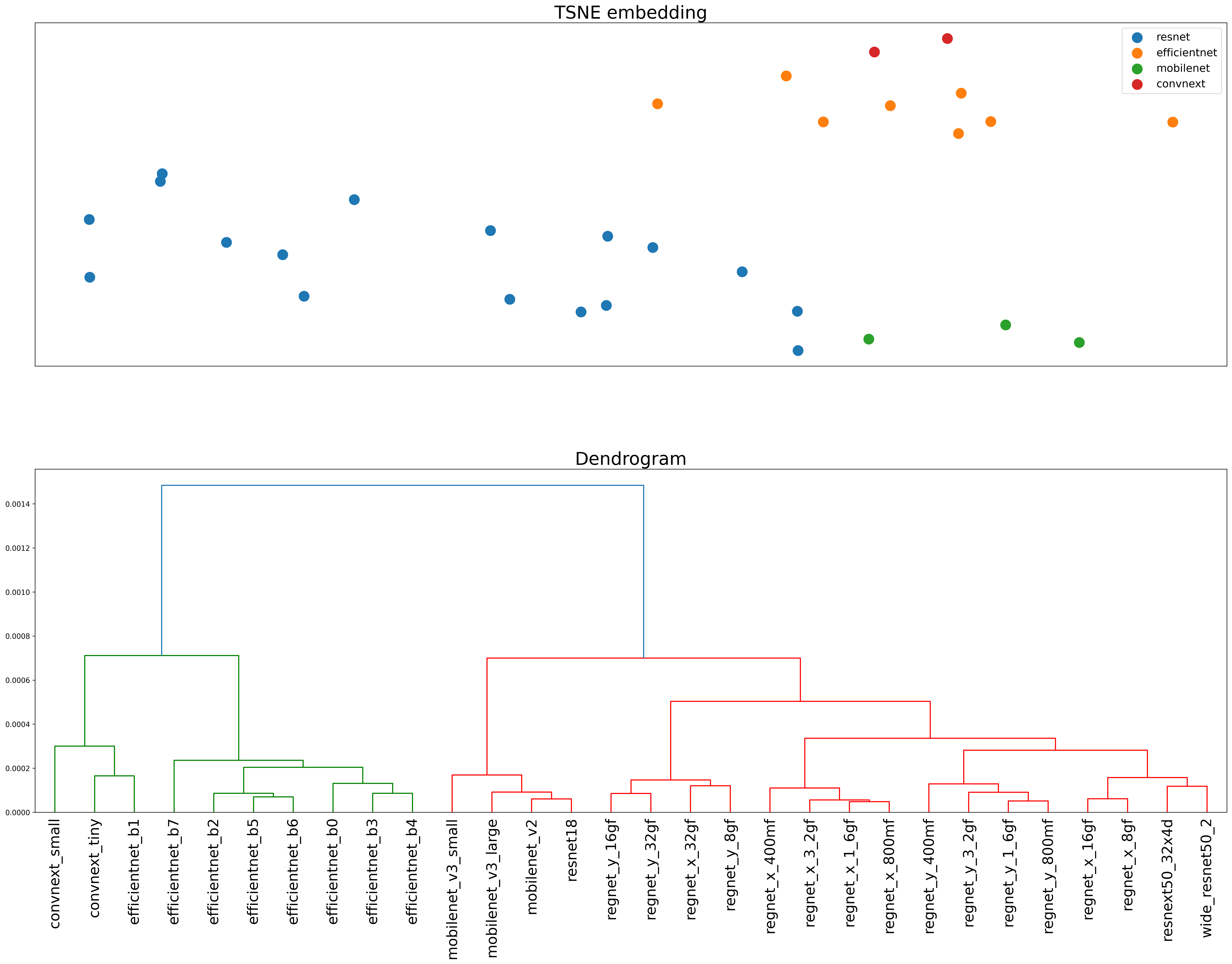}
\caption{Clustering based on \metricstname distance is sensitive to differences in architectures of neural network models.}\label{DendrogramandtSNE}
\end{center}
\vspace{-4mm}
\end{figure}

\subsection{Ability of \metricstname to identify differences in architectures and inductive biases} \label{ImageNet experiments experiments}
 A key source of inductive biases in neural network models is their architecture, with features such as residual connections and variations in convolutional filter complexity shaping the representations learned during training. As a pseudometric over feature space, the \metricstname distance is expected to capture intrinsic differences in these inductive biases, which are known to impact generalization performance across tasks. To explore this, we analyze representations from 35 pre-trained neural network architectures used for image classification, described in detail in Section \ref{Additional ImageNet experiments} of the Appendix.

 We estimate pairwise \metricstname distances between model representations using 3,000 images from the validation set of the ImageNet dataset \cite{krizhevsky2012imagenet}, a regularization parameter $\lambda=1$ and a Gaussian kernel with bandwidth $\sigma=10$. The tSNE embedding method is then used to embed these representations into 2-D space utilizing the distance measures given by the \metricstname pseudometric. Concurrently, we perform an agglomerative (bottom-up) hierarchical clustering of the representations based on the pairwise \metricstname distances and obtain the corresponding dendrogram. We observe in Fig. \ref{DendrogramandtSNE} that similar architectures which share important properties, such as the Regnets and Resnets are clustered together, while they are well separated from smaller efficient architectures such as MobileNets and ConvNexts. This demonstrates that the \metricstname distance effectively captures notions of similarity and dissimilarity aligned with interpretable notions based on inductive biases. Further comparisons with baseline measures, such as GULP and CKA, presented in Fig. \ref{ImageNet dendrograms additional} in Section \ref{Additional ImageNet experiments} of the Appendix demonstrate that \metricstname often provides superior clustering quality. We would like to note here that the choice of the kernel function for the \metricstname pseudometric should be driven by the nature of inductive bias that will be useful for the tasks for which the representations/features of interest will be used. Additional discussion regarding kernel (and kernel parameter) selection is provide in Section \ref{Additional ImageNet experiments} of the Appendix.

\section{Conclusion and future work} \label{conclusion}

This paper introduces the \metricstname pseudometric, a novel method for comparing model representations based on their predictive performance in kernel ridge regression tasks. It is shown to be easily interpretable, efficient, and capable of encoding inductive biases, supported by theoretical proofs and experimental validation. Therefore, the \metricstname pseudometric can serve as an useful and versatile exploratory tool for comparison of model representations, including representations learnt by black-box models such as neural networks, deep learning models and Large Language Models (LLMs). Future research could focus on using \metricstname for model selection, hyperparameter tuning, and enhancing its computational efficiency for large-scale models, such as deep neural networks, to better suit real-world applications.

\section*{References}
\bibliographystyle{plainnat}
\bibliography{main.bib}
\nocite{*}

\newpage

\appendix


\section{Proofs}\label{Proofs}
In this appendix, we present the missing proofs of the paper.
\subsection{Proof of Lemma \ref{Characterization of pseudometric in terms of expected squared distance}}\label{Proof of Lemma 1}
\begin{proof}
    Consider a fixed population regression function $\eta(x) = \E(Y \mid X = x)$ corresponding to a fixed joint distribution of $(X, Y)$. Note that, for any $f \in \Hone$, we have
    \[
    \begin{aligned}
        &\E\left[Y-f(X)\right]^{2} + \lambda \norm{f}_{\Hone}^{2}
        = \E\left[Y-\inprod{f,K_{\repone}(\cdot,X)}_{\Hone}\right]^{2} + \lambda \norm{f}_{\Hone}^{2}\\
        =&\E(Y^{2}) - 2\E\left[Y\inprod{f,K_{\repone}(\cdot,X)}_{\Hone}\right] + \E\left[\inprod{f,K_{\repone}(\cdot,X)}_{\Hone}^{2}\right] + \lambda \norm{f}_{\Hone}^{2}\\
        =&\E(Y^{2}) - 2\E\left[\eta(X)\inprod{f,K_{\repone}(\cdot,X)}_{\Hone}\right] + \E\inprod{f,\left[K_{\repone}(\cdot,X) \otimes_{\Hone}K_{\repone}(\cdot,X)\right] f}_{\HS(\Hone)} + \lambda \inprod{f,f}_{\Hone}\\
        =&\E(Y^{2}) -2\inprod{f,\I_{\repone}^{*}\eta}_{\Hone} + \inprod{f,(\Sigma_{\repone}+\lambda I)f}_{\Hone}\\
        =&\E(Y^{2}) + \norm{\left(\Sigma_{\repone}+\lambda I\right)^{\frac{1}{2}}f - \left(\Sigma_{\repone}+\lambda I\right)^{-\frac{1}{2}}\I_{\repone}^{*}\eta}_{\Hone}^{2} - \norm{\left(\Sigma_{\repone}+\lambda I\right)^{-\frac{1}{2}}\I_{\repone}^{*}\eta}_{\Hone}^{2}.
    \end{aligned}
    \]

Therefore, the kernel ridge regression estimator of $\eta$ using the representation $\phi(X)$ is given by 
\[
\begin{aligned}
    \alpha_{\lambda} = & \underset{\alpha \in \Hone}{\argmin} \hspace{2pt} \E\left[Y-\alpha(X)\right]^{2} + \lambda \norm{\alpha}_{\Hone}^{2}
    = \Sigma_{\repone}^{-\lambda}\I_{\repone}^{*}\eta.
\end{aligned}
\]
Similarly, we can show that
\[
\begin{aligned}
    \beta_{\lambda} = & \underset{\beta \in \Htwo}{\argmin} \hspace{2pt} \E\left[Y-\beta(X)\right]^{2} + \lambda \norm{\beta}_{\Htwo}^{2}
    = \Sigma_{\reptwo}^{-\lambda}\I_{\reptwo}^{*}\eta.
\end{aligned}
\]

Now, 
\[
\begin{aligned}
    &\alpha_{\lambda}(x^{\prime}) = \int \eta(x) \left[\Sigma_{\repone}^{-\lambda}K_{\repone}(\cdot,x)\right](x^{\prime})dP_{X}(x)
    = \int \eta(x) \inprod{\Sigma_{\repone}^{-\lambda}K_{\repone}(\cdot,x),K_{\repone}(\cdot,x^{\prime})}_{\Hone}dP_{X}(x)\\
    = &\int \eta(x) \inprod{\Sigma_{\repone}^{-\frac{\lambda}{2}}K_{\repone}(\cdot,x),\Sigma_{\repone}^{-\frac{\lambda}{2}}K_{\repone}(\cdot,x^{\prime})}_{\Hone}dP_{X}(x)\\
    = & \inprod{\eta,\inprod{\Sigma_{\repone}^{-\frac{\lambda}{2}}K_{\repone}(\cdot,x),\Sigma_{\repone}^{-\frac{\lambda}{2}}K_{\repone}(\cdot,x^{\prime})}_{\Hone}}_{\LPtwo}\\
    = & \inprod{\eta,\inprod{\widetilde{K}_{\repone}(\cdot,x),\widetilde{K}_{\repone}(\cdot,x^{\prime})}_{\Hone}}_{\LPtwo}
     = \inprod{\eta,\inprod{\widetilde{K}_{\repone}(\cdot,x),\widetilde{K}_{\repone}(\cdot,x^{\prime})}_{\Hil}}_{\LPtwo}.
\end{aligned}
\]
Similarly,
\[
\begin{aligned}
    \beta_{\lambda}(x^{\prime}) = & \inprod{\eta,\inprod{\widetilde{K}_{\reptwo}(\cdot,x),\widetilde{K}_{\reptwo}(\cdot,x^{\prime})}_{\Htwo}}_{\LPtwo}
     =  \inprod{\eta,\inprod{\widetilde{K}_{\reptwo}(\cdot,x),\widetilde{K}_{\reptwo}(\cdot,x^{\prime})}_{\Hil}}_{\LPtwo}.
\end{aligned}
\]

Therefore, we have that
\[
\begin{aligned}
&\left[d_{\lambda,K}^{\text{\metricstname}}(\repone,\reptwo)\right]^{2}
    = \underset{\norm{\eta}_{\LPtwo} \leq 1}{\sup} \E\left[\alpha_{\lambda}(X)-\beta_{\lambda}(X)\right]^{2}\\
    =& \underset{\norm{\eta}_{\LPtwo} \leq 1}{\sup} \E\left\langle\eta, \inprod{\widetilde{K}_{\repone}(\cdot,\cdot),\widetilde{K}_{\repone}(\cdot,X)}_{\Hone} \right.\left.-\inprod{\widetilde{K}_{\reptwo}(\cdot,\cdot),\widetilde{K}_{\reptwo}(\cdot,X)}_{\Htwo}\right\rangle_{\LPtwo}^{2}
    \end{aligned}
    \]
    \[
    \begin{aligned}
    =& \E \left\|\inprod{\widetilde{K}_{\repone}(\cdot,\cdot),\widetilde{K}_{\repone}(\cdot,X)}_{\Hone} \right.\left.-\inprod{\widetilde{K}_{\reptwo}(\cdot,\cdot),\widetilde{K}_{\reptwo}(\cdot,X)}_{\Htwo}\right\|_{\LPtwo}^{2}\\
    =& \E  \left[\inprod{\Sigma_{\repone}^{-\frac{\lambda}{2}}K_{\repone}(\cdot,X),\Sigma_{\repone}^{-\frac{\lambda}{2}}K_{\repone}(\cdot,X^{\prime})}_{\Hone} \right.  \left.-\inprod{\Sigma_{\reptwo}^{-\frac{\lambda}{2}}K_{\reptwo}(\cdot,X),\Sigma_{\reptwo}^{-\frac{\lambda}{2}}K_{\reptwo}(\cdot,X^{\prime})}_{\Htwo}\right]^{2}\\
   = & \E  \left[\inprod{K_{\repone}(\cdot,X),\Sigma_{\repone}^{-\lambda}K_{\repone}(\cdot,X^{\prime})}_{\Hone} \right.  \left.-\inprod{K_{\reptwo}(\cdot,X),\Sigma_{\reptwo}^{-\lambda}K_{\reptwo}(\cdot,X^{\prime})}_{\Htwo}\right]^{2},
\end{aligned}
\]
where $X$ and $X^{\prime}$ are i.i.d observations drawn from $P_{X}$.
\end{proof}

\subsection{Proof of Proposition \ref{Proposition: Squared pseudometric using trace operators}}\label{Proof of Proposition 1}

\begin{proof}
    Using Lemma \ref{Characterization of pseudometric in terms of expected squared distance}, the squared \metricstname distance of $d_{\lambda,K}^{\text{\metricstname}}(\repone,\reptwo)$ between between representations $\repone(X)$ and $\reptwo(X)$ can be expressed as
    \[
    \begin{aligned}
        &\left[d_{\lambda,K}^{\text{\metricstname}}(\repone,\reptwo)\right]^{2}
        = \E\left[ \inprod{\widetilde{K}_{\repone}(\cdot,X),\widetilde{K}_{\repone}(\cdot,X^{\prime})}_{\Hone} \right.\left.-\inprod{\widetilde{K}_{\reptwo}(\cdot,X),\widetilde{K}_{\reptwo}(\cdot,X^{\prime})}_{\Htwo}\right]^{2}\\
        =& \E\left[\inprod{\widetilde{K}_{\repone}(\cdot,X)\otimes_{\Hone}\widetilde{K}_{\repone}(\cdot,X),\widetilde{K}_{\repone}(\cdot,X^{\prime})\otimes_{\Hone}\widetilde{K}_{\repone}(\cdot,X^{\prime})}_{\HS(\Hone)}\right.\\
        &\left.+\inprod{\widetilde{K}_{\reptwo}(\cdot,X)\otimes_{\Htwo}\widetilde{K}_{\reptwo}(\cdot,X),\widetilde{K}_{\reptwo}(\cdot,X^{\prime})\otimes_{\Htwo}\widetilde{K}_{\reptwo}(\cdot,X^{\prime})}_{\HS(\Htwo)}\right.\\
        &\left.-2\left\langle\widetilde{K}_{\repone}(\cdot,X)\otimes_{\HS(\Htwo,\Hone)}\widetilde{K}_{\reptwo}(\cdot,X),\right.\right.\left.\left.\widetilde{K}_{\repone}(\cdot,X^{\prime})\otimes_{\HS(\Htwo,\Hone)}\widetilde{K}_{\reptwo}(\cdot,X^{\prime})\right\rangle_{\HS(\Htwo,\Hone)}\right]\\
        =&\inprod{\Sigma_{\repone}^{-\frac{\lambda}{2}}\Sigma_{\repone}\Sigma_{\repone}^{-\frac{\lambda}{2}},\Sigma_{\repone}^{-\frac{\lambda}{2}}\Sigma_{\repone}\Sigma_{\repone}^{-\frac{\lambda}{2}}}_{\HS(\Hone)} + \inprod{\Sigma_{\reptwo}^{-\frac{\lambda}{2}}\Sigma_{\reptwo}\Sigma_{\reptwo}^{-\frac{\lambda}{2}},\Sigma_{\reptwo}^{-\frac{\lambda}{2}}\Sigma_{\reptwo}\Sigma_{\reptwo}^{-\frac{\lambda}{2}}}_{\HS(\Htwo)}\\
        &-2\inprod{\Sigma_{\repone}^{-\frac{\lambda}{2}} \Sigma_{\repone\reptwo} \Sigma_{\reptwo}^{-\frac{\lambda}{2}} ,\Sigma_{\repone}^{-\frac{\lambda}{2}} \Sigma_{\repone\reptwo} \Sigma_{\reptwo}^{-\frac{\lambda}{2}}}_{\HS(\Htwo,\Hone)}\\
        =& \Tr\left(\Sigma_{\repone}^{-\lambda}\Sigma_{\repone}\Sigma_{\repone}^{-\lambda}\Sigma_{\repone}\right) + \Tr\left(\Sigma_{\reptwo}^{-\lambda}\Sigma_{\reptwo}\Sigma_{\reptwo}^{-\lambda}\Sigma_{\reptwo}\right)-2\Tr\left(\Sigma_{\repone}^{-\lambda}\Sigma_{\repone\reptwo}\Sigma_{\reptwo}^{-\lambda}\Sigma_{\reptwo\repone}\right)
    \end{aligned}
    \]
    which completes the proof.
\end{proof}

\subsection{Proof of Theorem \ref{Theorem: Pseudometric}}\label{Proof of Theorem 1}

\begin{proof}
    The first three properties immediately follow from the characterization of $d_{\lambda,K}^{\text{\metricstname}}$ given in Lemma \ref{Characterization of pseudometric in terms of expected squared distance}. Note that,
    \[
    \begin{aligned}
        d_{\lambda,K}^{\text{\metricstname}}(\repone,\reptwo)
        =& \left(\E  \left[\inprod{K_{\repone}(\cdot,X),\Sigma_{\repone}^{-\lambda}K_{\repone}(\cdot,X^{\prime})}_{\Hone} \right.\right. \left.\left. -\inprod{K_{\reptwo}(\cdot,X),\Sigma_{\reptwo}^{-\lambda}K_{\reptwo}(\cdot,X^{\prime})}_{\Htwo}\right]^{2}\right)^{\frac{1}{2}}\\
        =& \left(\E  \left[\inprod{K_{\repone}(\cdot,X),\Sigma_{\repone}^{-\lambda}K_{\repone}(\cdot,X^{\prime})}_{\Hone} \right.\right.  -\inprod{K_{\repthree}(\cdot,X),\Sigma_{\repthree}^{-\lambda}K_{\repthree}(\cdot,X^{\prime})}_{\Hthree}\\
        &+\inprod{K_{\repthree}(\cdot,X),\Sigma_{\repthree}^{-\lambda}K_{\repthree}(\cdot,X^{\prime})}_{\Hthree}\left.\left. -\inprod{K_{\reptwo}(\cdot,X),\Sigma_{\reptwo}^{-\lambda}K_{\reptwo}(\cdot,X^{\prime})}_{\Htwo}\right]^{2}\right)^{\frac{1}{2}}
        \end{aligned}
        \]
        \[
        \begin{aligned}
        &\overset{\dagger}{\leq} \left(\E  \left[\inprod{K_{\repone}(\cdot,X),\Sigma_{\repone}^{-\lambda}K_{\repone}(\cdot,X^{\prime})}_{\Hone} \right.\right. \left.\left. -\inprod{K_{\repthree}(\cdot,X),\Sigma_{\repthree}^{-\lambda}K_{\repthree}(\cdot,X^{\prime})}_{\Hthree}\right]^{2}\right)^{\frac{1}{2}}\\
        &+ \left(\E  \left[\inprod{K_{\repthree}(\cdot,X),\Sigma_{\repthree}^{-\lambda}K_{\repthree}(\cdot,X^{\prime})}_{\Hthree} \right.\right. \left.\left. -\inprod{K_{\reptwo}(\cdot,X),\Sigma_{\reptwo}^{-\lambda}K_{\reptwo}(\cdot,X^{\prime})}_{\Htwo}\right]^{2}\right)^{\frac{1}{2}}\\
        &=  d_{\lambda,K}^{\text{\metricstname}}(\repone,\repthree) + d_{\lambda,K}^{\text{\metricstname}}(\repthree,\reptwo),
    \end{aligned}
    \]
    where $\dagger$ follows using Minkowski's inequality for integrals. Thus, the $d_{\lambda,K}^{\text{\metricstname}}$ distance satisfies the triangle inequality along with the other three properties, and consequently, fulfills all the requirements of a pseudometric.
\end{proof}

\subsection{Proof of Lemma \ref{Invariance of regularized kernel inner product}}\label{Proof Lemma 2}

\begin{proof}
    The sufficiency of the condition is obvious, so we proceed to prove the necessity part.

    Under the given conditions on the kernel $K$, the integral operators $\mathcal{T}_{f}$ and $\mathcal{T}_{g}$ corresponding to the kernels $K_{f}$ and $K_{g}$ both admit spectral decompositions. Let $\left(\mu_{i}^{f},e_{i}^{f}\right)_{i=1}^{\infty}$ and $\left(\mu_{j}^{g},e_{j}^{g}\right)_{j=1}^{\infty}$ be the eigenvalue-eigenfunction pairs corresponding to the spectral decomposition of $\mathcal{T}_{f}$ and $\mathcal{T}_{g}$, respectively. Then, we have that
    \begin{equation*}\label{Spectral decomposition of Tf}
        \mathcal{T}_{f} = \sum_{i=1}^{\infty} \mu_{i}^{f} \left(e_{i}^{f} \otimes_{\LPtwo} e_{i}^{f}\right)
    \end{equation*}
    and
    \begin{equation*}\label{Spectral decomposition of Tg}
        \mathcal{T}_{g} = \sum_{j=1}^{\infty} \mu_{j}^{g} \left(e_{j}^{g} \otimes_{\LPtwo} e_{j}^{g}\right).
    \end{equation*}

    Since $K$ is a positive definite, symmetric, continuous and bounded kernel defined on a separable domain, $\mathcal{T}_{f}$ and $\mathcal{T}_{g}$ are compact, self-adjoint, trace-class operators. Therefore, we must have that $\mu_{i}^{f},\mu_{j}^{g} > 0$ and $\lim_{i \to \infty} \mu_{i}^{f} = \lim_{j \to \infty} \mu_{j}^{g} = 0$. Further, $(e_{i}^{f})_{i=1}^{\infty}$ and $(e_{j}^{g})_{j=1}^{\infty}$ constitute orthonormal bases of $\Hf$ and $\Hg$, respectively.

    The Mercer decompositions of the kernels $K_{f}$ and $K_{g}$ are given by,
    \begin{equation*}\label{Mercer decomposition of Kf}
        \mathcal{K}_{f}(x,x^{\prime}) = \sum_{i=1}^{\infty} \mu_{i}^{f} e_{i}^{f}(x) e_{i}^{f}(x^{\prime})
    \end{equation*}
    and 
    \begin{equation*}\label{Mercer decomposition of Kg}
        \mathcal{K}_{g}(x,x^{\prime}) = \sum_{j=1}^{\infty} \mu_{j}^{g} e_{j}^{g}(x) e_{j}^{g}(x^{\prime}).
    \end{equation*}

    Note that,
    \begin{equation}\label{Expressing regularized inner product If in terms of eigenvalues and eigenfunctions}
    \begin{aligned}[b]
    \mathcal{I}(f)(x,x^{\prime}) &= \inprod{\Sigma_{f}^{-\frac{\lambda}{2}}K_{f}(\cdot,x),\Sigma_{f}^{-\frac{\lambda}{2}}K_{f}(\cdot,x^{\prime})}_{\Hf} 
     = \inprod{K_{f}(\cdot,x),\Sigma_{f}^{-\lambda}K_{f}(\cdot,x^{\prime})}_{\Hf}\\
    & = \sum_{i=1}^{\infty} \frac{\mu_{i}^{f}}{\mu_{i}^{f} + \lambda} e_{i}^{f}(x) e_{i}^{f}(x^{\prime}). 
    \end{aligned}
    \end{equation}

    Similarly, we have
    \begin{equation}\label{Expressing regularized inner product Ig in terms of eigenvalues and eigenfunctions}
    \mathcal{I}(g)(x,x^{\prime}) = \sum_{j=1}^{\infty} \frac{\mu_{j}^{g}}{\mu_{j}^{g} + \lambda} e_{j}^{g}(x) e_{j}^{g}(x^{\prime}). 
    \end{equation}

    Define $t_{ij} \coloneq \inprod{e_{i}^{f},e_{j}^{g}}_{\LPtwo}$ for all $i,j$. Further, define $V_{i} = \left\{j \in \mathbb{N} : t_{ij} \neq 0\right\}$ for all $i$ and $W_{j} = \left\{i \in \mathbb{N} : t_{ij} \neq 0\right\}$ for all $j$. Now, using \eqref{Expressing regularized inner product If in terms of eigenvalues and eigenfunctions} and \eqref{Expressing regularized inner product Ig in terms of eigenvalues and eigenfunctions}, we have that
    \begin{equation}\label{Equality of regularized inner products If and Ig part 1}
        \begin{aligned}[b]
            &\mathcal{I}(f) = \mathcal{I}(g)\\
            \iff &\sum_{i=1}^{\infty} \frac{\mu_{i}^{f}}{\mu_{i}^{f} + \lambda} e_{i}^{f}(\cdot) e_{i}^{f}(\cdot) = \sum_{j=1}^{\infty} \frac{\mu_{j}^{g}}{\mu_{j}^{g} + \lambda} e_{j}^{g}(\cdot) e_{j}^{g}(\cdot) .
        \end{aligned}
    \end{equation}

    Taking the $\LPtwo$ inner product of both the RHS and LHS of \eqref{Equality of regularized inner products If and Ig part 1} with $e_{j}^{g}$, we have that
    \begin{equation}\label{Equality of regularized inner products If and Ig part 2}
        \sum_{i=1}^{\infty} \frac{\mu_{i}^{f}t_{ij}}{\mu_{i}^{f} + \lambda} e_{i}^{f}(\cdot) = \frac{\mu_{j}^{g}}{\mu_{j}^{g} + \lambda} e_{j}^{g}(\cdot) .
    \end{equation}

    Taking the $\LPtwo$ inner product of both the RHS and LHS of \eqref{Equality of regularized inner products If and Ig part 2} with $e_{k}^{f}$, we have that
    \begin{equation}\label{Equality of regularized inner products If and Ig part 3}
        \begin{aligned}[b]
            &\frac{\mu_{k}^{f}t_{kj}}{\mu_{k}^{f} + \lambda} = \frac{\mu_{j}^{g}t_{kj}}{\mu_{j}^{g} + \lambda} \\
            \iff & t_{kj} \left(\frac{\mu_{k}^{f}}{\mu_{k}^{f} + \lambda} - \frac{\mu_{j}^{g}}{\mu_{j}^{g} + \lambda}  \right) = 0\\
            \iff & t_{kj} \left( \mu_{k}^{f} - \mu_{j}^{g}\right) = 0 .
        \end{aligned}
    \end{equation}

    Taking the $\LPtwo$ inner product of both the RHS and LHS of \eqref{Equality of regularized inner products If and Ig part 2} with $e_{k}^{g}$, we have that
    \begin{equation}\label{Equality of regularized inner products If and Ig part 4}
        \begin{aligned}[b]
        & \sum_{i=1}^{\infty} \frac{\mu_{i}^{f}t_{ij}t_{ik}}{\mu_{i}^{f} + \lambda} = \begin{cases}
            \frac{\mu_{j}^{g}}{\mu_{j}^{g} + \lambda} \textrm{ if } j = k \\
            0, \textrm{ if } j \neq k
        \end{cases} \\
        \iff & \sum_{i \in W_{j} \cap W_{k}}\frac{\mu_{i}^{f}t_{ij}t_{ik}}{\mu_{i}^{f} + \lambda} = \begin{cases}
            \frac{\mu_{j}^{g}}{\mu_{j}^{g} + \lambda} \textrm{ if } j = k \\
            0, \textrm{ if } j \neq k
        \end{cases}.
        \end{aligned}
    \end{equation}

    Using \eqref{Equality of regularized inner products If and Ig part 3} and \eqref{Equality of regularized inner products If and Ig part 4}, we have that

    \begin{equation}\label{Equality of regularized inner products If and Ig part 5}
        \begin{aligned}[b]
        \frac{\mu_{j}^{g}}{\mu_{j}^{g} + \lambda} \left[ \sum_{i \in W_{j}}t_{ij}^{2} - 1\right] = 0
        \end{aligned}
    \end{equation}
    and, if $j \neq k$,
    \begin{equation}\label{Equality of regularized inner products If and Ig part 6}
        \begin{aligned}[b]
        \frac{\mu_{j}^{g}}{\mu_{j}^{g} + \lambda} \left( \sum_{i \in W_{j}\cap W_{k}}t_{ij}t_{ik}\right) = 0.
        \end{aligned}
    \end{equation}

    Therefore, from \eqref{Equality of regularized inner products If and Ig part 5} and \eqref{Equality of regularized inner products If and Ig part 6}, we obtain that
    \begin{equation}\label{Equality of regularized inner products If and Ig part 7}
        \sum_{i \in W_{j}}t_{ij}^{2} = 1
    \end{equation}
    and, if $j \neq k$,
    \begin{equation}\label{Equality of regularized inner products If and Ig part 8}
        \sum_{i \in W_{j}\cap W_{k}}t_{ij}t_{ik}=0.
    \end{equation}

    In exactly analogous manner, we can also obtain 
    \begin{equation}\label{Equality of regularized inner products If and Ig part 9}
        \sum_{j \in V_{i}}t_{ij}^{2} = 1
    \end{equation}
    and, if $i \neq k$,
    \begin{equation}\label{Equality of regularized inner products If and Ig part 10}
        \sum_{j \in V_{i}\cap V_{k}}t_{ij}t_{kj}=0.
    \end{equation}

    Note that $(e_{j}^{g})_{j=1}^{\infty}$ can be extended to obtain an orthonormal basis for $\LPtwo$. Let $B=\left\{\cup_{j=1}^{\infty}e_{j}^{g}\right\} \cup \left\{\cup_{l=1}^{\infty}z_{l}^{g}\right\}$ be the resulting orthonormal basis of $\LPtwo$ obtained by said extension.

    Now,
    \begin{equation}\label{Expressing one set of eigenfunctions in terms of orthonormal basis}
        \begin{aligned}
            e_{i}^{f} &= \sum_{j=1}^{\infty} \inprod{e_{i}^{f},e_{j}^{g}}e_{j}^{g} + \sum_{l=1}^{\infty} \inprod{e_{i}^{f},z_{l}^{g}}z_{l}^{g}.
        \end{aligned}
    \end{equation}

    Therefore, using \eqref{Expressing one set of eigenfunctions in terms of orthonormal basis} and \eqref{Equality of regularized inner products If and Ig part 9} along with the orthonormality of $(e_{i}^{f})_{i=1}^{\infty}$, we have,
     \begin{equation*}\label{Expressing one set of eigenfunctions in terms of other set of eigenfunctions}
        \begin{aligned}[b]
            &\norm{e_{i}^{f}}_{\LPtwo} = 1\\
            \iff &\sum_{j=1}^{\infty} \inprod{e_{i}^{f},e_{j}^{g}}^{2} + \sum_{l=1}^{\infty} \inprod{e_{i}^{f},z_{l}^{g}}^{2} = 1\\
            \iff & \sum_{j \in V_{i}} t_{ij}^{2} + \sum_{l=1}^{\infty} \inprod{e_{i}^{f},z_{l}^{g}}^{2} = 1\\
            \iff &  \sum_{l=1}^{\infty} \inprod{e_{i}^{f},z_{l}^{g}}^{2} = 0\\
            \iff & \inprod{e_{i}^{f},z_{l}^{g}} = 0 \textrm{ for all l and i}.
        \end{aligned}
    \end{equation*}

    Hence, for all $i$, $e_{i}^{f} \in \operatorname{Span}\left\{e_{j}^{g}, j \in \mathbb{N}\right\}$. Consequently, $\mathcal{T}_{f} e_{j}^{g} = \sum_{i=1}^{\infty} \mu_{i}^{f}t_{ij}e_{i}^{f} \in \operatorname{Span}\left\{e_{i}^{f}, i \in \mathbb{N}\right\} \subset \operatorname{Span}\left\{e_{j}^{g}, j \in \mathbb{N}\right\}$. 

    Now, using \eqref{Equality of regularized inner products If and Ig part 8} and \eqref{Equality of regularized inner products If and Ig part 3}, for any $j \neq k$, we have
    \begin{equation*}
        \begin{aligned}[b]
            \inprod{\mathcal{T}_{f} e_{j}^{g},e_{k}^{g}}_{\LPtwo} = \sum_{i=1}^{\infty} \mu_{i}^{f}t_{ij}t_{ik}
            = \sum_{i \in W_{j}\cap W_{k}} \mu_{i}^{f}t_{ij}t_{ik}
            = \mu_{j}^{g} \sum_{i \in W_{j}\cap W_{k}} t_{ij}t_{ik}
            =0.
        \end{aligned}
    \end{equation*}

    Finally, using \eqref{Equality of regularized inner products If and Ig part 5} and \eqref{Equality of regularized inner products If and Ig part 3}, we have that
    \begin{equation*}
        \begin{aligned}[b]
            \inprod{\mathcal{T}_{f} e_{j}^{g},e_{j}^{g}}_{\LPtwo} = \sum_{i=1}^{\infty} \mu_{i}^{f}t_{ij}^{2}
            = \sum_{i \in W_{j}}\mu_{i}^{f}t_{ij}^{2}
            = \mu_{j}^{g}\sum_{i \in W_{j}}t_{ij}^{2}
            =\mu_{j}^{g} > 0.
        \end{aligned}
    \end{equation*}

    Therefore, $\mathcal{T}_{f} e_{j}^{g}=\mu_{j}^{g}  e_{j}^{g}$ for all $j$. Therefore, all the eigenfunctions of $\mathcal{T}_{g}$ are also eigenfunctions of $\mathcal{T}_{f}$. By symmetry, all the eigenfunctions of $\mathcal{T}_{f}$ are also eigenfunctions of $\mathcal{T}_{g}$. Therefore, $\mathcal{T}_{f}$ and $\mathcal{T}_{g}$ have exactly the same eigenfunctions. 

    Consequently, \eqref{Equality of regularized inner products If and Ig part 1} can be now written as
    \begin{equation}\label{Equality of regularized inner products If and Ig part 11}
        \begin{aligned}[b]
            &\mathcal{I}(f) = \mathcal{I}(g)\\
            \iff &\sum_{i=1}^{\infty} \frac{\mu_{i}^{f}}{\mu_{i}^{f} + \lambda} e_{i}^{f}(\cdot) e_{i}^{f}(\cdot) = \sum_{i=1}^{\infty} \frac{\mu_{i}^{g}}{\mu_{i}^{g} + \lambda} e_{i}^{g}(\cdot) e_{i}^{g}(\cdot) .
        \end{aligned}
    \end{equation}

    Taking the $\LPtwo$ inner product of both the RHS and LHS of \eqref{Equality of regularized inner products If and Ig part 11} with $e_{i}^{f}$ twice, we have that, for any $i$,
    \begin{equation*}
        \begin{aligned}[b]
            &\frac{\mu_{i}^{f}}{\mu_{i}^{f} + \lambda} = \frac{\mu_{i}^{g}}{\mu_{i}^{g} + \lambda}\\
            \iff & \mu_{i}^{f} = \mu_{i}^{g}.
        \end{aligned}
    \end{equation*}

    Therefore, we must have that the integral operators $\mathcal{T}_{f}$ and $\mathcal{T}_{g}$ have the same spectral decomposition. Consequently, their corresponding kernel functions and RKHS's must be the same. Therefore, we must have $ K_{f}(\cdot, \cdot) = K_{g}(\cdot, \cdot)$. This concludes the proof of the necessity part and consequently, the proof of Lemma \ref{Invariance of regularized kernel inner product}.
\end{proof}

\subsection{Proof of Corollary \ref{corollary 1}} \label{Proof of Corollary 1}

\begin{proof}
    Define the operator $\mathcal{I}$ as in Lemma \ref{Invariance of regularized kernel inner product}. Then, we have that, for any $h_{1},h_{2} \in \mathcal{H}$
    \[
    \begin{aligned}
         d_{\lambda,K}^{\text{\metricstname}}(h_{1} \circ \repone,h_{2} \circ \reptwo)
        =& \left(\E\left[\mathcal{I}(h_{1} \circ \repone)(X,X^{\prime}) - \mathcal{I}(h_{2} \circ \reptwo)(X,X^{\prime})\right]^{2}\right)^{\frac{1}{2}}\\
        =& \left(\E\left[\mathcal{I}(\repone)(X,X^{\prime}) - \mathcal{I}(\reptwo)(X,X^{\prime})\right]^{2}\right)^{\frac{1}{2}}
        = d_{\lambda,K}^{\text{\metricstname}}(\repone,\reptwo).
    \end{aligned}
    \]
If either $h_{1}$ or $h_{2}$ does not belong to $\mathcal{H}$, then using Lemma \ref{Invariance of regularized kernel inner product}, we have that \\$\left[\mathcal{I}(h_{1} \circ \repone)(X,X^{\prime}) - \mathcal{I}(h_{2} \circ \reptwo)(X,X^{\prime})\right]^{2}$ must be strictly positive on a set with positive measure w.r.t $P_{X}$. Therefore, we must have $d_{\lambda,K}^{\text{\metricstname}}(h_{1} \circ \repone,h_{2} \circ \reptwo) > 0$.
\end{proof}

\subsection{Proof of Theorem \ref{Theorem: Finite sample convergence}}\label{Proof of Theorem 2}

\begin{proof}
    Note that for any $x,y \in \R^{d}$, $\hat{\Sigma}_{\repone}^{-\lambda}\left[K_{\repone}(\cdot,x)\otimes_{\Hone}K_{\repone}(\cdot,x)\right]\hat{\Sigma}_{\repone}^{-\lambda}\left[K_{\repone}(\cdot,y)\otimes_{\Hone}K_{\repone}(\cdot,y)\right]$ is a rank-one operator with eigenvalue $\inprod{\hat{\Sigma}_{\repone}^{-\frac{\lambda}{2}}K_{\repone}(\cdot,x),\hat{\Sigma}_{\repone}^{-\frac{\lambda}{2}}K_{\repone}(\cdot,y)}_{\Hone}^{2}$ and eigenfunction $\frac{\hat{\Sigma}_{\repone}^{-\lambda}K_{\repone}(\cdot,x)}{\norm{\hat{\Sigma}_{\repone}^{-\lambda}K_{\repone}(\cdot,x)}_{\Hone}}$. Similarly, $\hat{\Sigma}_{\reptwo}^{-\lambda}\left[K_{\reptwo}(\cdot,x)\otimes_{\Htwo}K_{\reptwo}(\cdot,x)\right]\hat{\Sigma}_{\reptwo}^{-\lambda}\left[K_{\reptwo}(\cdot,y)\otimes_{\Htwo}K_{\reptwo}(\cdot,y)\right]$ is a rank-one operator with eigenvalue $\inprod{\hat{\Sigma}_{\reptwo}^{-\frac{\lambda}{2}}K_{\reptwo}(\cdot,x),\hat{\Sigma}_{\reptwo}^{-\frac{\lambda}{2}}K_{\reptwo}(\cdot,y)}_{\Htwo}^{2}$ and eigenfunction $\frac{\hat{\Sigma}_{\reptwo}^{-\lambda}K_{\reptwo}(\cdot,x)}{\norm{\hat{\Sigma}_{\reptwo}^{-\lambda}K_{\reptwo}(\cdot,x)}_{\Htwo}}$. Further, \\$\hat{\Sigma}_{\repone}^{-\lambda}\left[K_{\repone}(\cdot,x)\otimes_{\HS(\Hone,\Htwo)}K_{\reptwo}(\cdot,x)\right] \times\hat{\Sigma}_{\reptwo}^{-\lambda}\left[K_{\reptwo}(\cdot,y)\otimes_{\HS(\Hone,\Htwo)}K_{\repone}(\cdot,y)\right]$ is a rank-one operator with eigenvalue $\inprod{\hat{\Sigma}_{\repone}^{-\frac{\lambda}{2}}K_{\repone}(\cdot,x),\hat{\Sigma}_{\repone}^{-\frac{\lambda}{2}}K_{\repone}(\cdot,y)}_{\Hone}\times\inprod{\hat{\Sigma}_{\reptwo}^{-\frac{\lambda}{2}}K_{\reptwo}(\cdot,x),\hat{\Sigma}_{\reptwo}^{-\frac{\lambda}{2}}K_{\reptwo}(\cdot,y)}_{\Htwo}$ and eigenfunction $\frac{\hat{\Sigma}_{\repone}^{-\lambda}K_{\repone}(\cdot,x)}{\norm{\hat{\Sigma}_{\repone}^{-\lambda}K_{\repone}(\cdot,x)}_{\Hone}}$.

    Using these facts, we have that the squared V-statistic type estimator of $d_{\lambda,K}^{\text{\metricstname}}$ can be expressed as
    \[
    \begin{aligned}
&\left[\hat{d}_{\lambda}^{\text{\metricstname}}(\repone,\reptwo)\right]^{2}\\
        =&\frac{1}{n^{2}}\sum_{i=1}^{n}\sum_{j=1}^{n}\left[\inprod{\hat{\Sigma}_{\repone}^{-\frac{\lambda}{2}}K_{\repone}(\cdot,X_{i}),\hat{\Sigma}_{\repone}^{-\frac{\lambda}{2}}K_{\repone}(\cdot,X_{j})}_{\Hone}\right.\left.-\inprod{\hat{\Sigma}_{\reptwo}^{-\frac{\lambda}{2}}K_{\reptwo}(\cdot,X_{i}),\hat{\Sigma}_{\reptwo}^{-\frac{\lambda}{2}}K_{\reptwo}(\cdot,X_{j})}_{\Htwo}\right]^{2}.
    \end{aligned}
    \]
Let us define the following quantity
\[
\begin{aligned}
    &\left[\tilde{d}_{\lambda}^{\text{\metricstname}}(\repone,\reptwo)\right]^{2}\\
    \coloneq&\frac{1}{n^{2}}\sum_{i=1}^{n}\sum_{j=1}^{n}\left[\inprod{\Sigma_{\repone}^{-\frac{\lambda}{2}}K_{\repone}(\cdot,X_{i}),\Sigma_{\repone}^{-\frac{\lambda}{2}}K_{\repone}(\cdot,X_{j})}_{\Hone}\right.\left.-\inprod{\Sigma_{\reptwo}^{-\frac{\lambda}{2}}K_{\reptwo}(\cdot,X_{i}),\Sigma_{\reptwo}^{-\frac{\lambda}{2}}K_{\reptwo}(\cdot,X_{j})}_{\Htwo}\right]^{2} 
\end{aligned}
\]
which is $\left[\hat{d}_{\lambda}^{\text{\metricstname}}(\repone,\reptwo)\right]^{2}$ with $\hat{\Sigma}_{\repone}$ and $\hat{\Sigma}_{\repone}$ replaced by $\Sigma_{\repone}$ and $\Sigma_{\repone}$, respectively. We utilize the triangle inequality to bound the difference between the squared V-statistic type estimator $\left[\hat{d}_{\lambda}^{\text{\metricstname}}(\repone,\reptwo)\right]^{2}$ and the squared population distance $\left[d_{\lambda,K}^{\text{\metricstname}}(\repone,\reptwo)\right]^{2}$ as follows:
\begin{equation}\label{Triangle inequality for concentration result}
    \begin{aligned}
        &\left|\left[\hat{d}_{\lambda}^{\text{\metricstname}}(\repone,\reptwo)\right]^{2} - \left[d_{\lambda,K}^{\text{\metricstname}}(\repone,\reptwo)\right]^{2}\right|\\
        &\leq \underbrace{\left|\left[\hat{d}_{\lambda}^{\text{\metricstname}}(\repone,\reptwo)\right]^{2} - \left[\tilde{d}_{\lambda}^{\text{\metricstname}}(\repone,\reptwo)\right]^{2}\right|}_{\mathbf{A}} +\underbrace{\left|\left[\tilde{d}_{\lambda}^{\text{\metricstname}}(\repone,\reptwo)\right]^{2} - \left[d_{\lambda,K}^{\text{\metricstname}}(\repone,\reptwo)\right]^{2}\right|}_{\mathbf{B}}.
    \end{aligned}
\end{equation}

We now proceed to bound $\mathbf{A}$. Let us define
\[
\hat{A}_{ij,\repone} = \inprod{\hat{\Sigma}_{\repone}^{-\frac{\lambda}{2}}K_{\repone}(\cdot,X_{i}),\hat{\Sigma}_{\repone}^{-\frac{\lambda}{2}}K_{\repone}(\cdot,X_{j})}_{\Hone},
\]
\[
A_{ij,\repone} = \inprod{\Sigma_{\repone}^{-\frac{\lambda}{2}}K_{\repone}(\cdot,X_{i}),\Sigma_{\repone}^{-\frac{\lambda}{2}}K_{\repone}(\cdot,X_{j})}_{\Hone},
\]
\[
\hat{A}_{ij,\reptwo} = \inprod{\hat{\Sigma}_{\reptwo}^{-\frac{\lambda}{2}}K_{\reptwo}(\cdot,X_{i}),\hat{\Sigma}_{\reptwo}^{-\frac{\lambda}{2}}K_{\reptwo}(\cdot,X_{j})}_{\Htwo},
\]
\[
A_{ij,\reptwo} = \inprod{\Sigma_{\reptwo}^{-\frac{\lambda}{2}}K_{\reptwo}(\cdot,X_{i}),\Sigma_{\reptwo}^{-\frac{\lambda}{2}}K_{\reptwo}(\cdot,X_{j})}_{\Htwo}.
\]
Then, we have that 
\[
\left|\hat{A}_{ij,\repone}\right|\leq \norm{K_{\repone}(\cdot,X_{i})}_{\Hone}^{2} \times \norm{\hat{\Sigma}_{\repone}^{-\frac{\lambda}{2}}}_{\Op(\Hone)}^{2}\leq \frac{\kappa}{\lambda}.
\]
Similarly, we can show that $\left|\hat{A}_{ij,\reptwo}\right|\leq \frac{\kappa}{\lambda}$,$\left|A_{ij,\repone}\right|\leq \frac{\kappa}{\lambda}$ and $\left|A_{ij,\reptwo}\right|\leq \frac{\kappa}{\lambda}$. Now, we have that
\[
\begin{aligned}
    \left|\hat{A}_{ij,\repone} - A_{ij,\repone}\right|
    =&\left|\inprod{K_{\repone}(\cdot,X_{i}),\left(\hat{\Sigma}_{\repone}^{-\lambda}-\Sigma_{\repone}^{-\lambda}\right)K_{\repone}(\cdot,X_{j})}_{\Hone}\right|\\
    &\leq \kappa \norm{\hat{\Sigma}_{\repone}^{-\lambda}-\Sigma_{\repone}^{-\lambda}}_{\Op(\Hone)}
    \leq \kappa \norm{\hat{\Sigma}_{\repone}^{-\lambda}-\Sigma_{\repone}^{-\lambda}}_{\HS(\Hone)}.
\end{aligned}
\]
Similarly, we have that
\[
\begin{aligned}
    \left|\hat{A}_{ij,\reptwo} - A_{ij,\reptwo}\right|=&\left|\inprod{K_{\reptwo}(\cdot,X_{i}),\left(\hat{\Sigma}_{\reptwo}^{-\lambda}-\Sigma_{\reptwo}^{-\lambda}\right)K_{\reptwo}(\cdot,X_{j})}_{\Htwo}\right|\\
    &\leq \kappa \norm{\hat{\Sigma}_{\reptwo}^{-\lambda}-\Sigma_{\reptwo}^{-\lambda}}_{\Op(\Htwo)}
    \leq \kappa \norm{\hat{\Sigma}_{\reptwo}^{-\lambda}-\Sigma_{\reptwo}^{-\lambda}}_{\HS(\Htwo)}.
\end{aligned}
\]

Note that,
\[
\begin{aligned}
    &\norm{\hat{\Sigma}_{\repone}^{-\lambda}-\Sigma_{\repone}^{-\lambda}}_{\Op(\Hone)}\\
    =&\left\|\left(\hat{\Sigma}_{\repone}+\lambda I\right)^{-1}\left(\Sigma_{\repone}+\lambda I\right)\left(\Sigma_{\repone}+\lambda I\right)^{-1} \right.\left.- \left(\hat{\Sigma}_{\repone}+\lambda I\right)^{-1}\left(\hat{\Sigma}_{\repone}+\lambda I\right)\left(\Sigma_{\repone}+\lambda I\right)^{-1}\right\|_{\Op(\Hone)}\\
    =&\norm{\hat{\Sigma}_{\repone}^{-\lambda}\left[\left(\Sigma_{\repone}+\lambda I\right)-\left(\hat{\Sigma}_{\repone}+\lambda I\right)\right]\Sigma_{\repone}^{-\lambda}}_{\Op(\Hone)}\\
    \leq& \norm{\hat{\Sigma}_{\repone}^{-\lambda}}_{\Op(\Hone)} \norm{\Sigma_{\repone} - \hat{\Sigma}_{\repone}}_{\Op(\Hone)} \norm{\Sigma_{\repone}^{-\lambda}}_{\Op(\Hone)}\\
    \leq& \frac{1}{\lambda^{2}}\norm{\Sigma_{\repone} - \hat{\Sigma}_{\repone}}_{\Op(\Hone)}
    \leq \frac{1}{\lambda^{2}}\norm{\Sigma_{\repone} - \hat{\Sigma}_{\repone}}_{\HS(\Hone)}.
\end{aligned}
\]
Similarly, $\norm{\hat{\Sigma}_{\reptwo}^{-\lambda}-\Sigma_{\reptwo}^{-\lambda}}_{\Op(\Htwo)}\leq \frac{1}{\lambda^{2}}\norm{\Sigma_{\reptwo} - \hat{\Sigma}_{\reptwo}}_{\Op(\Htwo)}\leq \frac{1}{\lambda^{2}}\norm{\Sigma_{\reptwo} - \hat{\Sigma}_{\reptwo}}_{\HS(\Htwo)}$.

Therefore, we have that 
\begin{equation}\label{Bound on A}
\begin{aligned}[b]
    \mathbf{A}
    =&\left|\frac{1}{n^{2}}\sum_{i=1}^{n}\sum_{j=1}^{n}\left[\left(\hat{A}_{ij,\repone}-\hat{A}_{ij,\reptwo}\right)^{2} - \left(A_{ij,\repone}-A_{ij,\reptwo}\right)^{2}\right]\right|\\
    =& \left|\frac{1}{n^{2}}\sum_{i=1}^{n}\sum_{j=1}^{n}\left[\left(\hat{A}_{ij,\repone}-\hat{A}_{ij,\reptwo}\right) - \left(A_{ij,\repone}-A_{ij,\reptwo}\right)\right]\right.\left.\left[\left(\hat{A}_{ij,\repone}-\hat{A}_{ij,\reptwo}\right) + \left(A_{ij,\repone}-A_{ij,\reptwo}\right)\right]\right|\\
    \leq& \kappa \left(\norm{\hat{\Sigma}_{\repone}^{-\lambda}-\Sigma_{\repone}^{-\lambda}}_{\Op(\Hone)} + \norm{\hat{\Sigma}_{\reptwo}^{-\lambda}-\Sigma_{\reptwo}^{-\lambda}}_{\Op(\Htwo)}\right)\times \left(\frac{2\kappa}{\lambda}+\frac{2\kappa}{\lambda}\right)\\
    =&\frac{4\kappa^{2}}{\lambda}\left[\norm{\hat{\Sigma}_{\repone}^{-\lambda}-\Sigma_{\repone}^{-\lambda}}_{\Op(\Hone)} + \norm{\hat{\Sigma}_{\reptwo}^{-\lambda}-\Sigma_{\reptwo}^{-\lambda}}_{\Op(\Htwo)}\right]\\
    \leq&\frac{4\kappa^{2}}{\lambda^{3}}\left[\norm{\hat{\Sigma}_{\repone}-\Sigma_{\repone}}_{\HS(\Hone)} + \norm{\hat{\Sigma}_{\reptwo}-\Sigma_{\reptwo}}_{\HS(\Htwo)}\right].
\end{aligned}
\end{equation}

Let us define $Z_{i}^{\repone} = K_{\repone}(\cdot,X_{i})\otimes_{\Hone}K_{\repone}(\cdot,X_{i})$. Then, $Z_{i}^{\repone}$'s are i.i.d random variables, $\E(Z_{i}^{\repone}) = \Sigma_{\repone}$ and $\hat{\Sigma}_{\repone} - \Sigma_{\repone} = \frac{1}{n}\sum_{i=1}^{n}\left[Z_{i}^{\repone}-\E(Z_{i}^{\repone})\right]$. Similarly, let us define $Z_{i}^{\reptwo} = K_{\reptwo}(\cdot,X_{i})\otimes_{\Htwo}K_{\reptwo}(\cdot,X_{i})$. Then $Z_{i}^{\reptwo}$'s are i.i.d random variables, $\E(Z_{i}^{\reptwo}) = \Sigma_{\reptwo}$ and $\hat{\Sigma}_{\reptwo} - \Sigma_{\reptwo} = \frac{1}{n}\sum_{i=1}^{n}\left[Z_{i}^{\reptwo}-\E(Z_{i}^{\reptwo})\right]$.

Note that,
\[
\begin{aligned}
    \norm{Z_{i}^{\repone}}_{\HS(\Hone)}    &=\sqrt{\inprod{Z_{i}^{\repone},Z_{i}^{\repone}}_{\HS(\Hone)}}
    =\inprod{K_{\repone}(\cdot,X_{i}),K_{\repone}(\cdot,X_{i})}_{\Hone}
    =K_{\repone}(X_{i},X_{i})
    \leq \kappa \coloneq B.
\end{aligned}
\]

Further, 
\[
\begin{aligned}
   &\E\norm{Z_{i}^{\repone}-\E(Z_{i}^{\repone})}_{\HS(\Hone)}^{2}
   =\E\left[\inprod{Z_{i}^{\repone},Z_{i}^{\repone}}_{\HS(\Hone)}\right] - \inprod{\Sigma_{\repone},\Sigma_{\repone}}_{\HS(\Hone)}
   \leq& \E\left[\inprod{Z_{i}^{\repone},Z_{i}^{\repone}}_{\HS(\Hone)}\right]\\
   =&\E\left[\inprod{K_{\repone}(\cdot,X_{i}),K_{\repone}(\cdot,X_{i})}_{\Hone}^{2}\right]
   =\E\left[K_{\repone}(X_{i},X_{i})^{2}\right]
   \leq \kappa^{2} \coloneq \theta^{2}.
\end{aligned}
\]
Similarly, we can show that $\norm{Z_{i}^{\reptwo}}_{\HS(\Htwo)}\leq \kappa = B$ and $\E\norm{Z_{i}^{\repone}-\E(Z_{i}^{\repone})}_{\HS(\Hone)}^{2}\leq \kappa^{2} = \theta^{2}$.

Note that since $K(\cdot,\cdot)$ is bounded and continuous,$\Hone$ and $\Htwo$ are separable Hilbert spaces. Now, using Bernstein's inequality for separable Hilbert spaces (Theorem D.1 in \citet{sriperumbudur2022approximate}), we have that, for any $0<\delta<1$,
\[
P\left(\norm{\hat{\Sigma}_{\repone}-\Sigma_{\repone}}_{\HS(\Hone)} \geq \frac{2\kappa\log(\frac{6}{\delta})}{n} + \sqrt{\frac{2\kappa^{2}\log(\frac{6}{\delta})}{n}}\right) \leq \frac{\delta}{3}
\]
and 
\[
P\left(\norm{\hat{\Sigma}_{\reptwo}-\Sigma_{\reptwo}}_{\HS(\Hone)} \geq \frac{2\kappa\log(\frac{6}{\delta})}{n} + \sqrt{\frac{2\kappa^{2}\log(\frac{6}{\delta})}{n}}\right) \leq \frac{\delta}{3}.
\]

Therefore, we have that, for any $0<\delta<1$,
\[
\begin{aligned}
    &P\left(\mathbf{A}=\left|\left[\hat{d}_{\lambda}^{\text{\metricstname}}(\repone,\reptwo)\right]^{2} - \left[\tilde{d}_{\lambda}^{\text{\metricstname}}(\repone,\reptwo)\right]^{2}\right|\right.\geq \left.\frac{8\kappa^{2}}{\lambda^{3}}\left[\frac{2\kappa\log(\frac{6}{\delta})}{n} + \sqrt{\frac{2\kappa^{2}\log(\frac{6}{\delta})}{n}}\right]\right) \leq \frac{2\delta}{3}.
\end{aligned}
\]

We now proceed to bound $\mathbf{B}$.

Let us define 
\[
\begin{aligned}
    b_{ij} \coloneq& \frac{1}{n^{2}}\left[\inprod{\Sigma_{\repone}^{-\frac{\lambda}{2}}K_{\repone}(\cdot,X_{i}),\Sigma_{\repone}^{-\frac{\lambda}{2}}K_{\repone}(\cdot,X_{j})}_{\Hone}-\right.\left.\inprod{\Sigma_{\reptwo}^{-\frac{\lambda}{2}}K_{\reptwo}(\cdot,X_{i}),\Sigma_{\reptwo}^{-\frac{\lambda}{2}}K_{\reptwo}(\cdot,X_{j})}_{\Htwo}\right]^{2}\\
    =&\frac{1}{n^{2}}\left[A_{ij,\repone}-A_{ij,\reptwo}\right]^{2}.
\end{aligned}
\]
Then, clearly, we have that $\left(b_{ij}\right)_{i,j=1,i \neq j}^{n}$'s are i.i.d random variables. Similarly, $\left(b_{ii}\right)_{i=1}^{n}$ are i.i.d random variables. Further, $\E(b_{ij})=\frac{\left[d_{\lambda,K}^{\text{\metricstname}}(\repone,\reptwo)\right]^{2}}{n^{2}}$ if $i\neq j$ and $|b_{ij}|\leq \frac{1}{n^{2}}\left[|A_{ij,\repone}|+|A_{ij,\reptwo}|\right]^{2}\leq \frac{4\kappa^{2}}{\lambda^{2} n^{2}}$ for any $i,j$. Therefore, $\left|\E(b_{ij})\right|\leq \E\left|b_{ij}\right|\leq  \frac{4\kappa^{2}}{\lambda^{2} n^{2}}$ for any $i,j$ and $\left[d_{\lambda,K}^{\text{\metricstname}}(\repone,\reptwo)\right]^{2} \leq \frac{4\kappa^{2}}{\lambda^{2}}$.

Now, we have that, $$\E\left[\tilde{d}_{\lambda}^{\text{\metricstname}}(\repone,\reptwo)\right]^{2} = \frac{n(n-1)}{n^{2}}\left[d_{\lambda,K}^{\text{\metricstname}}(\repone,\reptwo)\right]^{2} + n\E(b_{11}).$$ Consequently, $\left[d_{\lambda,K}^{\text{\metricstname}}(\repone,\reptwo)\right]^{2} - \E\left[\tilde{d}_{\lambda}^{\text{\metricstname}}(\repone,\reptwo)\right]^{2} = \frac{1}{n}\left[d_{\lambda,K}^{\text{\metricstname}}(\repone,\reptwo)\right]^{2} - n\E(b_{11})$. Therefore, $$\left|\left[d_{\lambda,K}^{\text{\metricstname}}(\repone,\reptwo)\right]^{2} - \E\left[\tilde{d}_{\lambda}^{\text{\metricstname}}(\repone,\reptwo)\right]^{2}\right|\leq \frac{8\kappa^{2}}{\lambda^{2} n}.$$ 

Now, using McDiarmid's inequality, we have that,
\[
\begin{aligned}
    P\left(\left|\left[\tilde{d}_{\lambda}^{\text{\metricstname}}(\repone,\reptwo)\right]^{2} - \E\left[\tilde{d}_{\lambda}^{\text{\metricstname}}(\repone,\reptwo)\right]^{2}\right|\geq \right. \left.\frac{4\kappa^{2}}{\lambda^{2}}\sqrt{\frac{2\log(\frac{6}{\delta})}{n}}\right)\leq \frac{\delta}{3}.
\end{aligned}
\]

Therefore, we have that,
\[
\begin{aligned}
    P\left(\mathbf{B} \geq \frac{\kappa^{2}}{\lambda^{2}}\left[\frac{8}{n} + 4\sqrt{\frac{2\log(\frac{6}{\delta})}{n}}\right]\right) \leq \frac{\delta}{3}.
\end{aligned}
\]

Finally, we have that,
\[
\begin{aligned}
P\left(\mathbf{A}+\mathbf{B} \leq \frac{8\kappa^{3}}{\lambda^{3}}\left[\frac{2\log(\frac{6}{\delta})}{n} + \sqrt{\frac{2\log(\frac{6}{\delta})}{n}}\right] + \frac{4\kappa^{2}}{\lambda^{2}}\left[\frac{2}{n} + \sqrt{\frac{2\log(\frac{6}{\delta})}{n}}\right]\right)
\geq 1-\delta,
\end{aligned}
\]
which completes the proof.
\end{proof}

\newpage

\section{Additional Experiments}
In this appendix, we provide additional experimental results.
\subsection{MNIST experiments}
\label{MNIST Experiments additional}

\paragraph{Training details} We have already described the architectures of the 50 ReLU networks we trained for experiments using the MNIST dataset in Section \ref{MNIST experiments}. We used the uniform Kaiming initialization \cite{he2015delving} for initializing the network weights for every network with a specific width and depth, while the biases are set to zero at initialization. We used a single A100 GPU on the Google Colab platform. We chose to use the Adam optimizer with a learning rate of $10^{-4}$ and a batch size of 100 to train the 50 ReLU networks. We follow a training scheme similar to that used in \citet{GULP}.

\paragraph{Clustering of representations based on UKP aligns with architectural characteristics of networks}

\begin{figure}[!h]
    \centering
    \begin{subfigure}[b]{0.3\textwidth}
        \includegraphics[width=\textwidth]{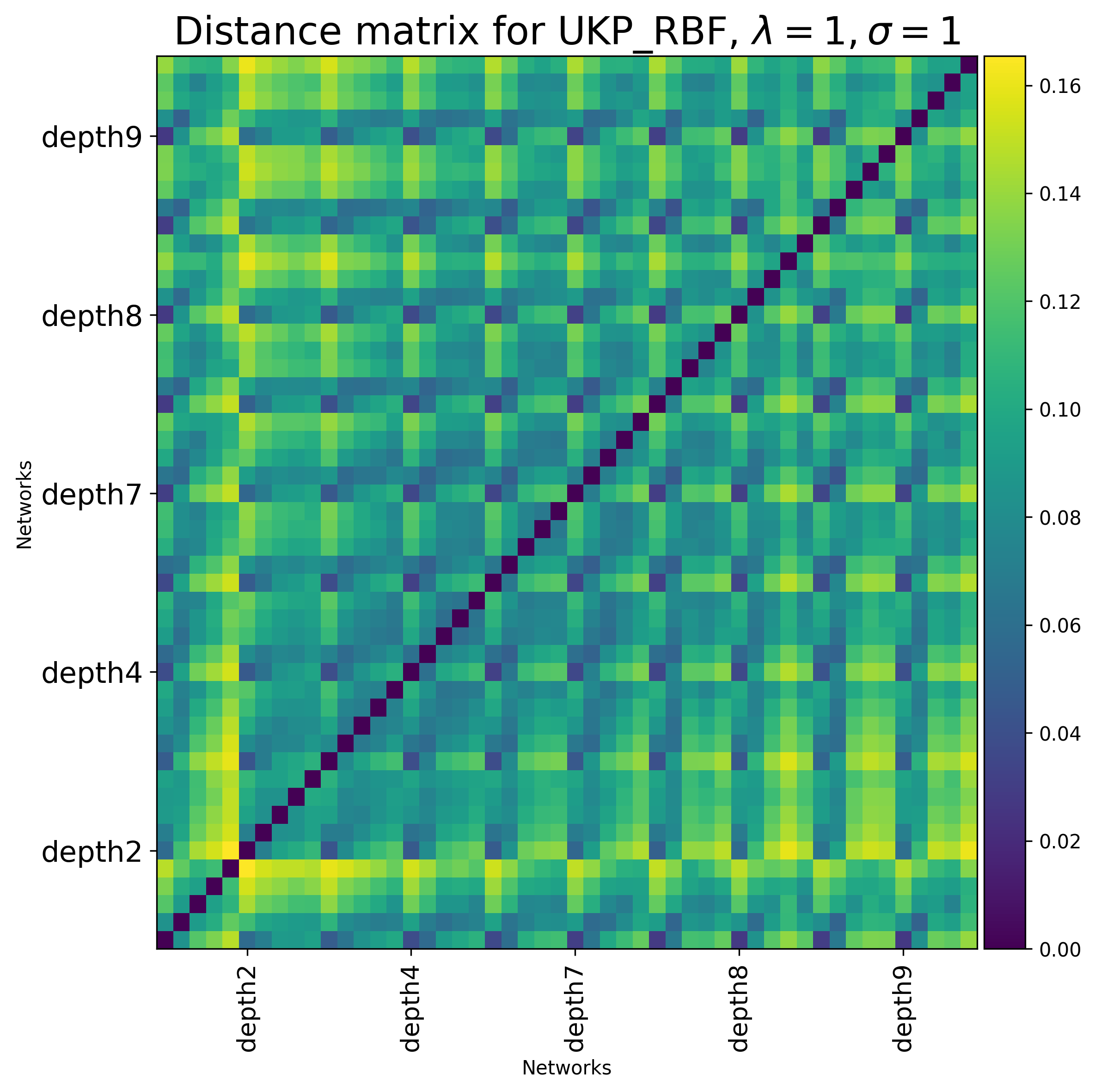}
    \end{subfigure}
    \hfill
    \begin{subfigure}[b]{0.3\textwidth}
        \includegraphics[width=\textwidth]{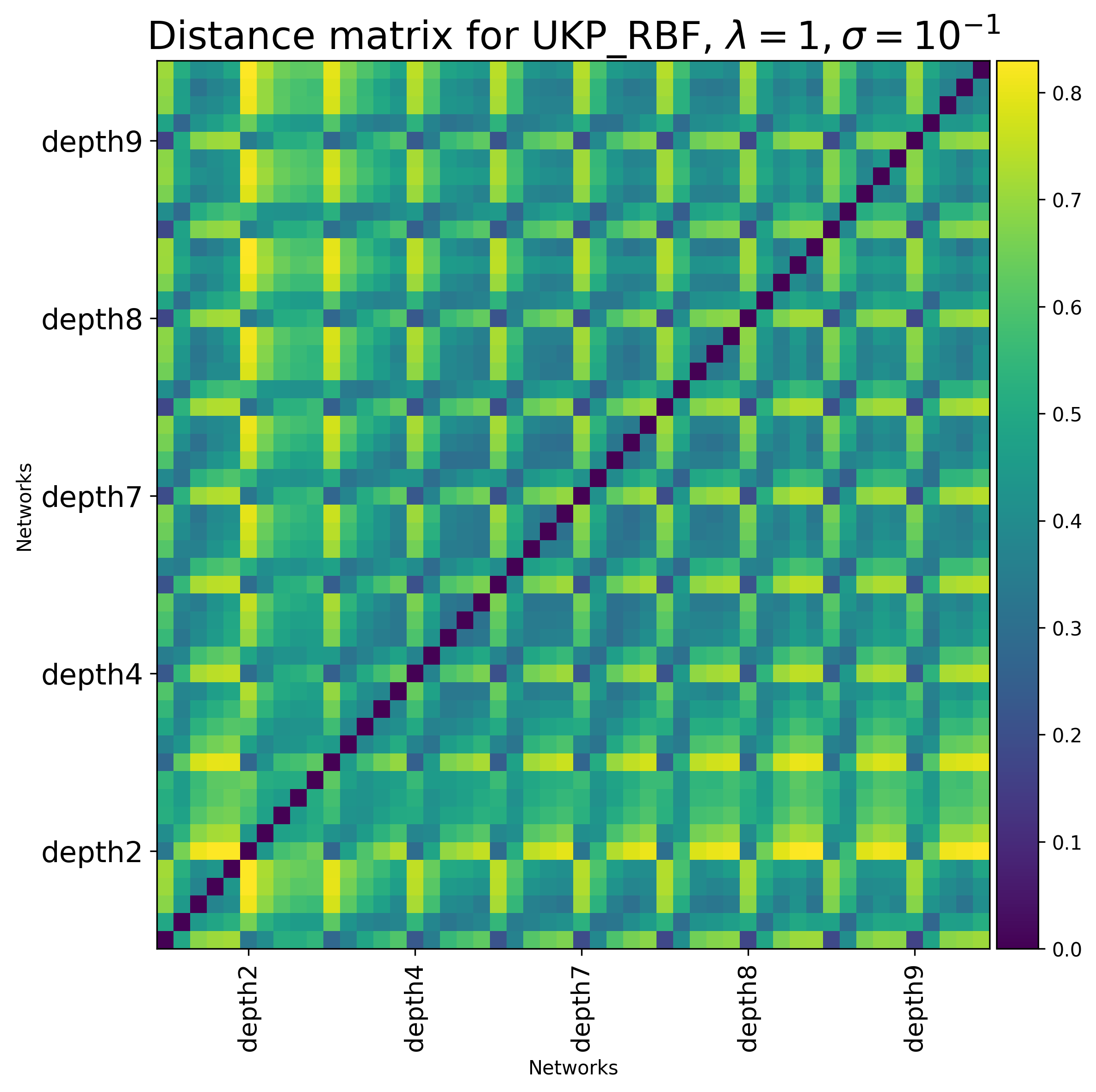}
    \end{subfigure}
    \hfill
    \begin{subfigure}[b]{0.3\textwidth}
        \includegraphics[width=\textwidth]{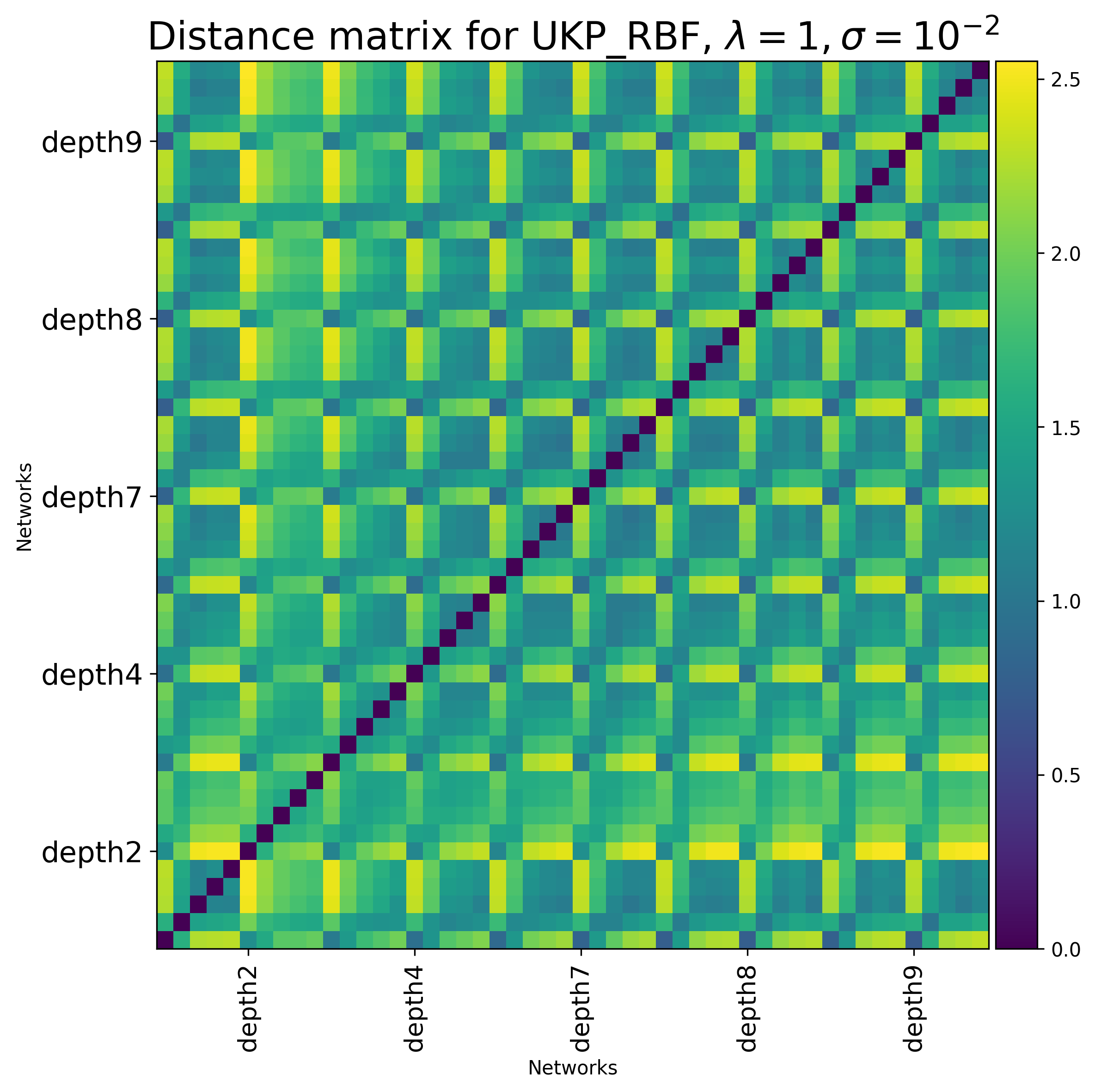}
    \end{subfigure}
    
    \vspace{0.5cm}  
    
    \begin{subfigure}[b]{0.3\textwidth}
        \includegraphics[width=\textwidth]{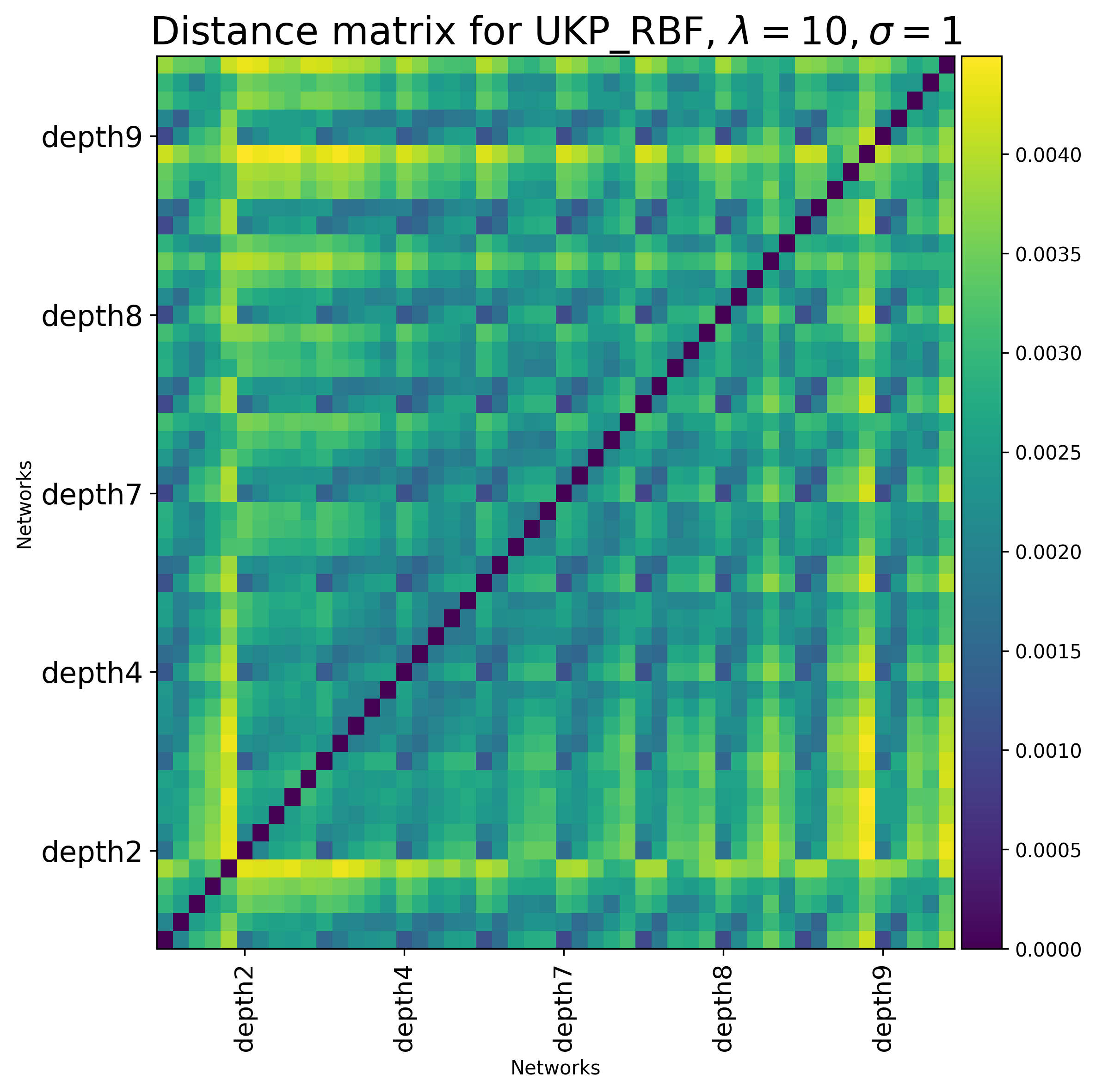}
    \end{subfigure}
    \hfill
    \begin{subfigure}[b]{0.3\textwidth}
        \includegraphics[width=\textwidth]{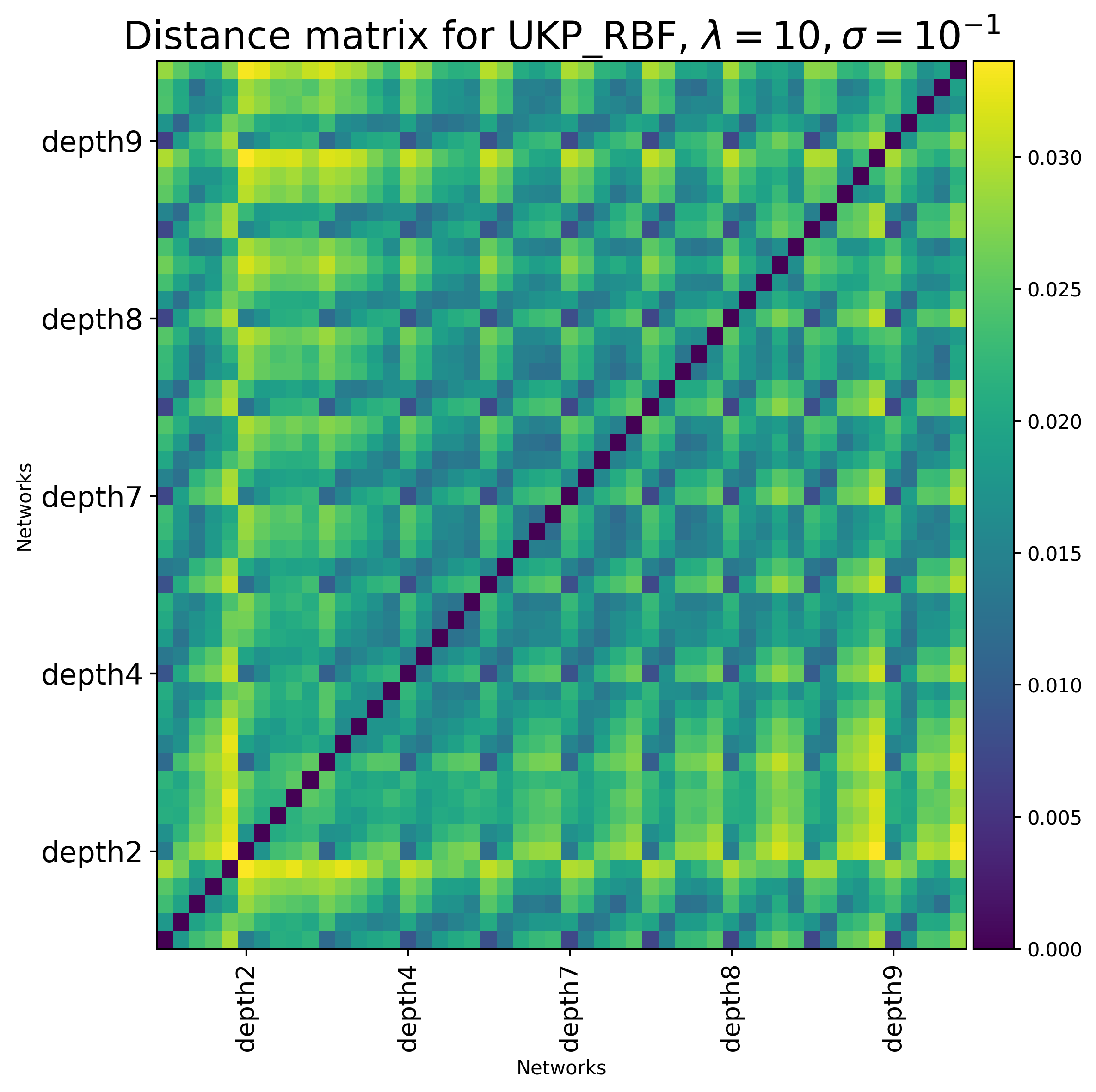}
    \end{subfigure}
    \hfill
    \begin{subfigure}[b]{0.3\textwidth}
        \includegraphics[width=\textwidth]{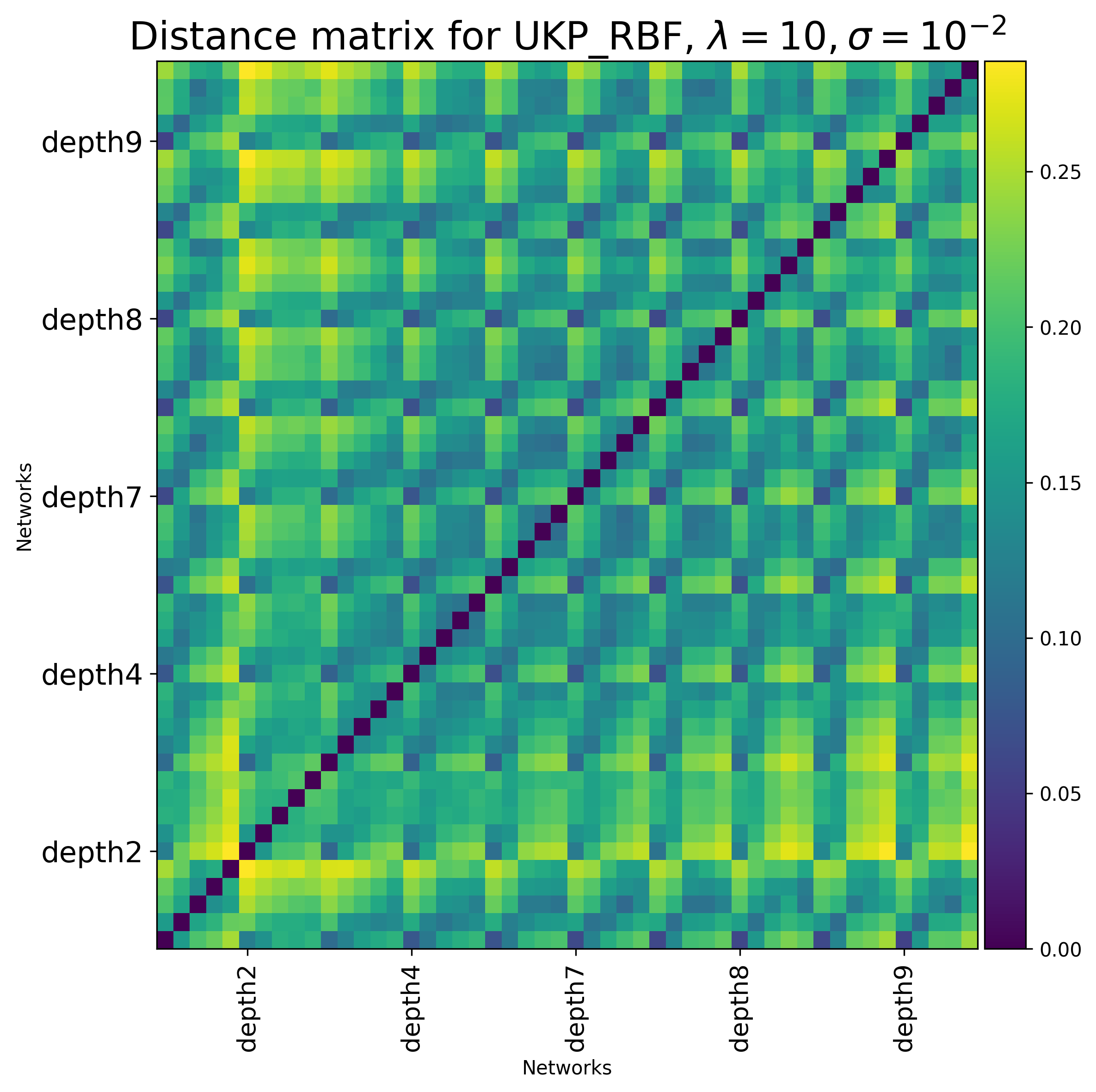}
    \end{subfigure}
    
    \caption{Heatmaps representing \metricstname distance between pairs of fully-connected ReLU networks of different depths and widths. We choose the kernel for the \metricstname distance to be the Gaussian RBF kernel with bandwidth $\sigma \in \left\{1,10^{-1},10^{-2}\right\}$ along with the regularization parameter $\lambda \in \left\{1,10\right\}$. Along the rows and columns of each of the heatmaps, the ReLU networks are first arranged in order of increasing depth, and then in order of increasing width inside each specific depth level. Darker colors indicate smaller value of \metricstname distance according to the scale attached to each heatmap.}
    \label{MNIST Heatmaps}
\end{figure}

\begin{figure}[!h]
    \centering
    \begin{subfigure}[b]{0.45\textwidth}
        \includegraphics[width=\textwidth]{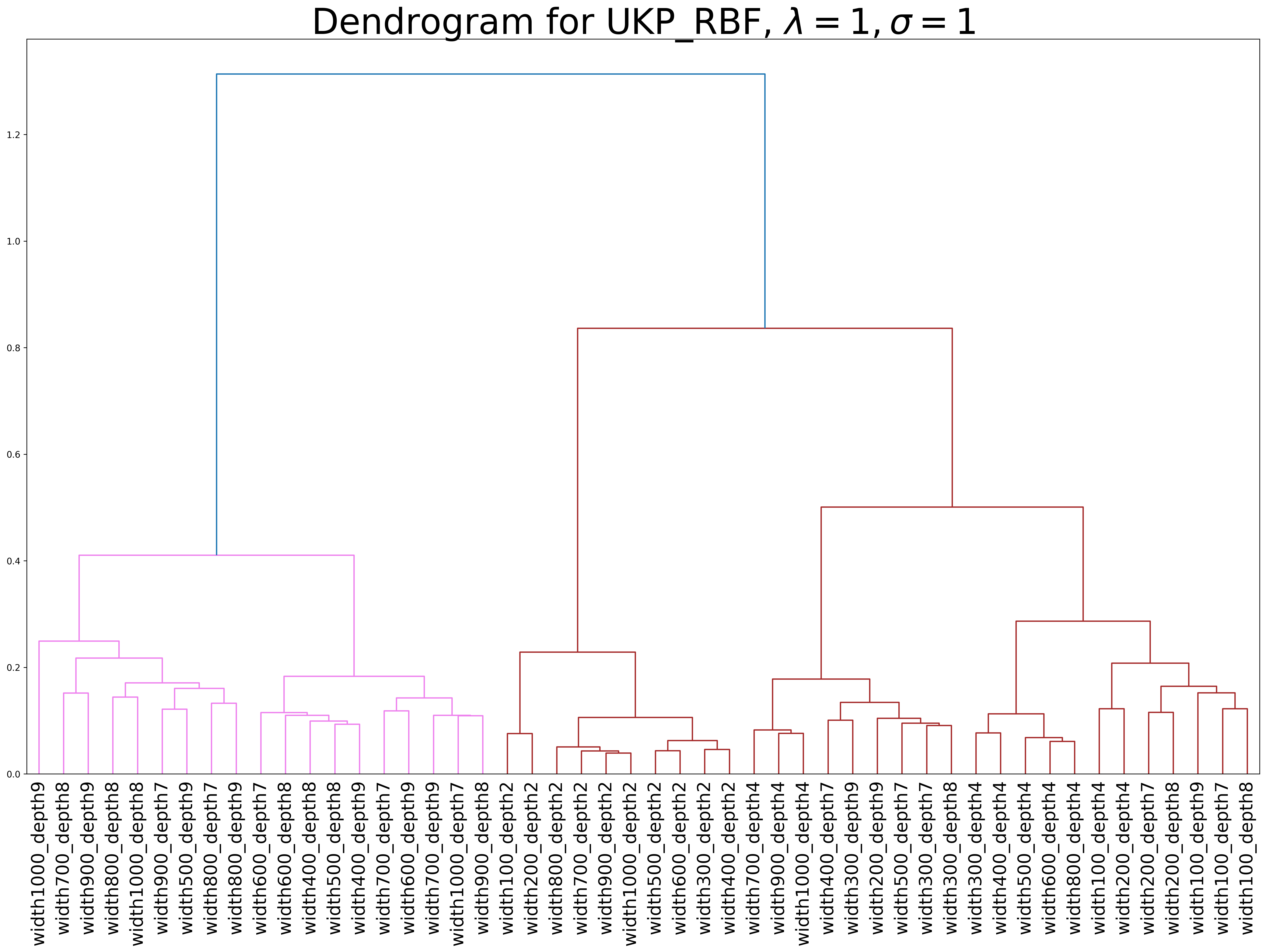}
    \end{subfigure}
    \hfill
    \begin{subfigure}[b]{0.45\textwidth}
        \includegraphics[width=\textwidth]{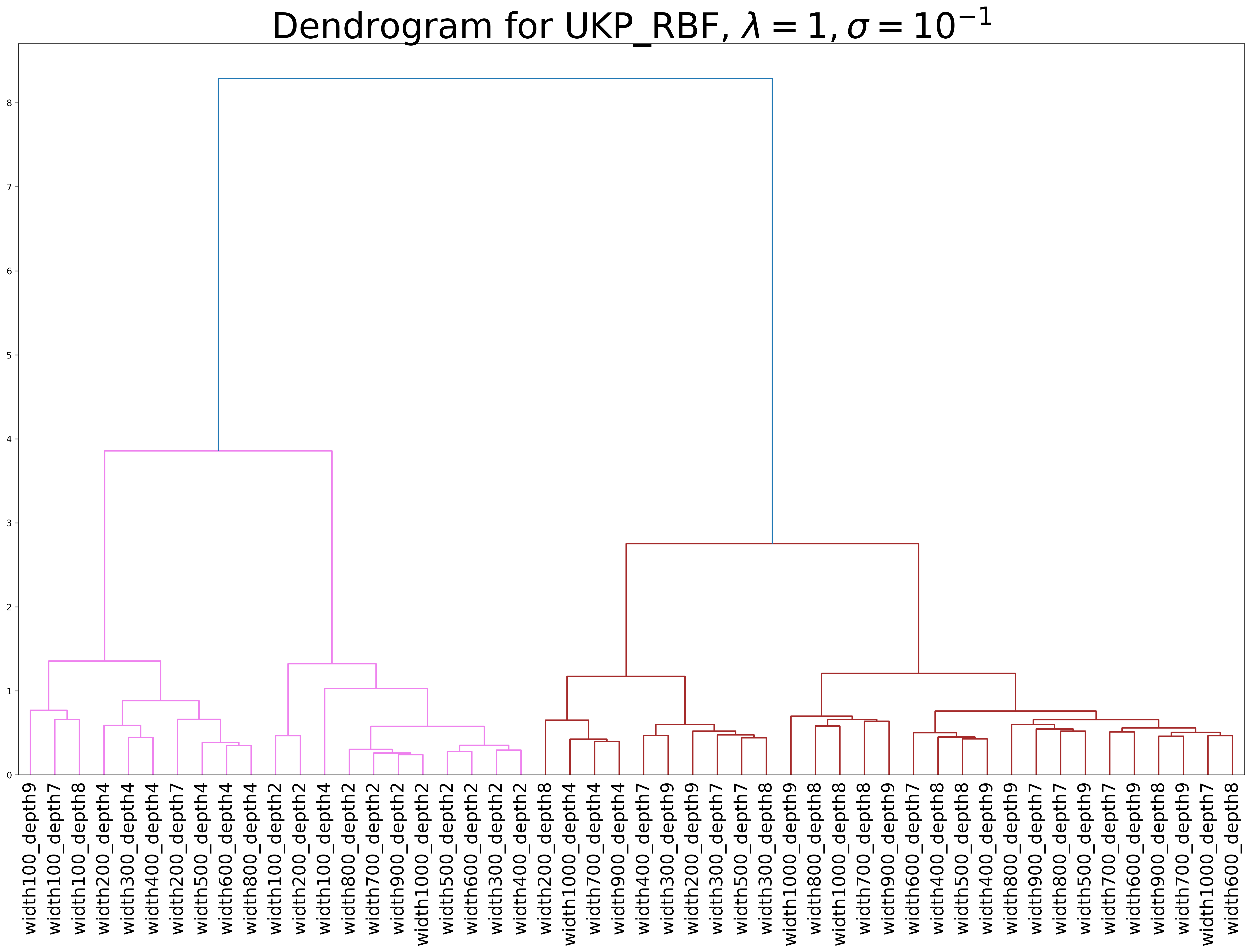}
    \end{subfigure}
    
    \vspace{0.5cm}  
    
    \begin{subfigure}[b]{0.45\textwidth}
        \includegraphics[width=\textwidth]{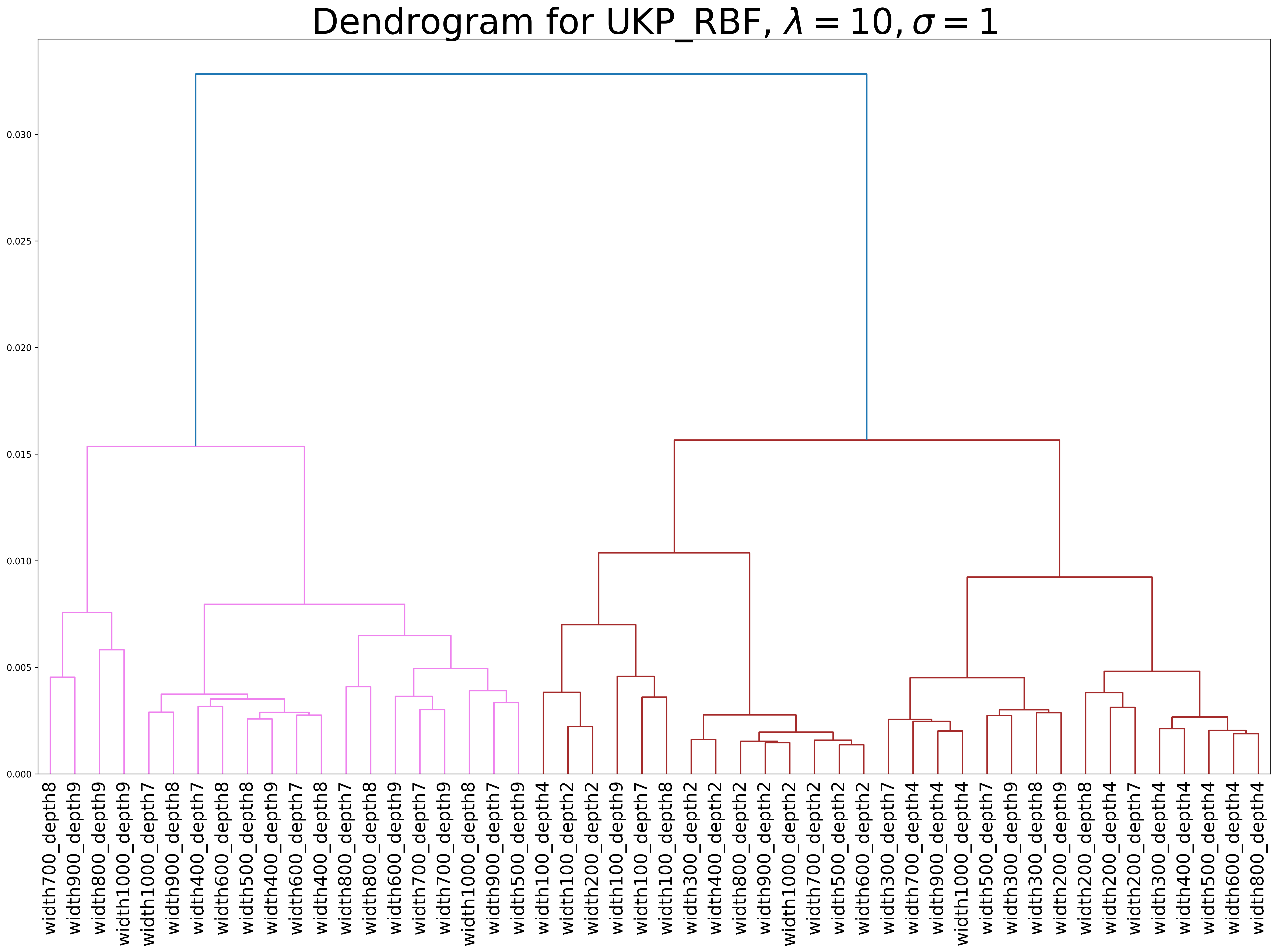}
    \end{subfigure}
    \hfill
    \begin{subfigure}[b]{0.45\textwidth}
        \includegraphics[width=\textwidth]{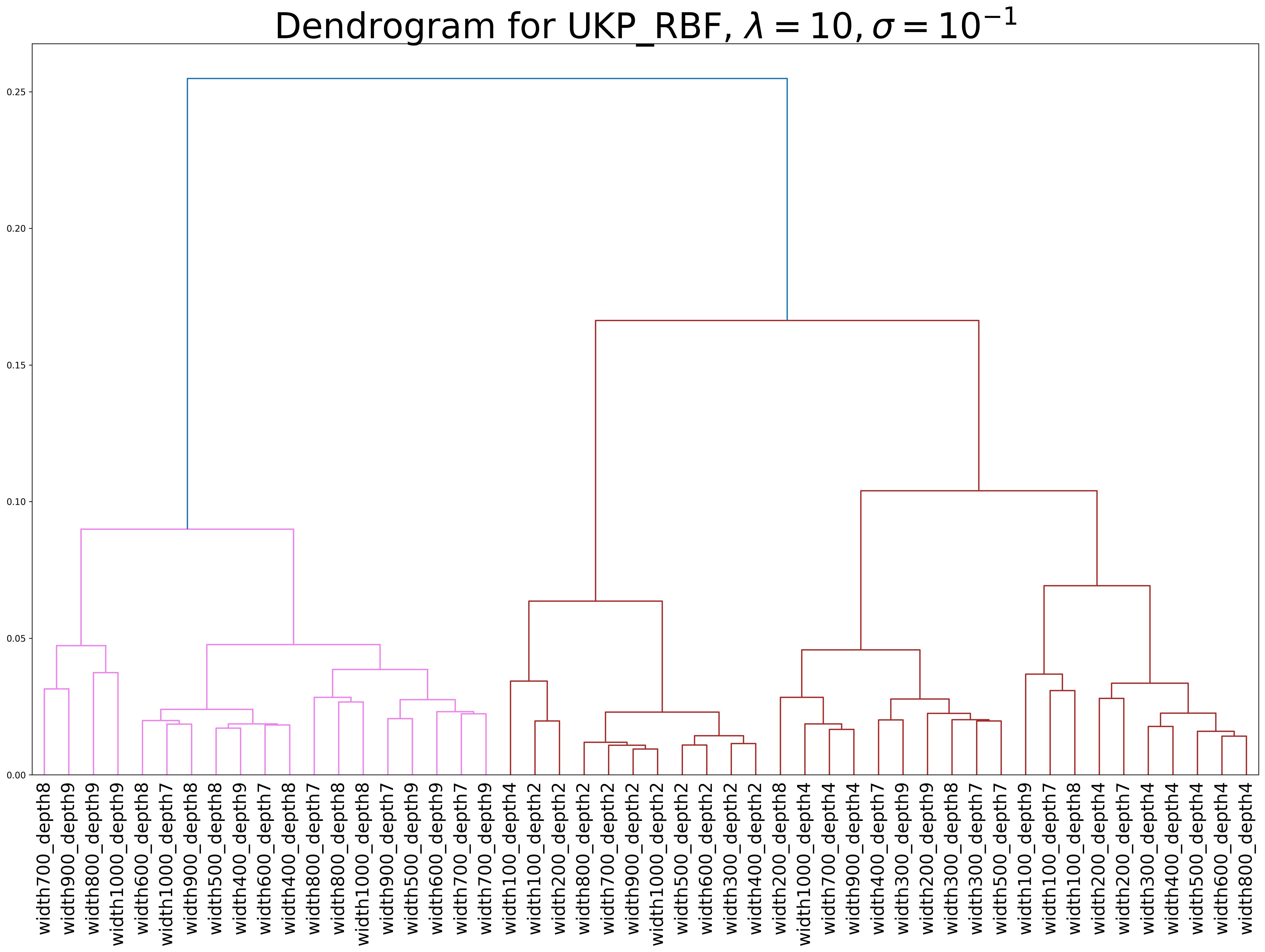}
    \end{subfigure}
    
    \caption{Dendrograms corresponding to agglomerative hierarchical clustering of representations of 50 ReLU networks based on \metricstname distance}
    \label{MNIST dendrograms}
\end{figure}

We observe in Fig. \ref{MNIST Heatmaps} that a repeating block structure emerges in each heatmap, with each block corresponding to networks with the same depth. Within each block, i.e., same depth, the pairwise similarities between networks with different widths are higher if the difference of widths of the pair of networks is small, and the similarities are lower otherwise. Further, it seems that the relative difference between networks with different depths is amplified (in terms of the \metricstname distance) if the depths of the networks are larger. For e.g. the contrast between a width 500 and width 600 network is higher when the depth is 9 for both networks, compared to the scenario where both networks have depth 2. We also perform an agglomerative (bottom-up) hierarchical clustering of the representations based on the pairwise \metricstname distances and obtain the corresponding dendrograms as shown in Fig. \ref{MNIST dendrograms}. The dendrograms also exhibit separation between deeper networks (depths 7,8 and 9) and shallow networks (depths 2,4 and 6) over a range of $(\lambda,\sigma)$ choices for the \metricstname distance with Gaussian RBF kernel. This indicates that the \metricstname distance is able to capture the relevant differences in predictive performance that are induced by architectural differences in these networks, over a wide range of values of its tuning parameters.

\raggedbottom

\paragraph{Generalization ability on kernel ridge regression tasks}

\begin{figure}[!h]
    \centering
    \begin{subfigure}[b]{0.45\textwidth}
        \includegraphics[width=0.8\textwidth]{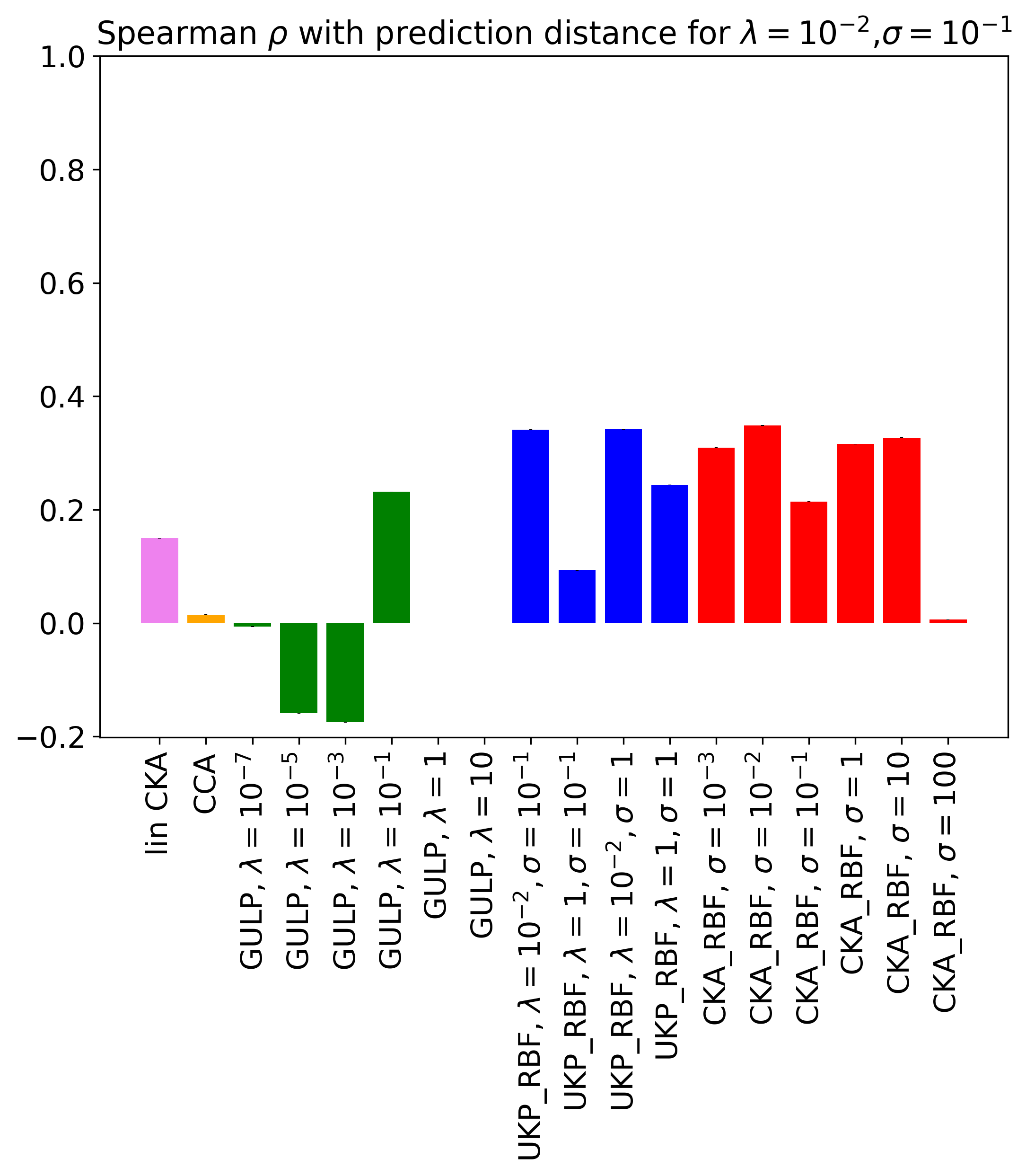}
    \end{subfigure}
    \hfill
    \begin{subfigure}[b]{0.45\textwidth}
        \includegraphics[width=0.8\textwidth]{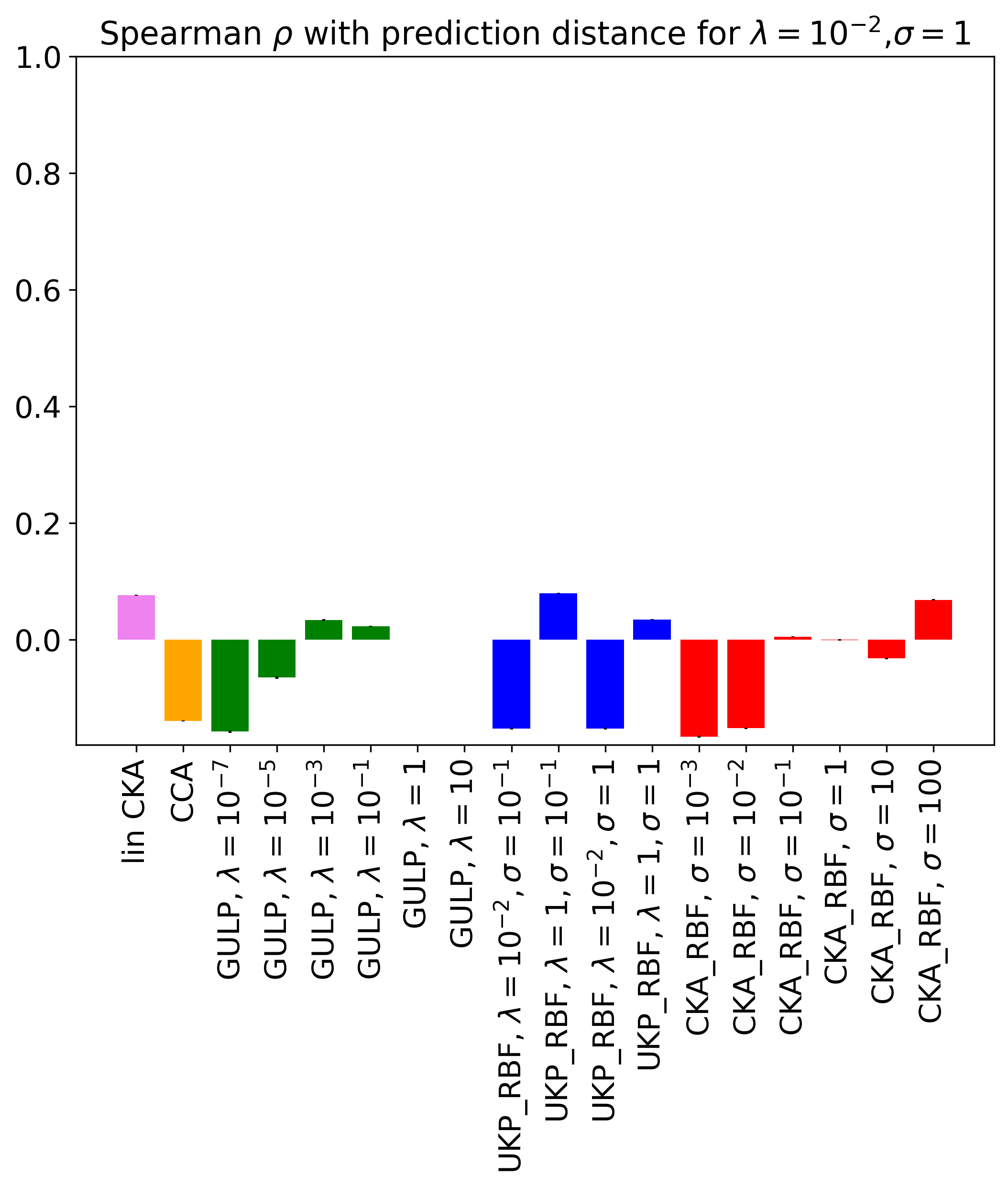}
    \end{subfigure}
    
    \vspace{0.5cm}  
    
    \begin{subfigure}[b]{0.45\textwidth}
        \includegraphics[width=0.8\textwidth]{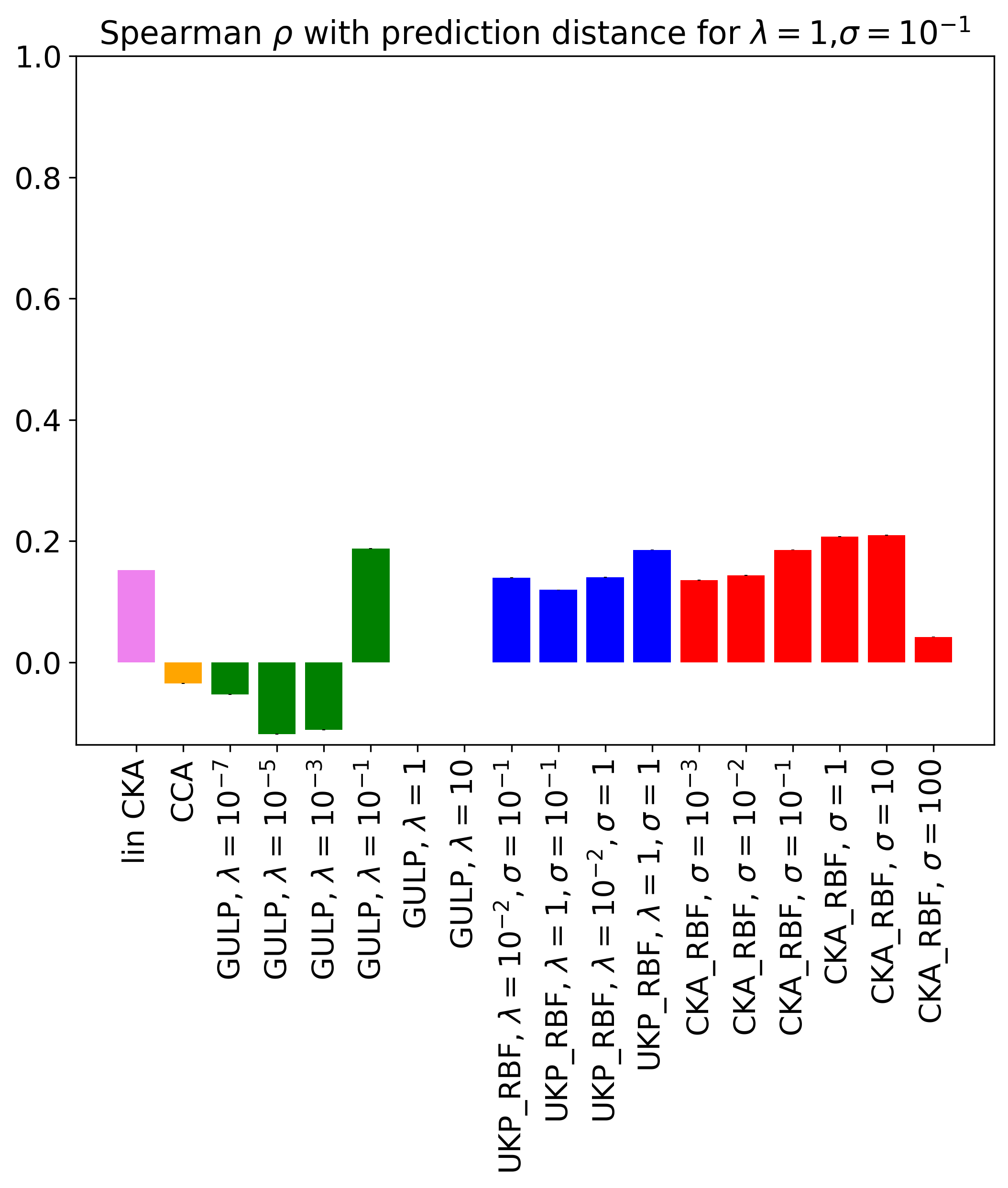}
    \end{subfigure}
    \hfill
    \begin{subfigure}[b]{0.45\textwidth}
        \includegraphics[width=0.8\textwidth]{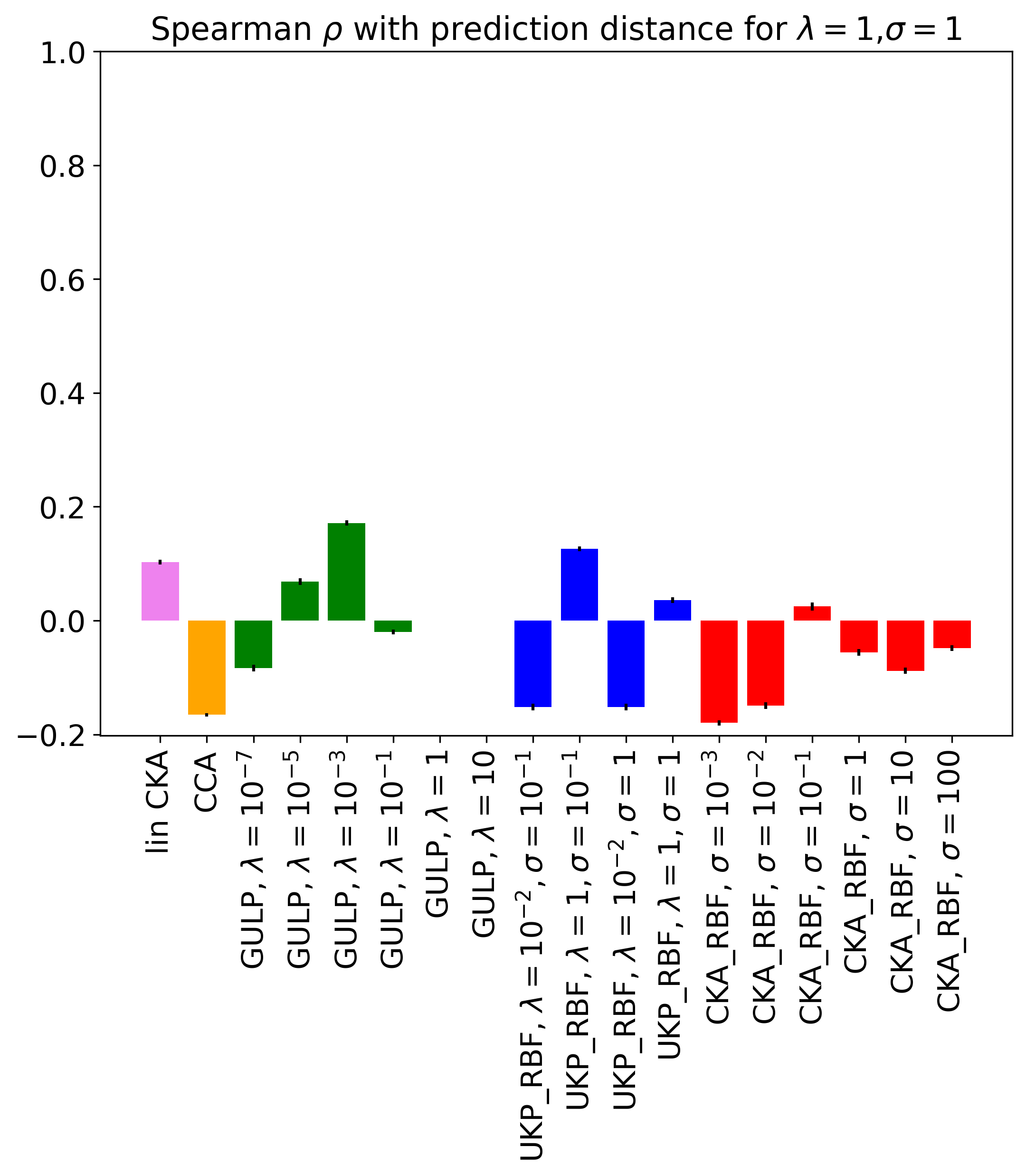}
    \end{subfigure}
    
    \caption{Spearman's $\rho$ rank correlation coefficient between generalization of kernel ridge regression-based predictors with various distance measures between representations. We report the average correlation across 10 random synthetic kernel ridge regression tasks. Results are similar for 30 trials. Error bars are negligibly small and hence not visible.}
    \label{MNIST generalization plots}
\end{figure}

We consider the same setup as discussed in Section \ref{MNIST experiments}. Supplementing our choices of $\lambda=10^{-2}$ and $\sigma=10^{-1}$ corresponding to synthetic kernel ridge regression tasks with Gaussian RBF kernel, we now consider $\lambda \in \left\{10^{-2},1\right\}$  and $\sigma \in \left\{10^{-1},1\right\}$. In Fig. \ref{MNIST generalization plots}, we plot the Spearman's $\rho$ rank correlation coefficient between the $err_{\repone,\reptwo}$'s as defined in Section \ref{MNIST experiments} and the pairwise distances between the representations using the following distances - CCA, linear CKA, nonlinear CKA with Gaussian RBF kernel, GULP and UKP with Gaussian RBF kernel. 

When $(\lambda = 10^{-2},\sigma = 10^{-1})$ and $(\lambda=1,\sigma=10^{-1})$, we observe from Fig. \ref{MNIST generalization plots} that the pairwise \metricstname distance is positively correlated to a moderate extent with the collection of $err_{\repone,\reptwo}$'s, as evident from the large positive values of the blue bars. In contrast, GULP distances show inconsistent behavior across different levels of regularization, while CCA and linear CKA distances show a much lower positive correlation with generalization performance (with CCA even showing negative correlation when $(\lambda=1,\sigma=10^{-1})$). For the remaining choices, none of the distance measures show any consistent behavior, which indicates that an increase in the number of samples used to approximate the model representations may improve the performance of these distance measures.

\raggedbottom

\begin{figure}[!h]
    \centering
    \begin{subfigure}[b]{0.45\textwidth}
        \includegraphics[width=\textwidth]{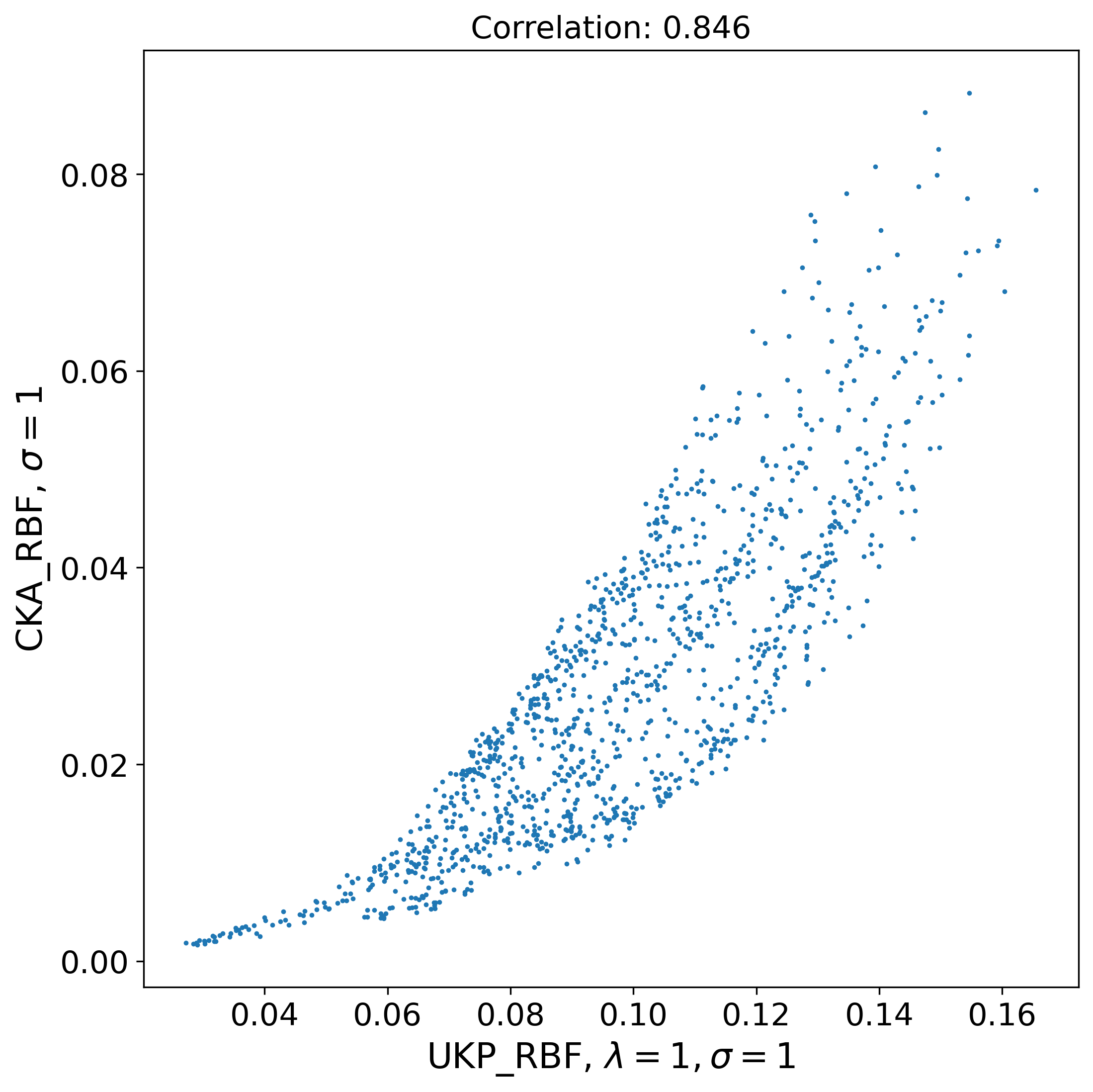}
    \end{subfigure}
    \hfill
    \begin{subfigure}[b]{0.45\textwidth}
        \includegraphics[width=\textwidth]{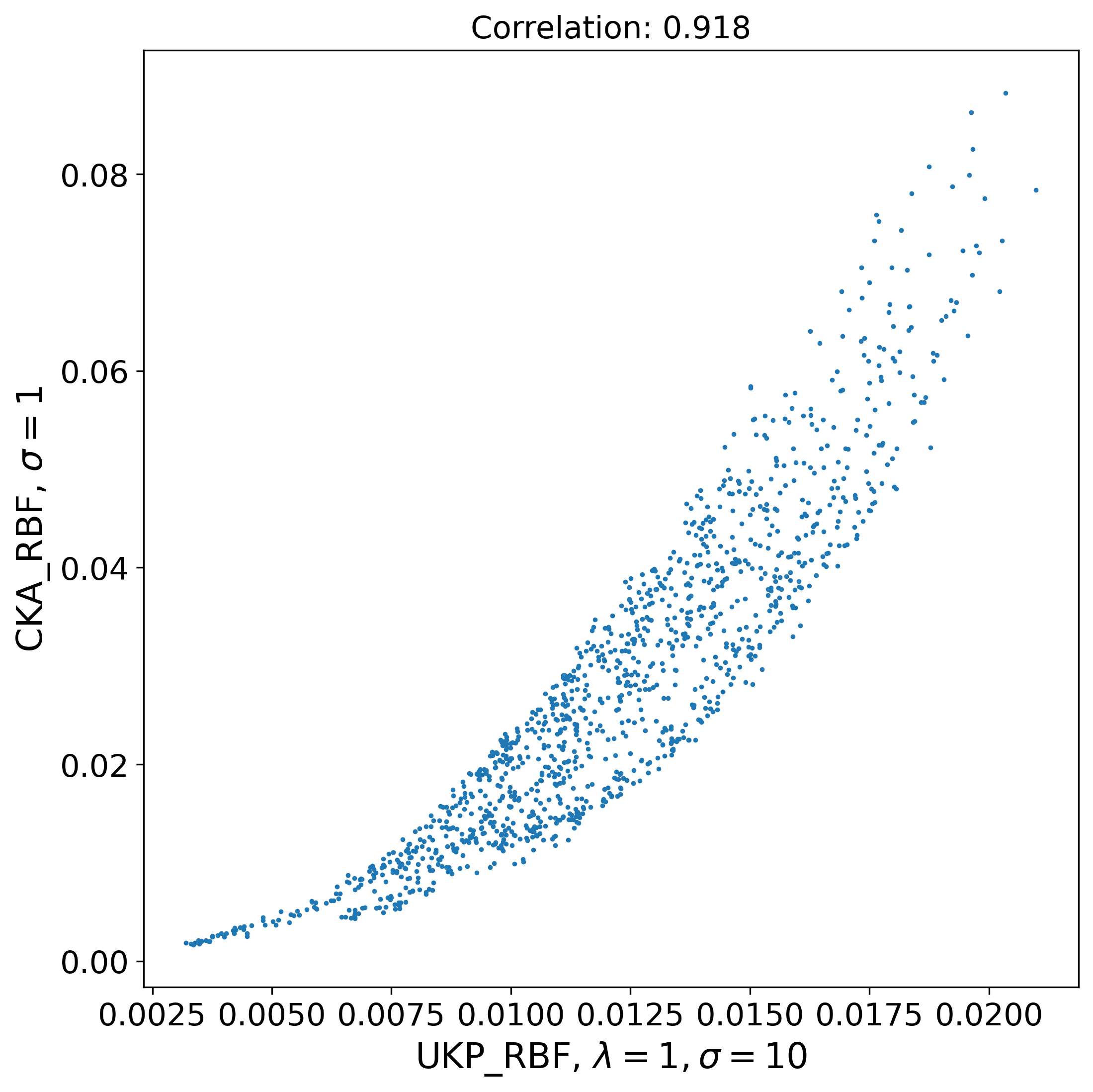}
    \end{subfigure}
    
    \vspace{0.5cm}  
    
    \begin{subfigure}[b]{0.45\textwidth}
        \includegraphics[width=\textwidth]{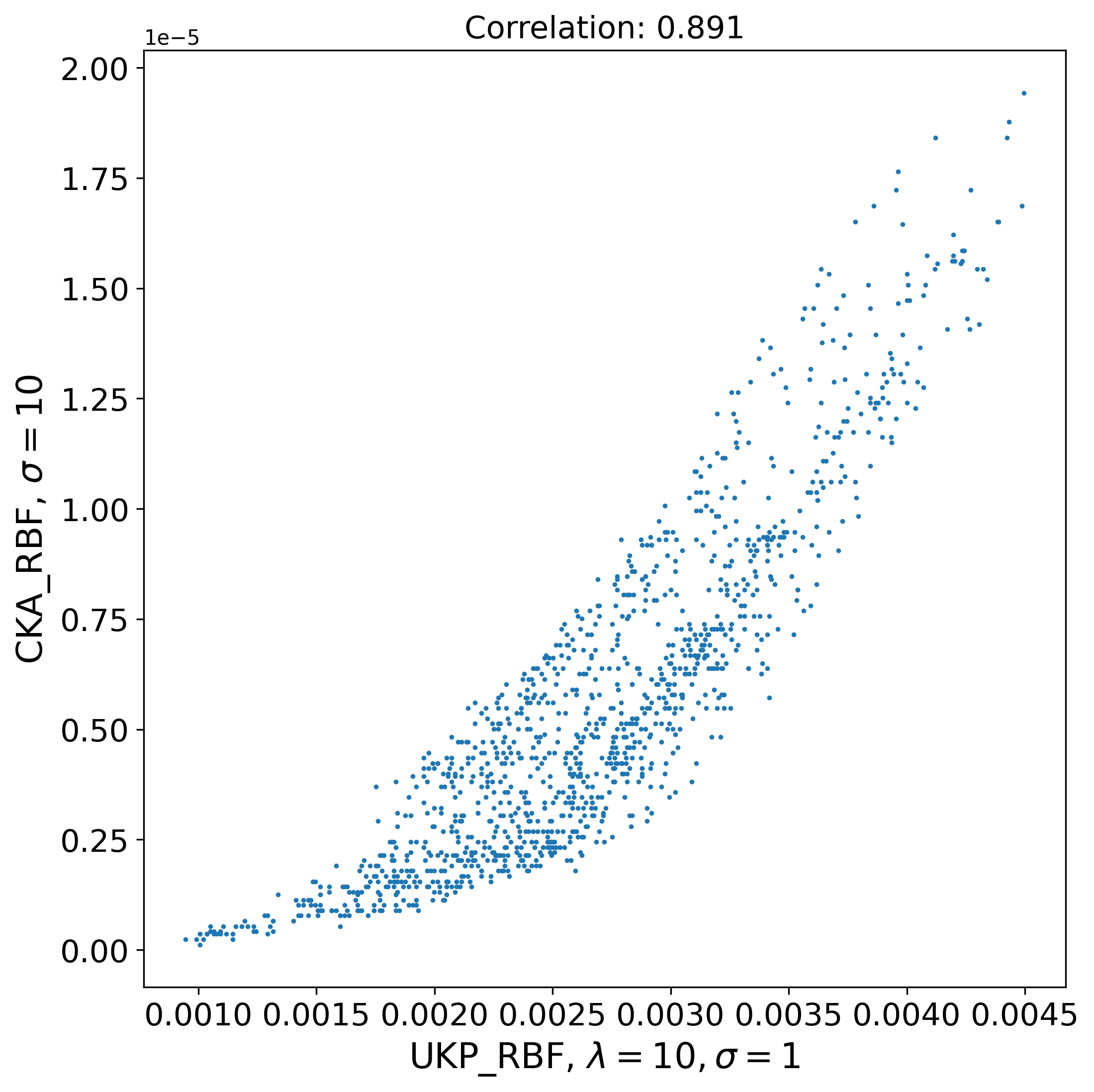}
    \end{subfigure}
    \hfill
    \begin{subfigure}[b]{0.45\textwidth}
        \includegraphics[width=\textwidth]{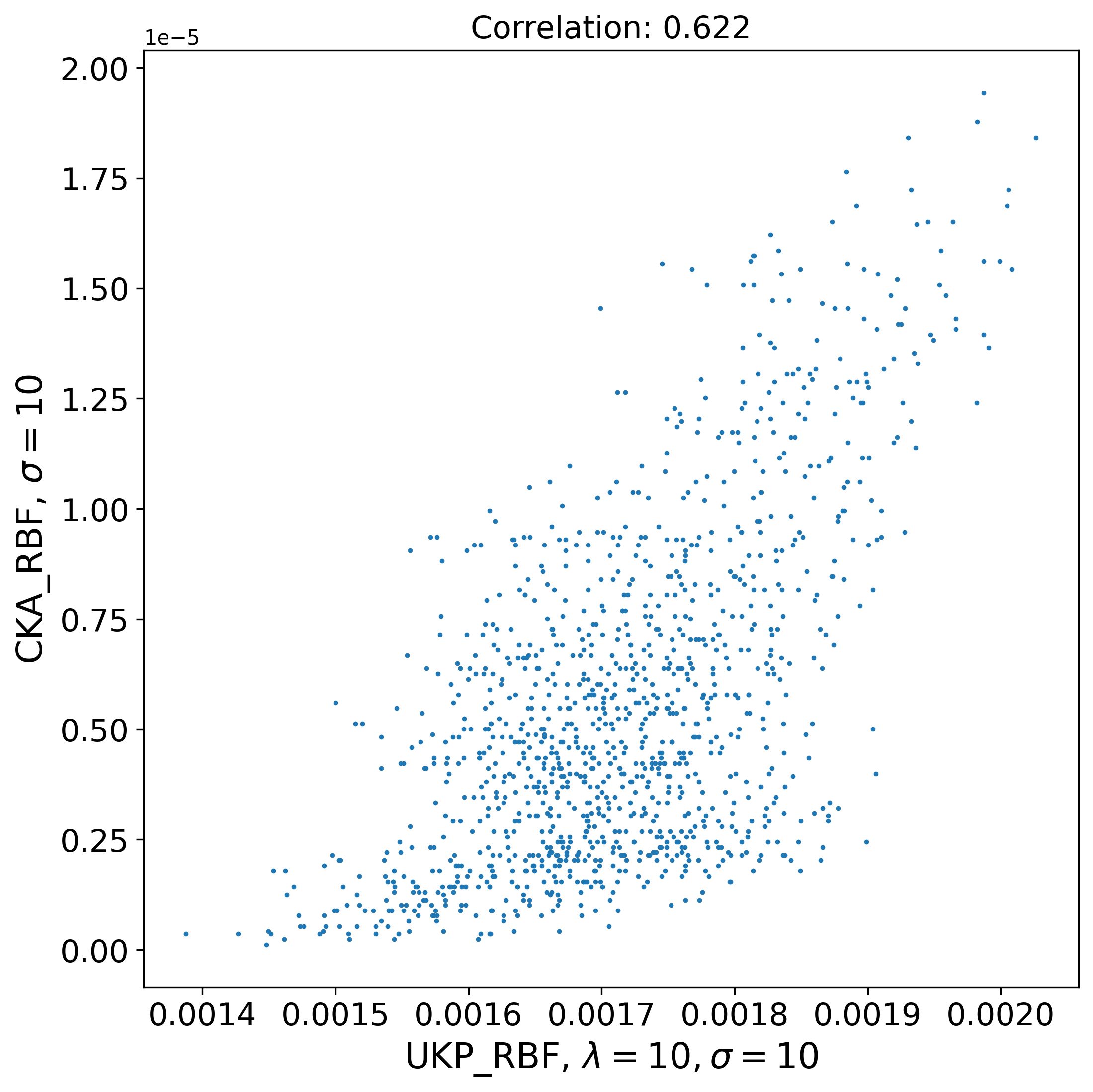}
    \end{subfigure}
    
    \caption{Correlation plots between UKP and CKA measures with Gaussian RBF kernel between $\binom{50}{2}$ pairs of ReLU networks trained on MNIST data. Plot titles display the Pearson product-moment correlation coefficient between the distance measures on the two axes.}

    \label{MNIST correlation plots bw UKP CKA}
\end{figure}

Unsurprisingly, as a consequence of the relationship between CKA and \metricstname, as discussed in Section \ref{Relation to other comparison measures}, the performance of the CKA distance, when using the Gaussian RBF kernel (with the corresponding bars shown in red), is comparable to that of \metricstname with the same choice of kernel. This similarity in the information conveyed by these two measures can be empirically observed through their scatterplots and the Pearson product-moment correlation coefficient under various choices of tuning parameters. As shown in Fig. \ref{MNIST correlation plots bw UKP CKA}, the nearly linear positive relationship between \metricstname and CKA distances, when both are used with a Gaussian RBF kernel, along with the high positive correlation coefficient, suggests that either measure could be effectively used in practice for comparing representations. However, the \metricstname distance may be preferred over the CKA distance due to its pseudometric properties, particularly the triangle inequality, which proves to be especially useful. In contrast, CKA, being a measure akin to a normalized inner product bounded between 0 and 1, does not satisfy the properties of a pseudometric and may lead to misleading intuitions when comparing different representations.

\raggedbottom

\FloatBarrier

\subsection{ImageNet experiments}
\label{Additional ImageNet experiments}

\paragraph{Architectures used and data description} In our experiments, we utilized 35 pretrained models known for achieving state-of-the-art (SOTA) performance in the ImageNet Object Localization Challenge on Kaggle \cite{imagenet-object-localization-challenge}, available from \citet{pytorch}. These models are categorized based on their architectural types as follows:

\begin{itemize}
    \item \textbf{ResNets (17 models):} regnet\_x\_16gf, regnet\_x\_1\_6gf, regnet\_x\_32gf, regnet\_x\_3\_2gf, regnet\_x\_400mf, regnet\_x\_800mf, regnet\_x\_8gf, regnet\_y\_16gf, regnet\_y\_1\_6gf, regnet\_y\_32gf, regnet\_y\_3\_2gf, regnet\_y\_400mf, regnet\_y\_800mf, regnet\_y\_8gf, resnet18, resnext50\_32x4d, wide\_resnet50\_2
    \item \textbf{EfficientNets (8 models):} efficientnet\_b0, efficientnet\_b1, efficientnet\_b2, efficientnet\_b3, efficientnet\_b4, efficientnet\_b5, efficientnet\_b6, efficientnet\_b7
    \item \textbf{MobileNets (3 models):} mobilenet\_v2, mobilenet\_v3\_large, mobilenet\_v3\_small
    \item \textbf{ConvNeXts (2 models):} convnext\_small, convnext\_tiny
    \item \textbf{Other Architectures (5 models):} alexnet, googlenet, inception, mnasnet, vgg16 .
\end{itemize}

The penultimate layer dimensions for these networks, corresponding to the representation sizes, vary from 400 to 4096 depending on the architecture. Each model processes input data as 3-channel RGB images, with each channel having dimensions of 224 × 224 pixels. To approximate the model representations learned by these models using finite-dimensional representations, we used 3000 images from the validation set of the ImageNet dataset. These images were normalized with a mean of (0.485, 0.456, 0.406) and a standard deviation of (0.229, 0.224, 0.225) for each RGB channel. Our choice of models and input preprocessing parameters is similar to those used in \citet{GULP}.

\begin{figure}[!h]
    \centering
    \begin{subfigure}[b]{0.3\textwidth}
        \includegraphics[width=\textwidth]{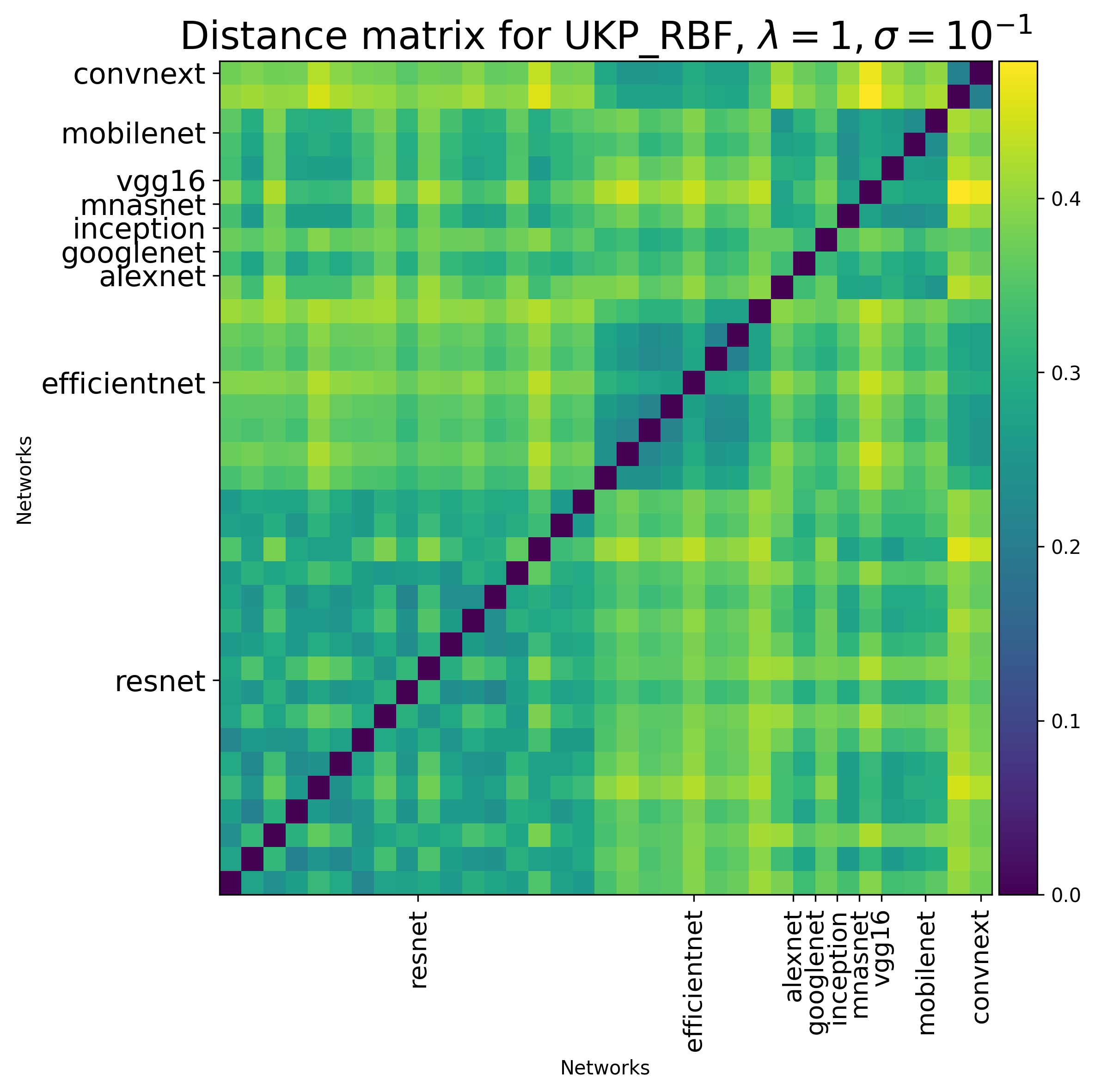}
    \end{subfigure}
    \hfill
    \begin{subfigure}[b]{0.3\textwidth}
        \includegraphics[width=\textwidth]{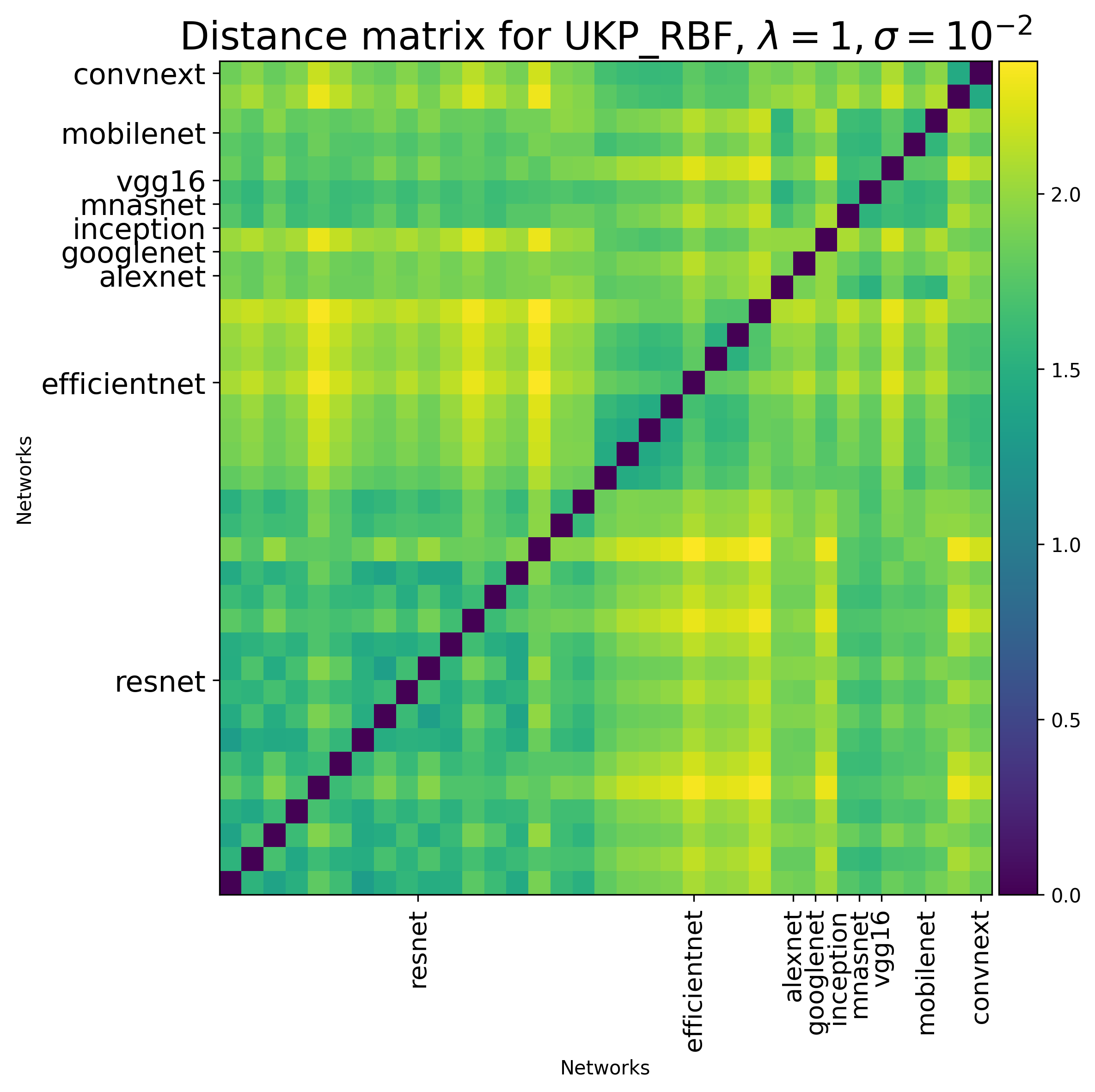}
    \end{subfigure}
    \hfill
    \begin{subfigure}[b]{0.3\textwidth}
        \includegraphics[width=\textwidth]{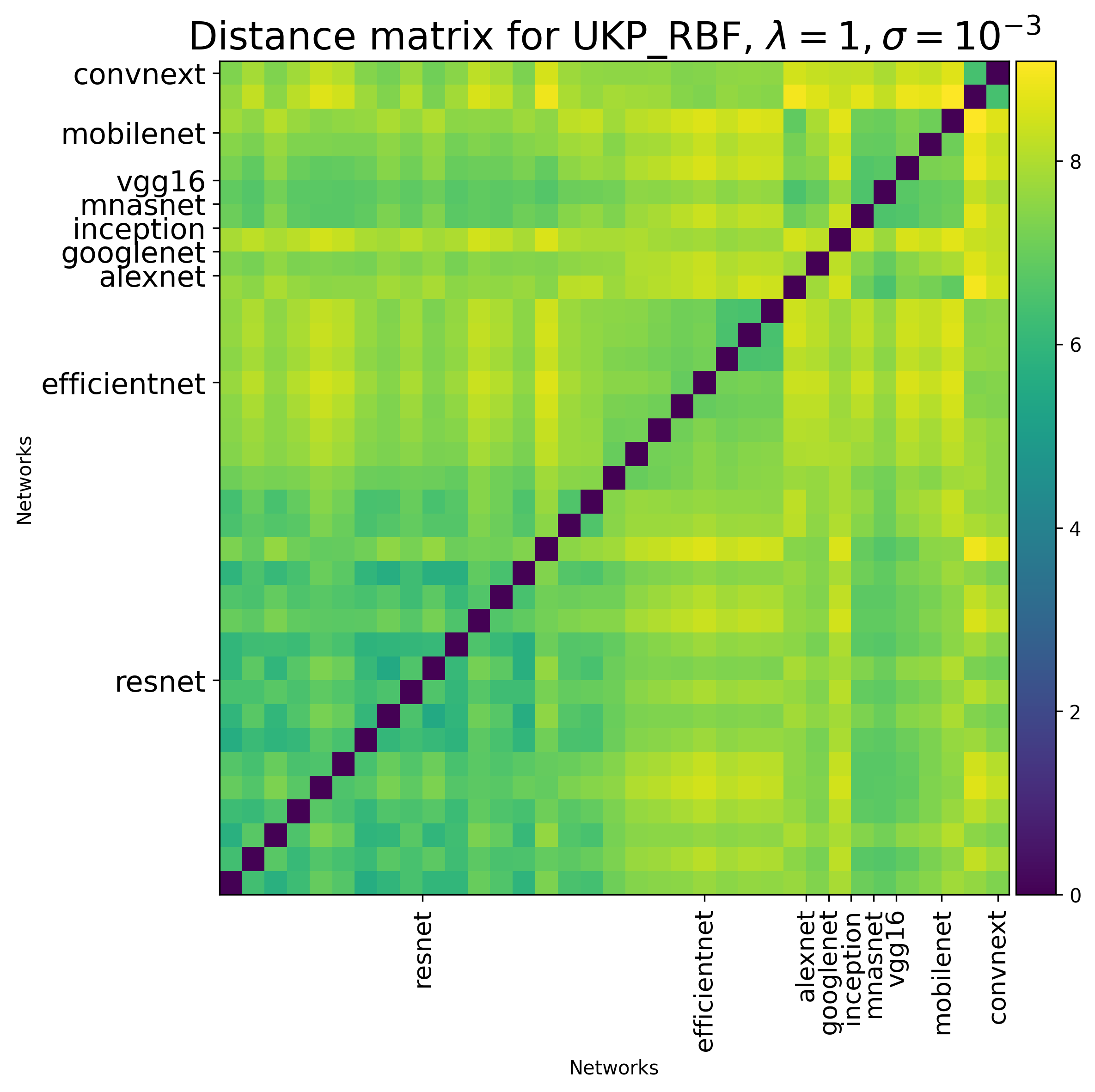}
    \end{subfigure}
    
    \vspace{0.5cm}  
    
    \begin{subfigure}[b]{0.3\textwidth}
        \includegraphics[width=\textwidth]{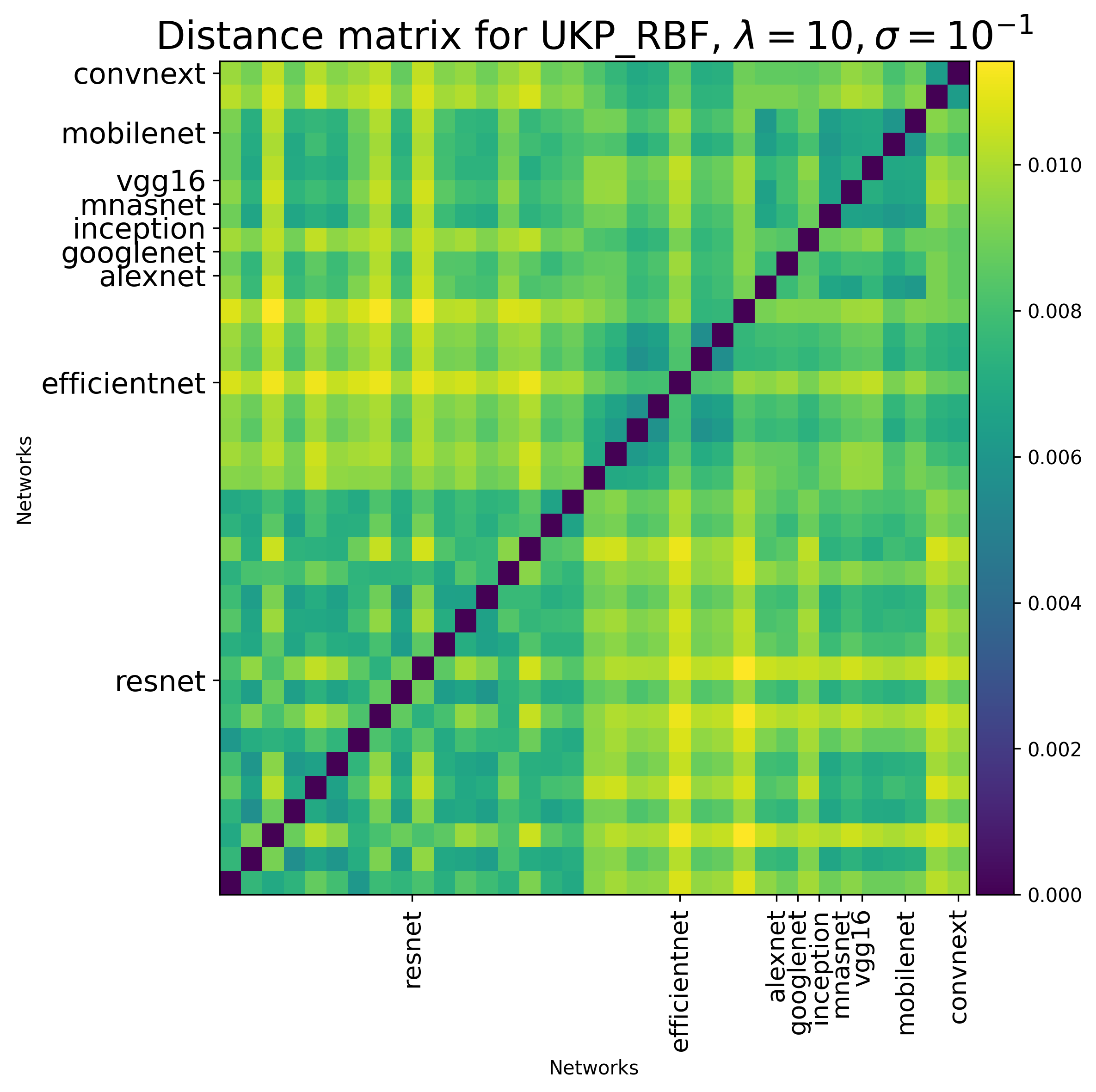}
    \end{subfigure}
    \hfill
    \begin{subfigure}[b]{0.3\textwidth}
        \includegraphics[width=\textwidth]{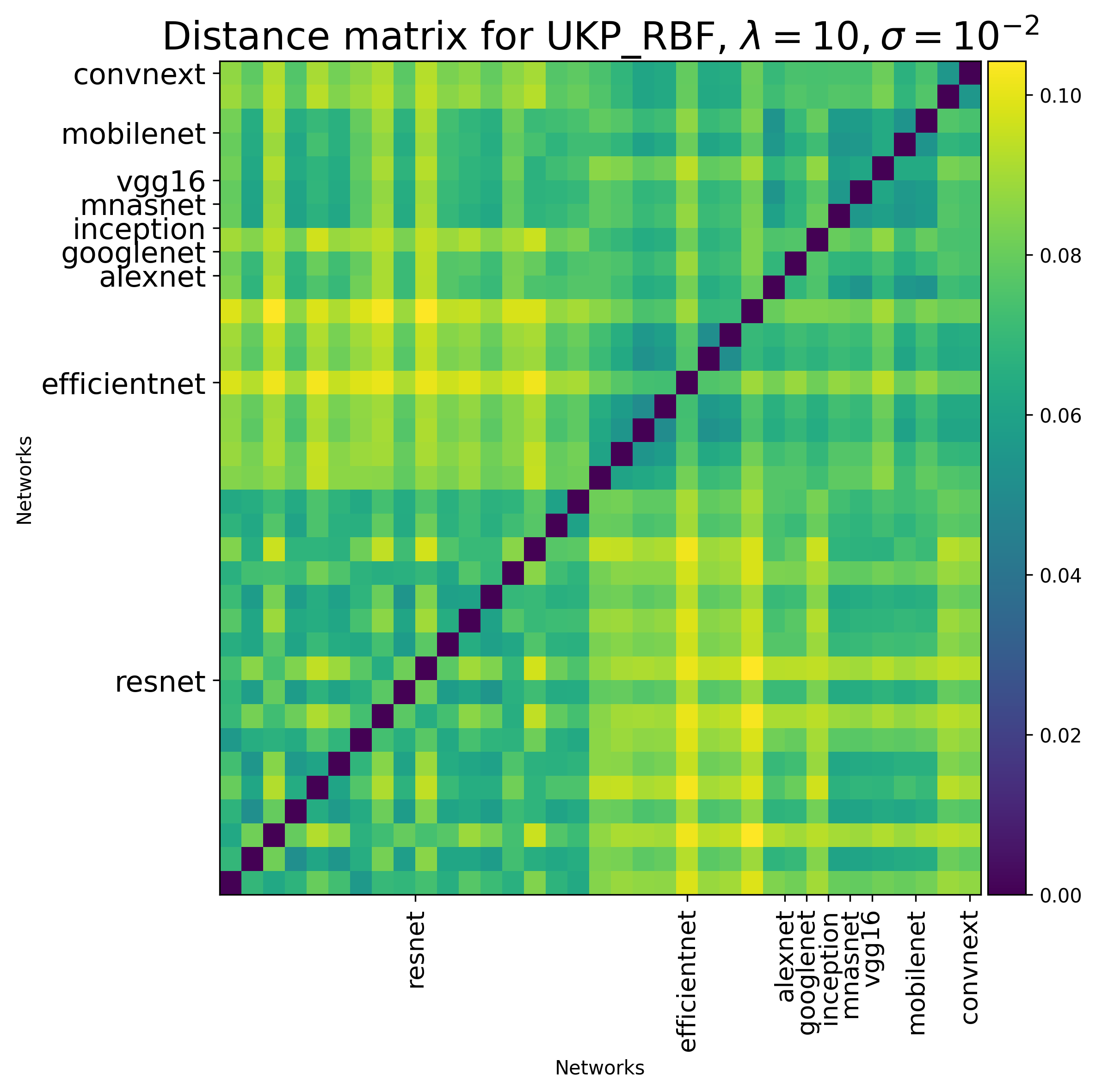}
    \end{subfigure}
    \hfill
    \begin{subfigure}[b]{0.3\textwidth}
        \includegraphics[width=\textwidth]{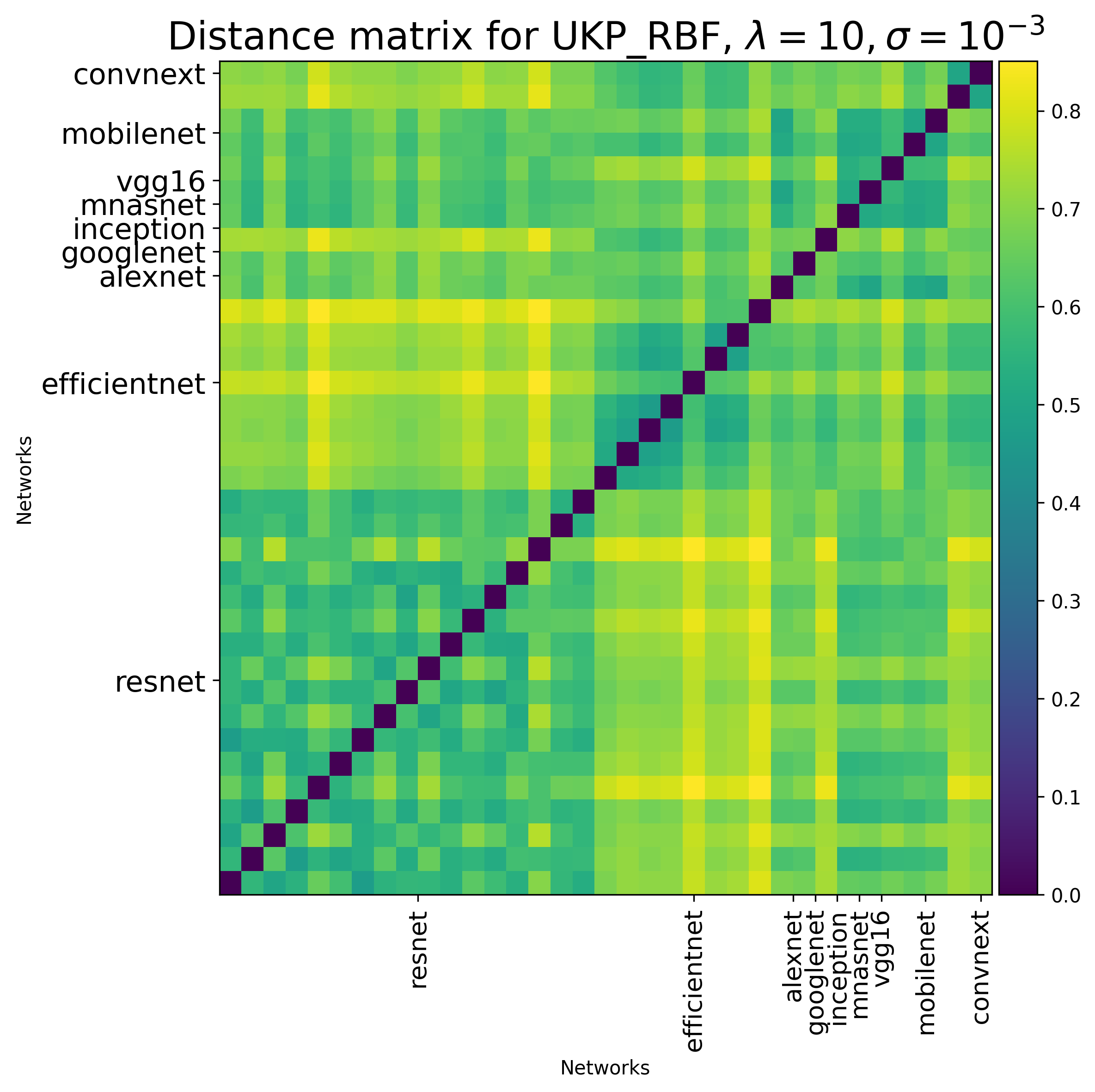}
    \end{subfigure}
    
     \caption{Heatmaps representing \metricstname distance between pairs of networks of different architecture, pretrained on ImageNet data. We choose the kernel for the \metricstname distance to be the Gaussian RBF kernel with bandwidth $\sigma \in \left\{10^{-1},10^{-2},10^{-3}\right\}$ along with the regularization parameter $\lambda \in \left\{1,10\right\}$. Along the rows and columns of each of the heatmaps, the networks are arranged in the following order from left to right and top to bottom - ResNets, EfficientNets, Other Architectures, MobileNets and ConvNexts. Darker colors indicate smaller value of \metricstname distance according to the scale attached to each heatmap.}
    \label{ImageNet heatmaps}
\end{figure}

\paragraph{Clustering of representations based on UKP aligns with architectural characteristics of networks}

We are interested in observing whether the \metricstname pseudometric is capable of capturing intrinsic differences in predictive performances of different representations. Such intrinsic differences are often the result of the different inductive biases we encode into networks through the choice of architectures, among other factors. 

We first discuss the main architectural similarities and differences between ResNet, RegNet, EfficientNet, MobileNet, alexnet, googlenet, inception, mnasnet, and vgg16, which are controlled by how they address depth, efficiency, and feature extraction. Alexnet and vgg16 are older architectures that use standard convolutional layers arranged in sequential blocks, with vgg16 deepening the network significantly compared to alexnet. Googlenet introduced Inception modules, which combine multiple convolution filters of different sizes to capture multi-scale features, making it more efficient than alexNet and vgg16. Different Inception architectures have been built using the Inception module of Googlenet. ResNet brought the innovation of residual connections (skip connections) to address the vanishing gradient problem, enabling very deep networks, while RegNet refined this concept by creating more regular, scalable structures without explicit skip connections. EfficientNet and mnasnet focus on balanced scaling (depth, width, resolution) and use of MBConv blocks for efficiency, with EfficientNet employing a compound scaling formula. MobileNet, like mnasnet, emphasizes depthwise separable convolutions for lightweight, efficient models suitable for mobile devices. In terms of architectural similarities, resNet and regNet share a focus on structured deep architectures, while EfficientNet and MobileNet share efficiency-driven designs for varied hardware constraints. Alexnet, vgg16, and googlenet represent early convolutional architectures, with googlenet’s Inception modules providing a bridge to more modern designs. In contrast, vgg16 and ResNet are quite different, with vgg16 being sequential and deep, and ResNet leveraging residual connections.

\begin{figure}[!h]
    \centering
    \begin{subfigure}[b]{0.45\textwidth}
        \includegraphics[width=\textwidth]{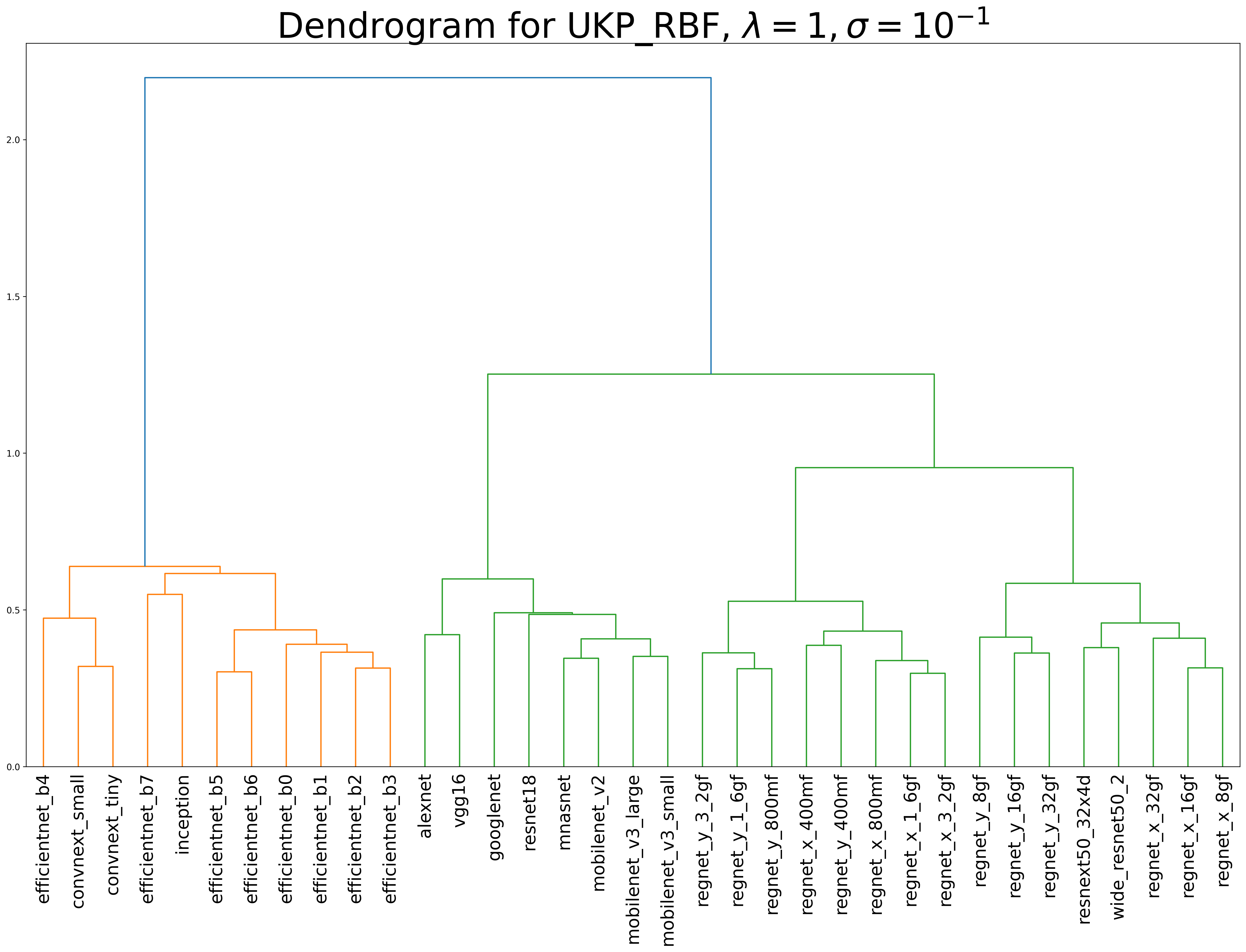}
    \end{subfigure}
    \hfill
    \begin{subfigure}[b]{0.45\textwidth}
        \includegraphics[width=\textwidth]{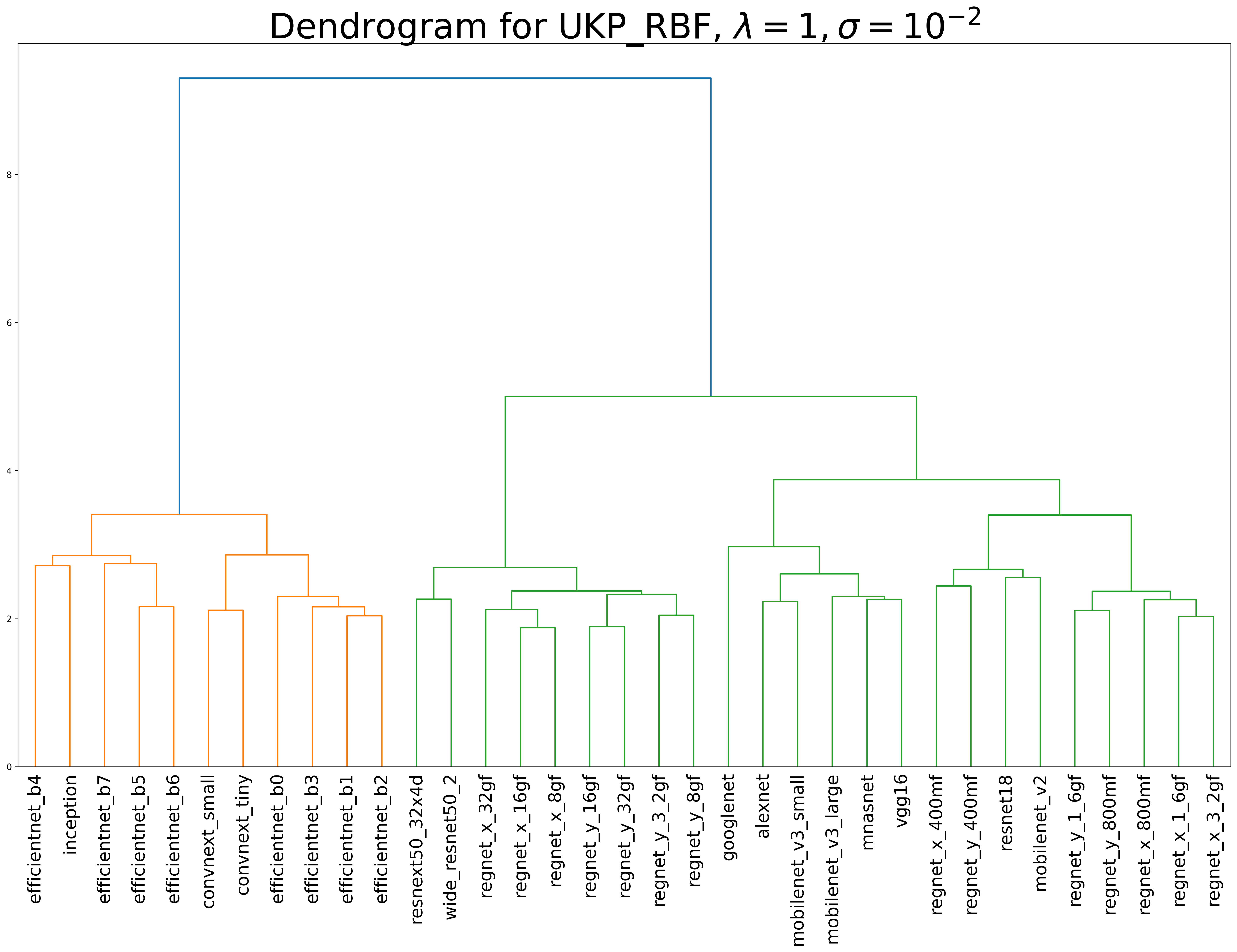}
    \end{subfigure}
    
    \vspace{0.5cm}  
    
    \begin{subfigure}[b]{0.45\textwidth}
        \includegraphics[width=\textwidth]{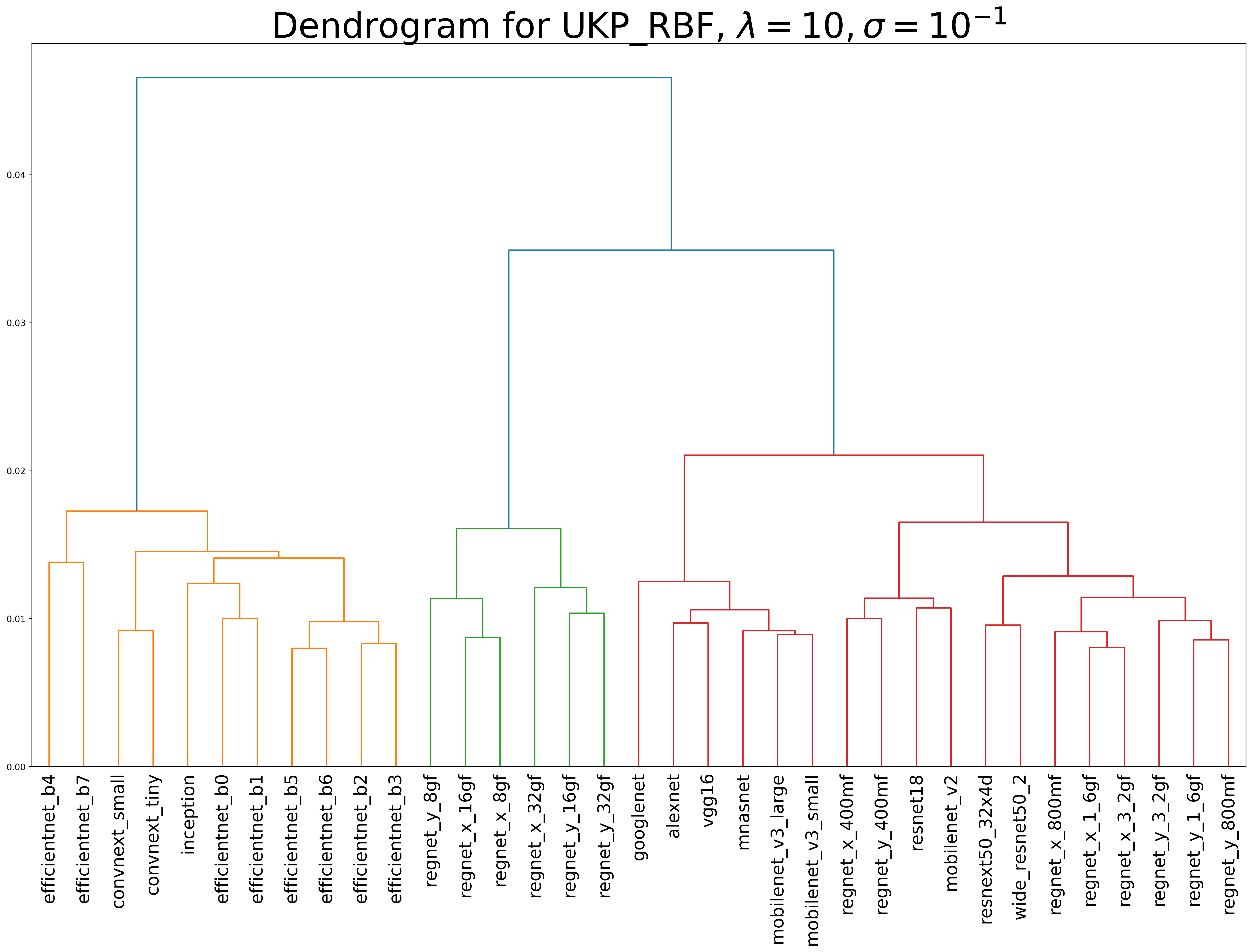}
    \end{subfigure}
    \hfill
    \begin{subfigure}[b]{0.45\textwidth}
        \includegraphics[width=\textwidth]{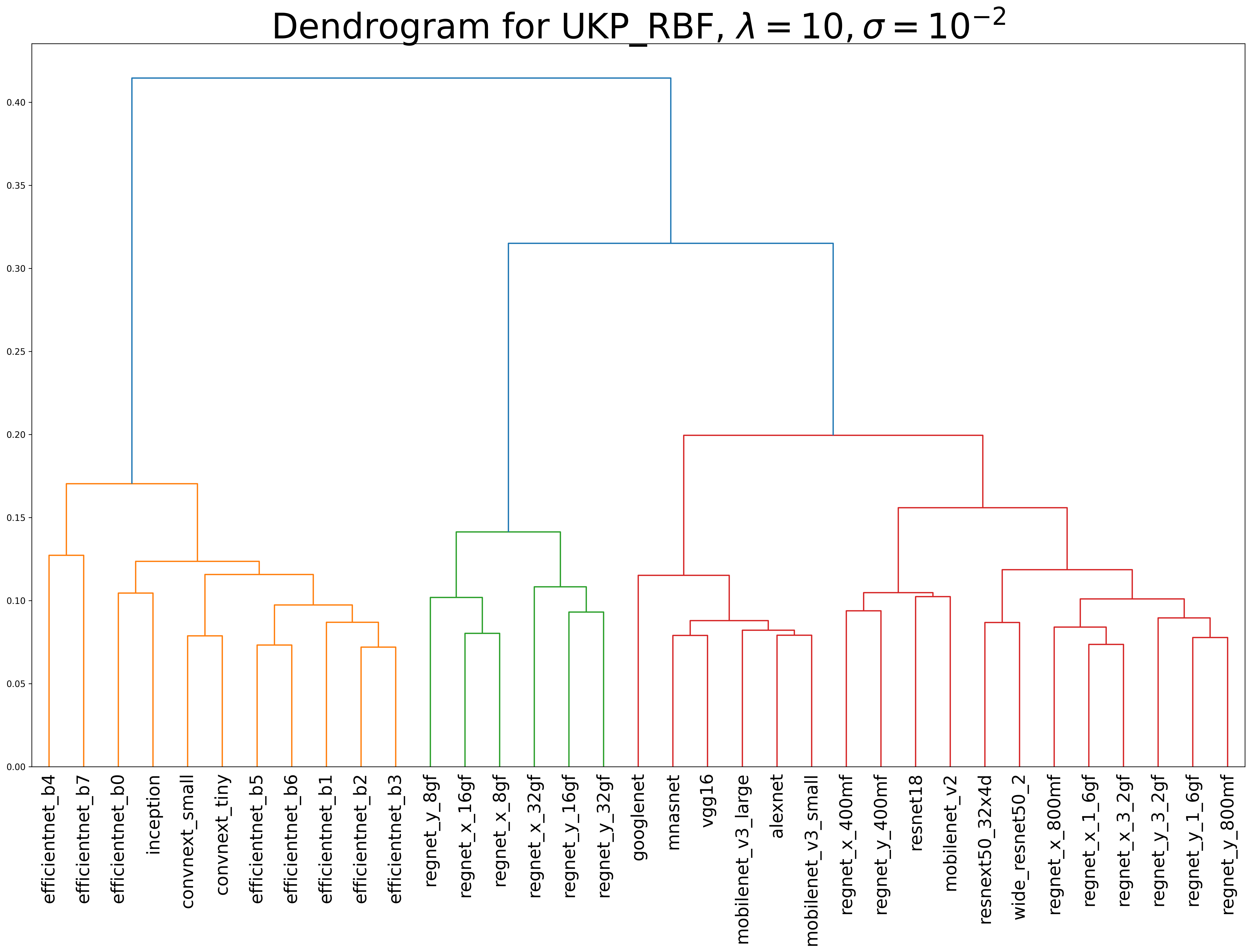}
    \end{subfigure}
    
    \caption{Dendrograms corresponding to agglomerative hierarchical clustering of representations of 35 pretrained ImageNet networks based on \metricstname distance}
    \label{ImageNet dendrograms}
\end{figure}

We observe in Fig. \ref{ImageNet heatmaps} that a block structure emerges in the heatmaps across different choices of the tuning parameters for the \metricstname distance, especially corresponding to the 4 major groups of architectures ResNets, EfficientNets, MobileNets and ConvNeXts. We also perform an agglomerative (bottom-up) hierarchical clustering of the representations based on the pairwise \metricstname distances and obtain the corresponding dendrograms as shown in Fig. \ref{ImageNet dendrograms}. The dendrograms exhibit a clear separation between the ResNets/RegNets and the remaining architectures over a range of $(\lambda,\sigma)$ choices for the \metricstname distance with Gaussian RBF kernel. This indicates that, for the class of pretrained ImageNet models we consider, the \metricstname distance captures the relevant differences in predictive performance that are induced by architectural differences in these networks, over a wide range of values of its tuning parameters.

\begin{figure}[!h]
    \centering

    \hspace*{\fill}
    \begin{subfigure}[b]{0.45\textwidth}
        \includegraphics[width=\textwidth]{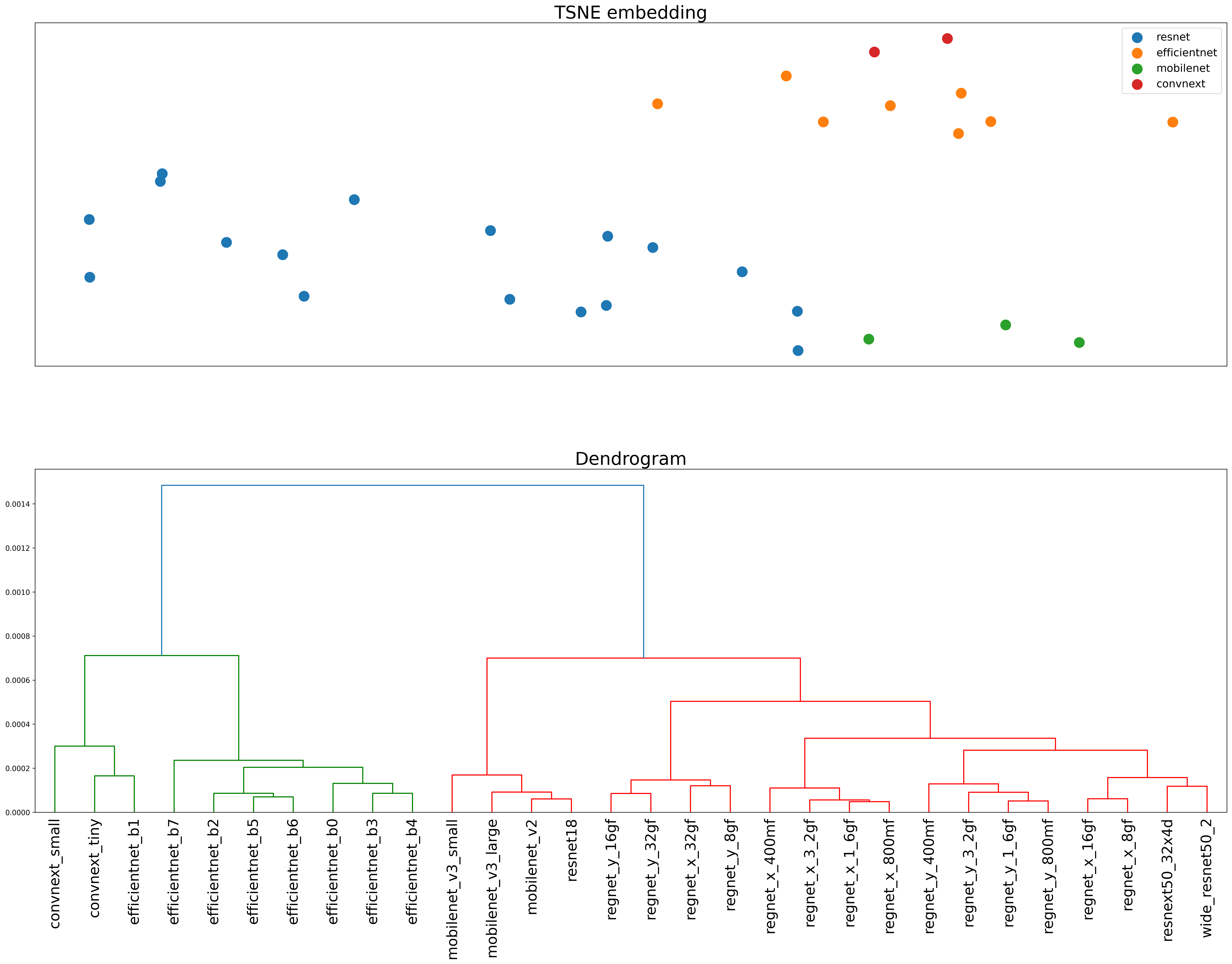}
    \subcaption{UKP (with RBF kernel, regularization parameter $\lambda=1$ and kernel bandwidth $\sigma=10$)}
    \end{subfigure}
    \hspace*{\fill}

    \vspace{0.5cm}  
    
    \begin{subfigure}[b]{0.45\textwidth}
        \includegraphics[width=\textwidth]{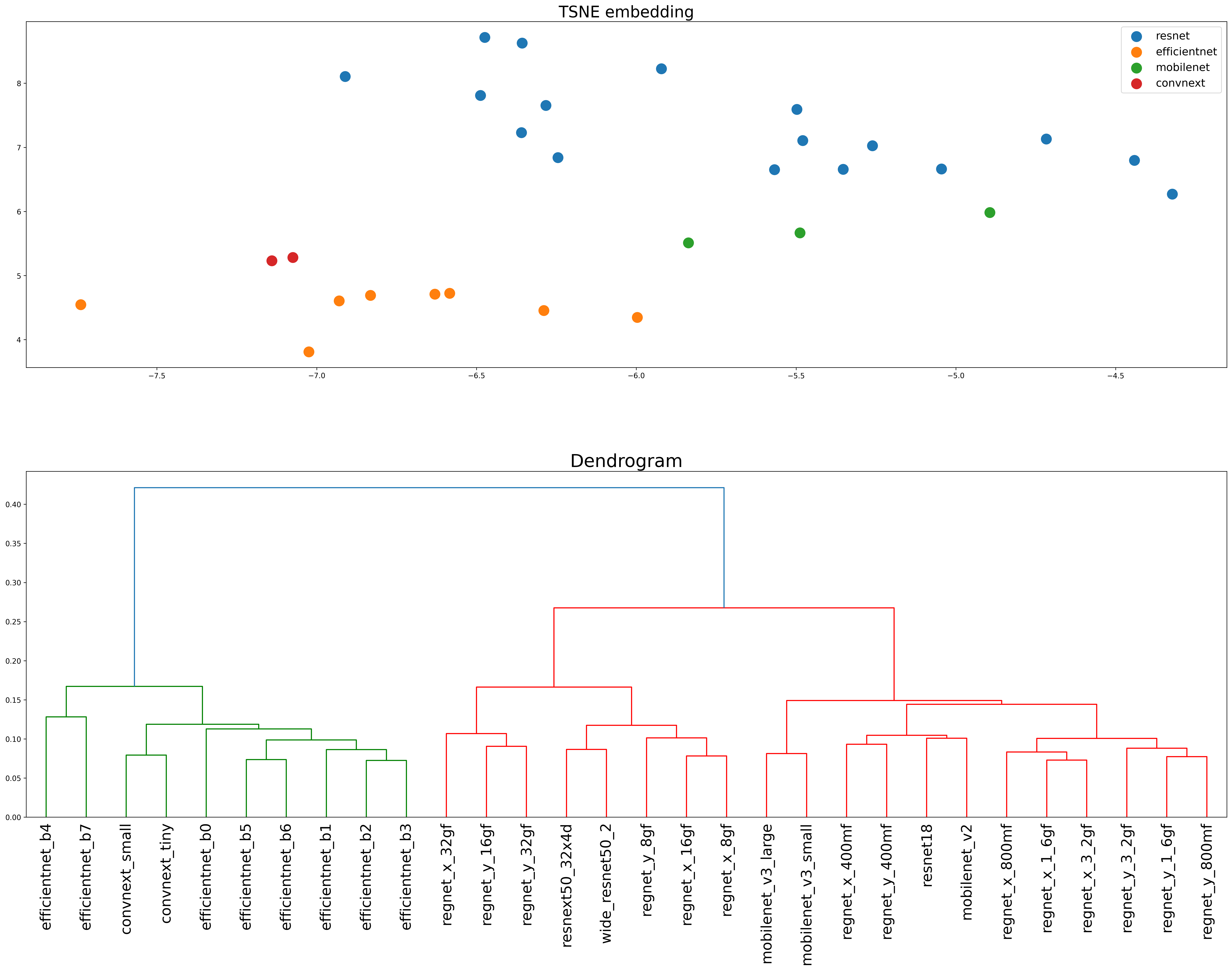}
    \subcaption{GULP (with regularization parameter $\lambda=1$)}
    \end{subfigure}
    \hfill
    \begin{subfigure}[b]{0.45\textwidth}
        \includegraphics[width=\textwidth]{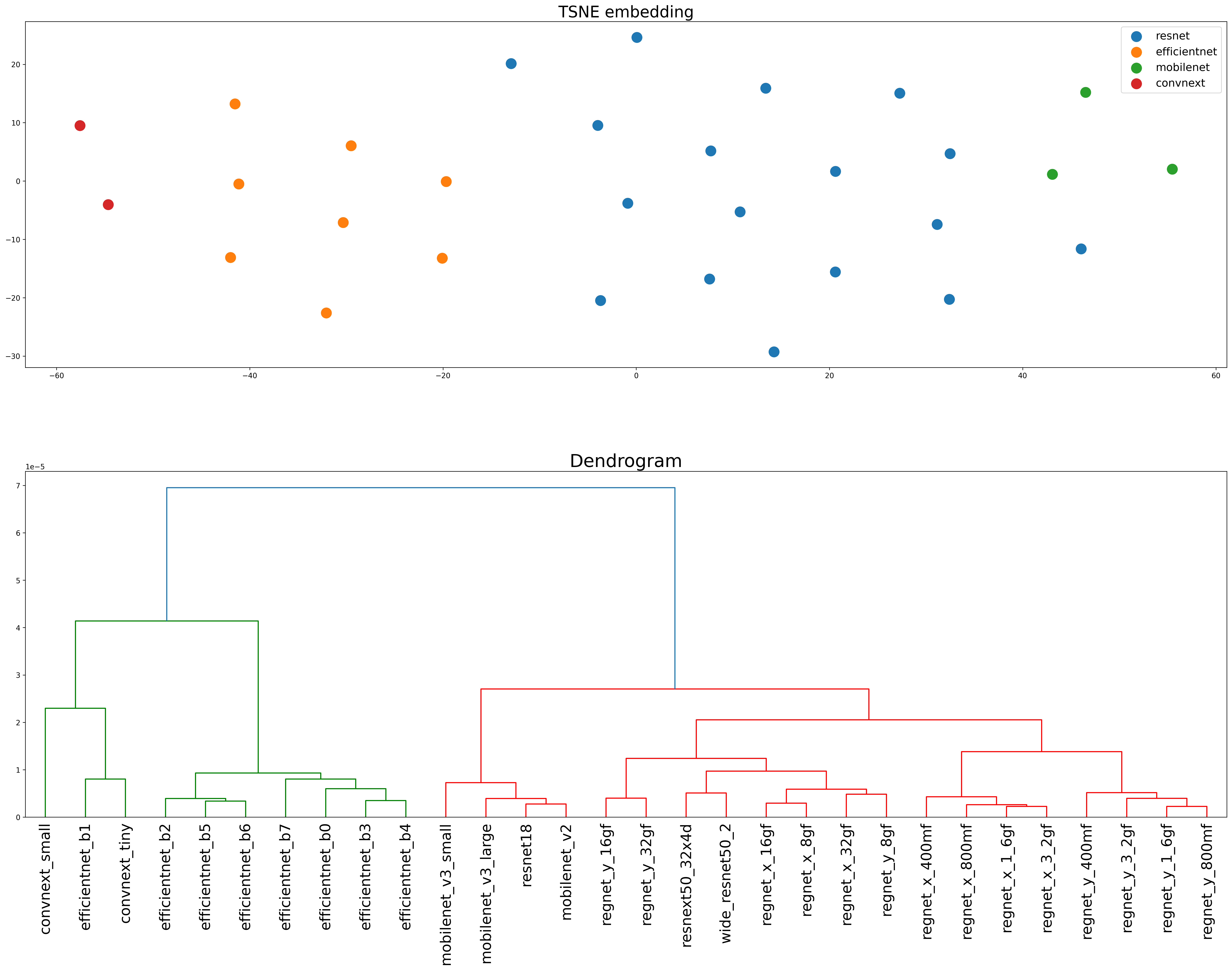}
        \subcaption{CKA (with RBF kernel and kernel bandwidth $\sigma=10$)}
    \end{subfigure}

     \caption{tSNE embeddings and dendrograms corresponding to agglomerative hierarchical clustering of representations of 35 pretrained ImageNet networks based on \metricstname (with Gaussian RBF kernel, regularization parameter $\lambda=1$ and kernel bandwidth $\sigma=10$), GULP (with regularization parameter $\lambda=1$) and CKA (with Gaussian RBF kernel and kernel bandwidth $\sigma=10$) distance}
    \label{ImageNet dendrograms additional}
\end{figure}

To illustrate that the performance of the \metricstname pseudometric is reasonably robust to the choice of the regularization parameter $\lambda$ and kernel parameters (such as bandwidth parameter $\sigma$ for the Gaussian RBF kernel), we have compared the performance of \metricstname's performance with other popular baseline measures such as GULP and CKA. As observed from Fig. \ref{ImageNet dendrograms additional}, the separation between the different classes of networks is more pronounced in the case of \metricstname than GULP. Additionally, the clustering behaviour within the primary classes of networks is much weaker for the CKA compared to the \metricstname and GULP measures, and the separation between the different classes is not clear in the case of CKA.

\paragraph{Relationship between UKP and CKA measures}

\begin{figure}[h!]
    \centering
    \begin{subfigure}[b]{0.45\textwidth}
        \includegraphics[width=\textwidth]{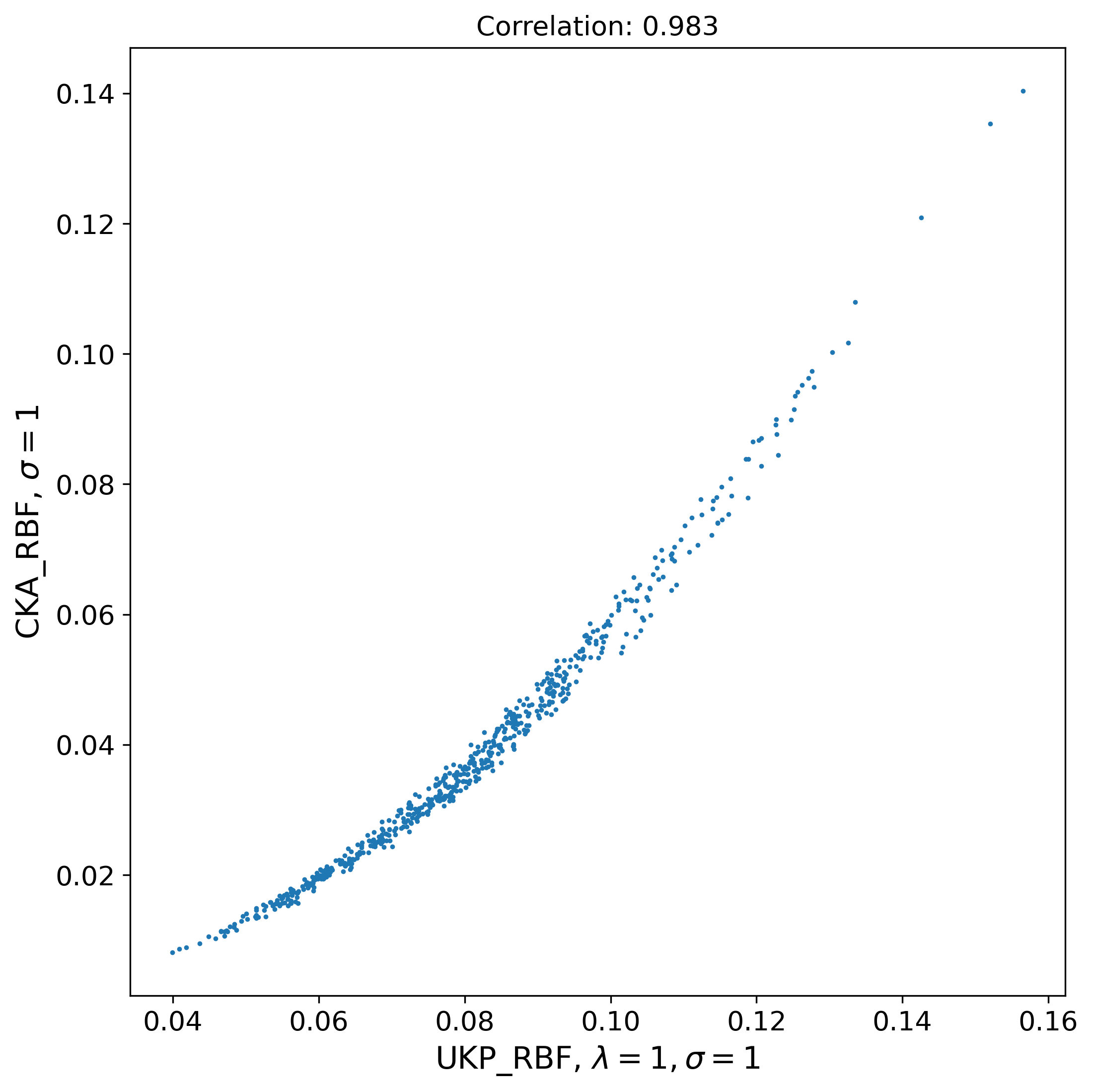}
    \end{subfigure}
    \hfill
    \begin{subfigure}[b]{0.45\textwidth}
        \includegraphics[width=\textwidth]{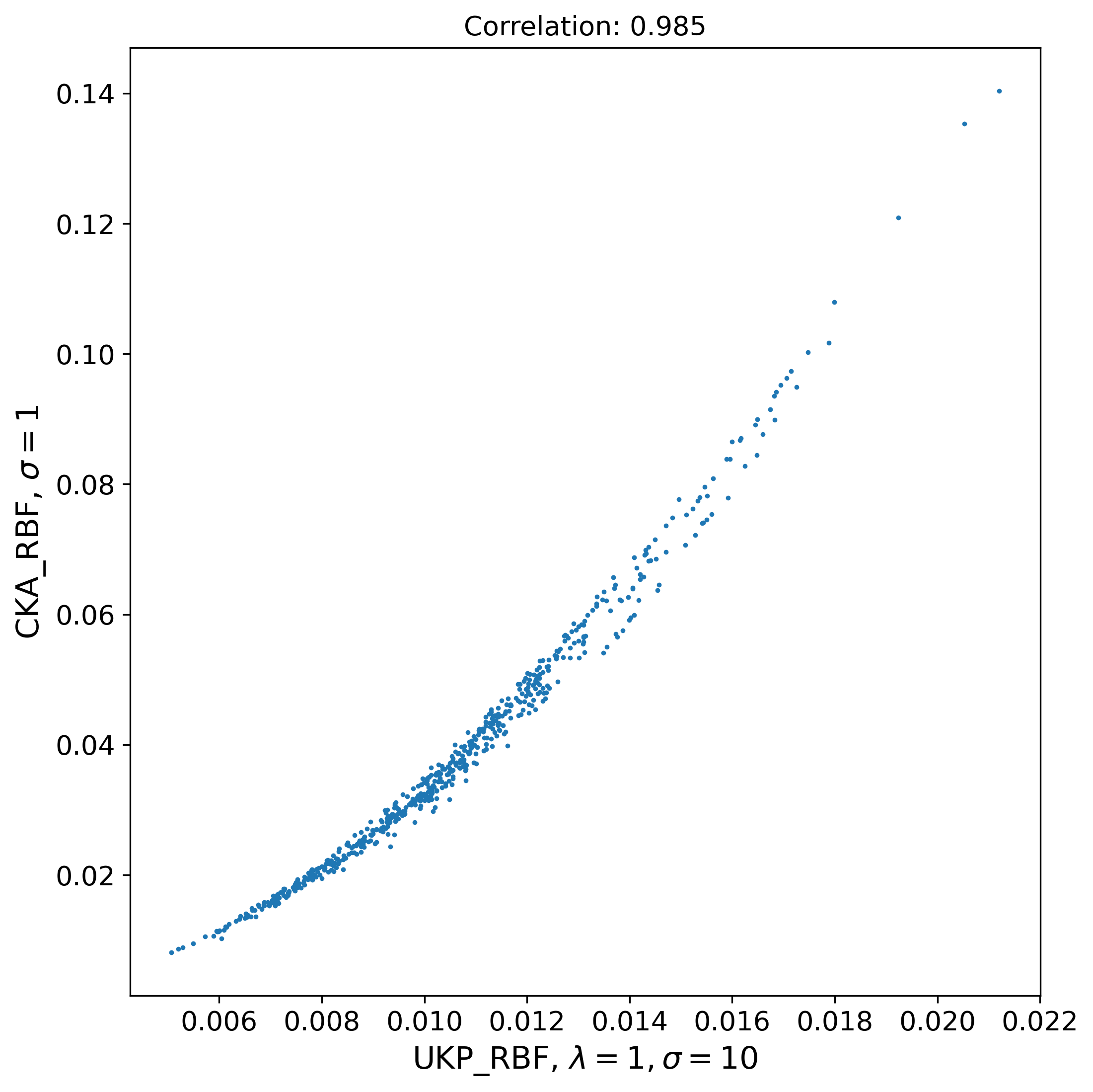}
    \end{subfigure}
    
    \vspace{0.5cm}  
    
    \begin{subfigure}[b]{0.45\textwidth}
        \includegraphics[width=\textwidth]{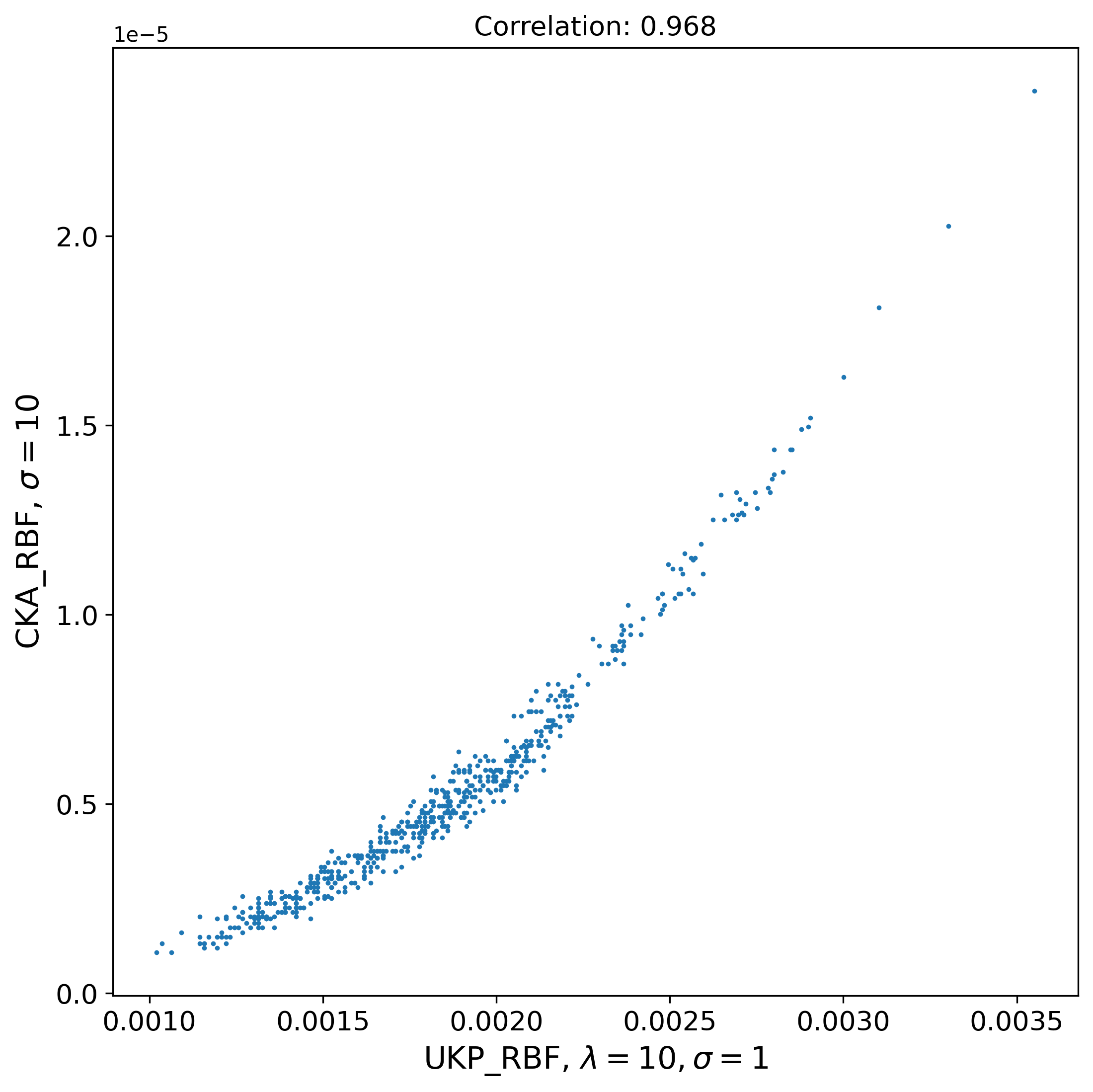}
    \end{subfigure}
    \hfill
    \begin{subfigure}[b]{0.45\textwidth}
        \includegraphics[width=\textwidth]{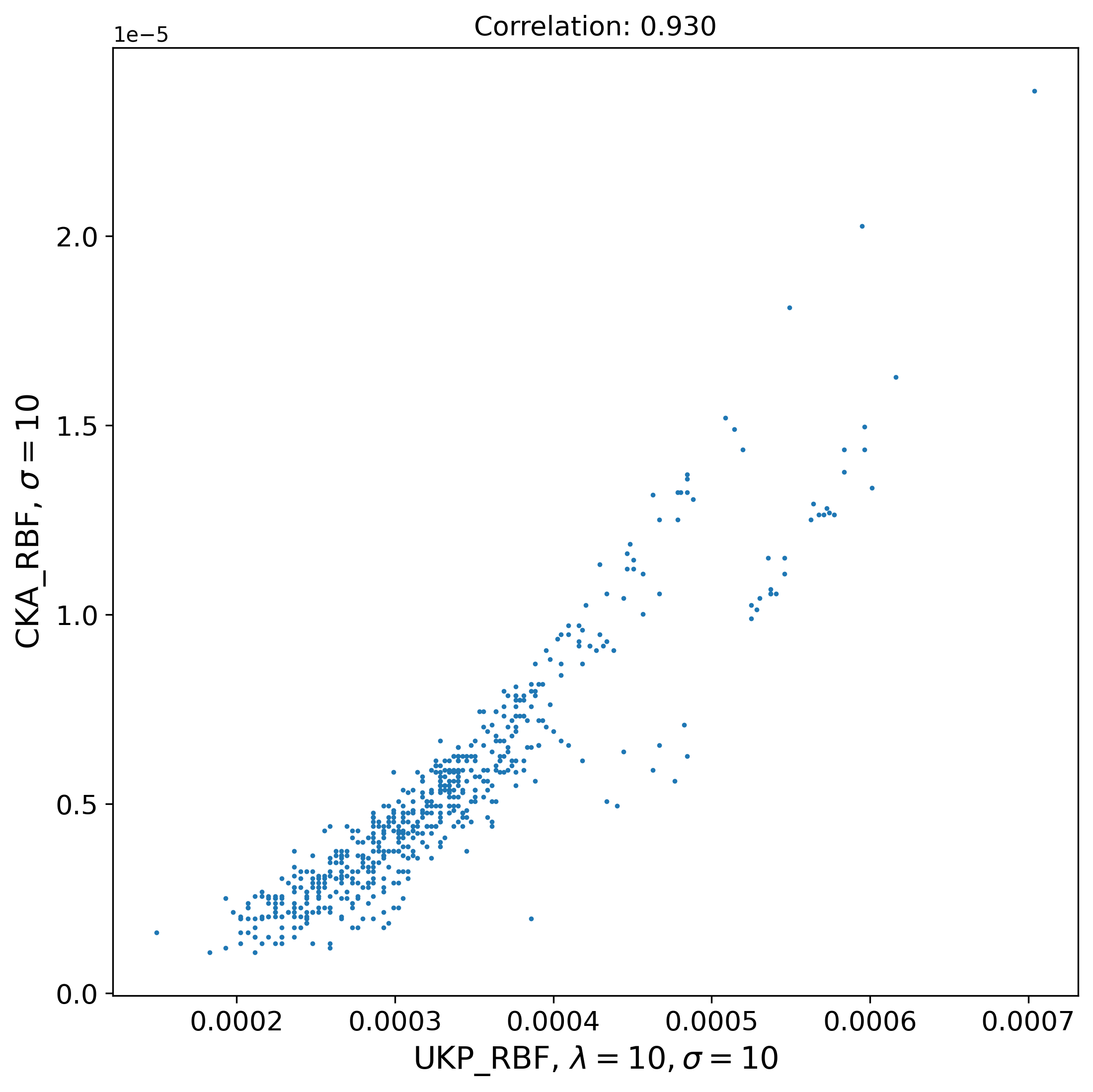}
    \end{subfigure}
    
    \caption{Correlation plots between UKP and CKA measures with Gaussian RBF kernel between $\binom{35}{2}$ pairs of networks with different architectures trained on  ImageNet data. Plot titles display the Pearson product-moment correlation coefficient between the distance measures on the two axes.}
    \label{ImageNet correlation plots UKP CKA}
\end{figure}

The MNIST experiments, along with the theoretical analysis in section \ref{Relation to other comparison measures}, reveal a similarity between the information conveyed by the \metricstname and CKA measures when both use the same kernel. This similarity is also empirically confirmed in the ImageNet experiments, as demonstrated by their scatterplots and the Pearson correlation coefficient across different tuning parameters. As illustrated in Fig. \ref{ImageNet correlation plots UKP CKA}, there is an almost linear positive relationship between \metricstname and CKA distances when both utilize a Gaussian RBF kernel. The strong positive correlation suggests that either measure could be effectively used for comparing representations. However, as previously discussed in Section \ref{Relation to other comparison measures}, \metricstname may be preferred over CKA due to its pseudometric properties, particularly the triangle inequality, which is especially advantageous. In contrast, CKA, being a measure similar to a normalized inner product bounded between 0 and 1, does not satisfy pseudometric properties and may lead to misleading interpretations when comparing different representations.

\paragraph{Choice of kernel function} The choice of kernel function for the \metricstname pseudometric should be guided by the inductive bias most relevant to the tasks for which the representations or features of interest will be used. For instance, consider an image classification task where the model's predictions should remain unaffected by image rotations. In this case, we can incorporate this inductive bias into the \metricstname pseudometric by selecting a rotationally invariant kernel, such as the Gaussian RBF kernel, as the kernel function for UKP. This approach is particularly useful for comparing the generalization performance of two representations: one obtained through a training or optimization procedure that explicitly enforces rotational invariance and another trained without such constraints.

Furthermore, even when the true inductive bias is unknown, probing the nature of representations encoded by different models can still provide valuable insights. In this context, the terms ``well-specified" and ``misspecified" kernels refer, respectively, to choices of kernels for the UKP pseudometric that either capture or fail to capture the required inductive bias for a specific class of downstream tasks utilizing the representations or features of interest. Each kernel choice can be viewed as a selection of particular characteristics of the representations that we aim to investigate.

If we have a set of characteristics in mind that we wish to probe, we should select a corresponding set of kernels whose feature maps encode some or all of those characteristics and then analyze the conclusions drawn from using each kernel as the kernel function for the \metricstname pseudometric. When the kernels are ``well-specified", clustering representations based on \metricstname values can help identify useful pairs of representations for specific downstream tasks. In contrast, when the kernels are ``misspecified", the \metricstname values may still cluster representations with characteristics aligned with the feature maps of the ``misspecified" kernels. However, in such cases, the clustering will not be informative for studying generalization performance on downstream tasks. Nonetheless, even with ``misspecified kernels", the \metricstname pseudometric can still provide insights into the characteristics of the representations, though its values will not reliably indicate generalization performance.

Cross-validation or selecting an ``optimal" value for the kernel parameters is not necessary in the context of this paper, as our focus is on an exploratory comparison of the inductive biases encoded by different representations. For example, consider a scenario where we hypothesize that rotational invariance is the key inductive bias required for good generalization performance, as in image classification tasks. In this case, the Gaussian RBF kernel is a natural choice. Since the Gaussian RBF kernel remains rotationally invariant for any value of its bandwidth parameter—which controls the ``scale" at which the kernel perceives the representations—the \metricstname pseudometric should, in principle, capture the extent to which different representations encode rotational invariance, regardless of the specific choice of bandwidth.

Of course, no experimental setup is ever exhaustive. In our study, we focus on datasets from the image domain (MNIST and ImageNet) to illustrate one of the simplest and most fundamental invariances—rotational invariance—which is relevant to most image-related tasks. This consideration motivated our choice of the Gaussian RBF kernel as the kernel function for the \metricstname pseudometric in our experiments.

\paragraph{Code implementation} The Python code for running all the experiments in this paper is available in the following GitHub repository: \url{https://github.com/Soumya-Mukherjee-Statistics/UKP-Arxiv}. The code for comparing our proposed \metricstname pseudometric to other distance measures has been adapted from \url{https://github.com/sgstepaniants/GULP}.

\end{document}